\documentclass[Afour,sageh,times]{sagej}

\usepackage{moreverb,url}

\usepackage[colorlinks,bookmarksopen,bookmarksnumbered,citecolor=red,urlcolor=red]{hyperref}
\usepackage{graphicx}
\usepackage{epstopdf}
\usepackage{subfig}
\usepackage{apalike}
\usepackage{setspace}
\usepackage{amsmath}
\usepackage{multirow}
\usepackage{tabu}
\usepackage{amssymb}
\usepackage{diagbox}
\usepackage[table]{xcolor}
\usepackage{natbib} 
\usepackage{caption}
\DeclareMathOperator*{\argminB}{argmin}   
\DeclareMathOperator*{\argmaxB}{argmax} 
\graphicspath{ {pic/} }
\newcommand\BibTeX{{\rmfamily B\kern-.05em \textsc{i\kern-.025em b}\kern-.08em
T\kern-.1667em\lower.7ex\hbox{E}\kern-.125emX}}

\def\volumeyear{2016}

\begin{document}

\runninghead{Shuang Liu, Mete Ozay, Takayuki Okatani, Hongli Xu, Kai Sun and Yang Lin}

\title{A vision based system for underwater docking}

\author{Shuang Liu\affilnum{1,2,3} , Mete Ozay\affilnum{3}, Takayuki Okatani\affilnum{3,4}, Hongli Xu\affilnum{1,2}, Kai Sun\affilnum{1} and Yang Lin\affilnum{1,2}}

\affiliation{\affilnum{1} State Key Laboratory of Robotics, Shenyang Institute of Automation, Chinese Academy of Sciences, Shenyang, China\\
\affilnum{2}  University of Chinese Academy of Sciences, Beijing, China \\
\affilnum{3} Graduate School of Information Sciences, Tohoku University, Sendai, Miyagi, Japan\\
\affilnum{4} RIKEN Center for Advanced Intelligence Project, Tokyo, Japan}

\corrauth{\{Shuang Liu, Mete Ozay\}, Graduate School of Information Sciences, Tohoku University, Sendai, Miyagi, Japan.}

\email{\{liushuang,mozay\}@vision.is.tohoku.ac.jp}

\begin{abstract}
Autonomous underwater vehicles (AUVs) have been deployed for underwater exploration. However, its potential is confined by its limited on-board battery energy and data storage capacity. This problem has been addressed using docking systems by underwater recharging and data transfer for AUVs. In this work, we propose a vision based framework for underwater docking following these systems. The proposed framework comprises two modules; (i) a detection module which provides location information on underwater docking stations in 2D images captured by an on-board camera, and (ii) a pose estimation module which recovers the relative 3D position and orientation between docking stations and AUVs from the 2D images. For robust and credible detection of docking stations, we propose a convolutional neural network called Docking Neural Network (DoNN). For accurate pose estimation, a perspective-n-point algorithm is integrated into our framework. In order to examine our framework in underwater docking tasks, we collected a dataset of 2D images, named Underwater Docking Images Dataset (UDID), in an experimental water pool. To the best of our knowledge, UDID is the first publicly available underwater docking dataset. In the experiments, we first evaluate performance of the proposed detection module on UDID and its deformed variations. Next, we assess the accuracy of the pose estimation module by ground experiments, since it is not feasible to obtain true relative position and orientation between docking stations and AUVs under water. Then, we examine the pose estimation module by underwater experiments in our experimental water pool. Experimental results show that the proposed framework can be used to detect docking stations and estimate their relative pose efficiently and successfully, compared to the state-of-the-art baseline systems.
\end{abstract}


\maketitle

\section{Introduction}
\label{l_intro}
In various underwater docking tasks, autonomous underwater vehicles (AUVs) aim to detect, localize and physically attach to docking stations \citep{bellingham2016autonomous}, such as sea survey \citep{smith2010planning,yoerger2007techniques} and in-water ship hull inspection \citep{englot2013three}. AUVs belong to the category of unmanned underwater vehicles (UUVs). UUVs are categorized into two main groups: i) remotely operated vehicles (ROVs), which require a user (human operator) input through a cable, and ii) autonomous underwater vehicles (AUVs), which provide stand-alone platforms without human supervision and support themselves by their on-board resources. Cables of ROVs supply adequate power and enable communication, but confine the scope of their activities. AUVs offer numerous advantages over ROVs such as a wider scope of activity, more compact size and higher efficiency, free from the limitation of a physical connection to an operator. However, its potential struggles with its finite on-board battery energy, processing and storage capacity. 

Underwater docking has been popularly used due to its ability of autonomous battery recharging and data transfer, by making long-term underwater residence possible. Underwater docking systems guide AUVs into predesignated docking stations by using compatible sensors. Three types of sensors are used for an underwater docking task: i) electromagnetic \citep{feezor2001autonomous}, ii) acoustic \citep{hong2003development}, and iii) optical sensors \citep{park2009experiments}.
Optical sensors outperform others in terms of good directional accuracy, low vulnerability to external detection and capacity for multiple tasks, but suffer from good propagation in an underwater environment owing to the speedy attenuation of light in the water \citep{deltheil2000simulating}. Therefore optical sensors are usually utilized to take responsibility of the final short-distance stage precise docking, and they are combined with other sensors which are superior in propagation but inferior in accuracy \citep{maki2013docking}. 

Widely used vision based underwater docking (VBUD) systems consist of docking stations, cameras mounted on AUVs and docking algorithms on AUVs. Dedicated landmarks on docking stations are necessarily used by AUVs to identify docking stations. Landmarks may either be passive or active. Passive landmarks, such as patterns drawn on a board, do not need energy supply, but they are only visible within a close range. Active landmarks, such as light beacons, have high visibility by emitting energy. Active landmarks are usually preferred for its high visibility in far distance compared to passive ones \citep{bianco2013comparative}. In underwater docking systems, either monocular or binocular cameras are used. Binocular cameras require large camera baselines and more computational resources for detection of distant objects \citep{negre2008robust}.

In this work, we consider VBUD systems equipped with active landmarks and a monocular camera. VBUD algorithms compute positions of docking stations using images captured by the on-board camera \citep{park2009experiments,li2015auv}. VBUD algorithms are employed in two phases: i) detection of docking stations, and ii) estimation of pose between AUVs and docking stations \citep{negre2008robust}. In the detection phase, docking stations are located in 2D images captured by cameras. Pose estimation recovers 3D relative position and orientation between AUVs and docking stations from the detected image patch.

Successful underwater docking algorithms demand on several conditions. First, detection of docking stations should be credible. Observing the docking station for large number of times (e.g. 100 times), but only successfully detecting once, will be inefficient and a crucial fault for AUVs which will run out of battery. Second, detection methods implemented by VBUD algorithms should be robust to blurring, color shift, contrast shift, mirror images, non-uniform illumination and noisy luminaries observed in non-stationary underwater environments. Finally, algorithms used for pose estimation are required to be fast, accurate and robust to noise.  

More precisely, underwater images are prone to be blurring, color shift, reduced contrast and non-uniform illumination in different underwater conditions due primarily to the optical properties of the water medium, in contrast to images in air \citep{kaeliillumination}. Water is a strong attenuator of electromagnetic radiation. Some part of the incident light energy is absorbed by water molecules, and some part is scattered out during its propagation through the water. The spectral absorption is wavelength and distance dependent. The light energy exponentially decays with respect to the propagation distance. Red light is more strongly absorbed due to its longer wavelength. In addition, absorption coefficient of natural water varies depending on water quality which is a sum of contribution by various constituents, such as dissolved salts, organic compounds and phytoplankton \citep{mobley1994light}. Factors above dominantly cause the varying degrees of color shift in underwater images. Scattering is divided into forward scattering and backward scattering depending on the scattering angle. The former results in blurring and low contrast while the latter results in a visible bright haze in images \citep{bryson2015true}. Moreover,  underwater noisy luminary may come from ambient noise light, water-surface bubble mixed with oil or other underwater light sources as shown in \cite{park2009experiments}. Noisy luminary annoys binary object detection methods crucially. Removing all possible noisy luminary requires employment of image pre-processing methods developed by domain experts. In addition, deformed objects are observed due to scale and rotation variance. Images of docking stations captured at different locations and viewpoints (orientation degrees) give rise to geometric deformations, addressing a challenging problem for underwater docking. Under certain conditions, mirror images of docking stations are also observed due to \textit{total internal reflection}. When light propagates from water to the air and crosses the water-air boundary, some part of the light is refracted while the rest is reflected back into the water. According to Snell's law, there exists a \textit{critical angle} where light is totally reflected back with zero refraction. Total internal reflection occurs at all angles  smaller than critical angle, and thereby water surface serves as a mirror. Mirror images are almost identical copies of original images of docking stations. Their similar appearances confuse detection and recognition algorithms implemented in AUVs. 

In order to address above problems, we propose a underwater docking framework that integrates a convolutional neural network (CNN) architecture proposed for robust detection of underwater docking stations, and a perspective-n-point algorithm employed for fast, accurate and robust pose estimation. 
Our contributions can be summarized as follows:
\begin{itemize}
    
    \item We provide our underwater docking dataset which was collected in our experimental water pool. To the best of our knowledge, it is the first publicly available dataset used for analysis of computer vision algorithms employed for underwater docking. We labeled bounding boxes of docking stations for each image in the dataset. The dataset can help researchers to develop underwater docking algorithms and validate their algorithms in absence of underwater docking infrastructures. In addition, a series of deformation methods are proposed in this study to generate deformed \textit{realistic} underwater images as close as possible to real-world undersea images. 
    
    \item We propose a convolutional neural network named DoNN for the detection of underwater docking stations. It has two main advantages compared to state-of-the-art networks. First, it is credible. It achieves $0.99964$ AUC (experimental analyses are given in Section \ref{sec:performevaluation}) for detection of underwater docking stations (see Section \ref{sec:detectperformoriginal}). Second, the proposed detection approach is robust to various deformations, such as blurring, color shift, contrast shift, mirror images, non-uniform illumination and noisy luminary, in various complex and dynamic underwater environments. It outperforms baseline models in terms of AUC using underwater images with blurring, color shift, contrast shift and mirror images. It achieves slightly inferior but acceptable AUC performance on underwater images with non-uniform illuminations and noisy luminaries compared to baseline models.
     
    \item We integrate a perspective-n-point algorithm termed RPnP, which is fast, accurate and robust to noise, into our framework for estimation of relative position and orientation between docking stations and AUVs. Ground experiments show that the average error of position and orientation of pose estimation module are $5.927$ mm and $1.970$ degree, respectively, and the running time is $0.036$ second per frame. It achieves $9.432$ mm and $2.353$ degree in terms of average error of position and orientation in presence of strong noise.
  
\end{itemize}

\section{Related work}
\label{gen_inst}
In recent years, several VBUD solutions have been proposed for underwater docking. It was first verified that optical terminal guidance acquisition ranges of 10 meters - 15 meters are possible even in very turbid water by \cite{cowen1997underwater}. In their study, docking algorithms guide AUVs by using distribution of landmarks in all four quadrants of images. Their method is simple but effective. False detection occurs in shallow water due to sunlight. In \cite{cowen1997underwater}, one light was used as an active landmark. \cite{hong2003development} designed a cone shape docking station with six color lights. Five of them form a pentagon while one lies outside to prevent mutual inference. \cite{park2009experiments} take advantage of docking stations in similar shape; they used four white lights on the rim of the upside of the circle and one light on the other semicircle. \cite{li2015auv} placed four 540 nm green lights around the rim of a cone shape docking station to achieve good underwater propagation of green lights. All landmarks mentioned above are arranged on a plane. \cite{maki2013docking} leveraged 3D landmarks with three green lights and one red light for underwater docking.

Configuration of landmarks is tightly coupled with the mechanism of VBUD algorithms. VBUD algorithms perform detection station detection and pose estimation tasks. Detection of docking station provides location of docking stations in 2D images captured by an on-board camera. Underwater docking detection algorithms fall into two general categories: binarization based methods and feature based methods. Binarization based methods binarize images using a threshold followed by a series of image preprocessing methods due to limited prior knowledge \citep{park2009experiments,ghosh2016reliable}. \cite{park2009experiments} first binarize images by a pre-defined fix threshold, and then remove salt-and-pepper noise using a $3 \times 3$ mask. \cite{ghosh2016reliable} improve the threshold selection method using an invariant histogram-based adaptive thresholding method. They did not use image pre-processing methods due to their relative ideal test environment. Performance of binarization based methods are sensitive to user defined thresholds, and require a well-round expert knowledge in different underwater environments in order to perform inference from non-uniform illumination, reduced contrast, blurring, color shift, mirror images and noisy luminaries. As an example of feature-based detection methods, \cite{li2015auv} first adopted Mean-Shift algorithm to extract light source area, and then took advantage of the Snake algorithm to recognize contour features. Finally, SVMs are used to detect features of landmarks. Although various detection approaches are proposed in previous underwater docking works, none of them reported their detection performance and analyzed credibility and robustness of detection methods in detail.

Pose estimation in underwater docking refers to recovering 3D relative position and orientation between docking stations and AUVs from 2D images. Both monocular and binocular cameras are available for this problem. Binocular methods estimate position information through disparity maps. They require longer processing time and large baselines for accurate estimation \citep{negre2008robust}. \cite{myint2016visual} estimate pose by using a binocular camera. \cite{li2015auv} proposed a pose estimation algorithm which combines binocular and monocular methods. \cite{park2009experiments} utilized a monocular camera for pose estimation. They established an algorithm that maps number of pixels to distance between AUVs and docking stations. In their case, exact locations of docking stations cannot be gained owing to factors like scattering. In order to obtain an exact pose by using monocular cameras,  perspective-n-points (PnP) algorithm is used. PnP algorithm is able to recover pose information given points correspondence between 2D images and 3D points. A unique solution can be obtained using more than three correspondence points ($n>3$). \cite{li2015auv} took advantages of P3P ($n=3$) to estimate pose in its monocular module. \cite{ghosh2016reliable} estimated poses by fitting ellipses. They mounted twelve lights around a cone shape docking station. An ellipse was fit, and parameters of the ellipse were used to estimate the pose. Their experiments were only carried out in the range of less than 140cm which is not convincing in real underwater docking tasks (10 meters - 15 meters). 

\begin{figure*}[t]
\centering
\captionsetup[subfigure]{justification=centering}
    \subfloat[][]{
    \label{fig:sialake}
    \includegraphics[width= 0.26\linewidth,height= 0.21\linewidth]{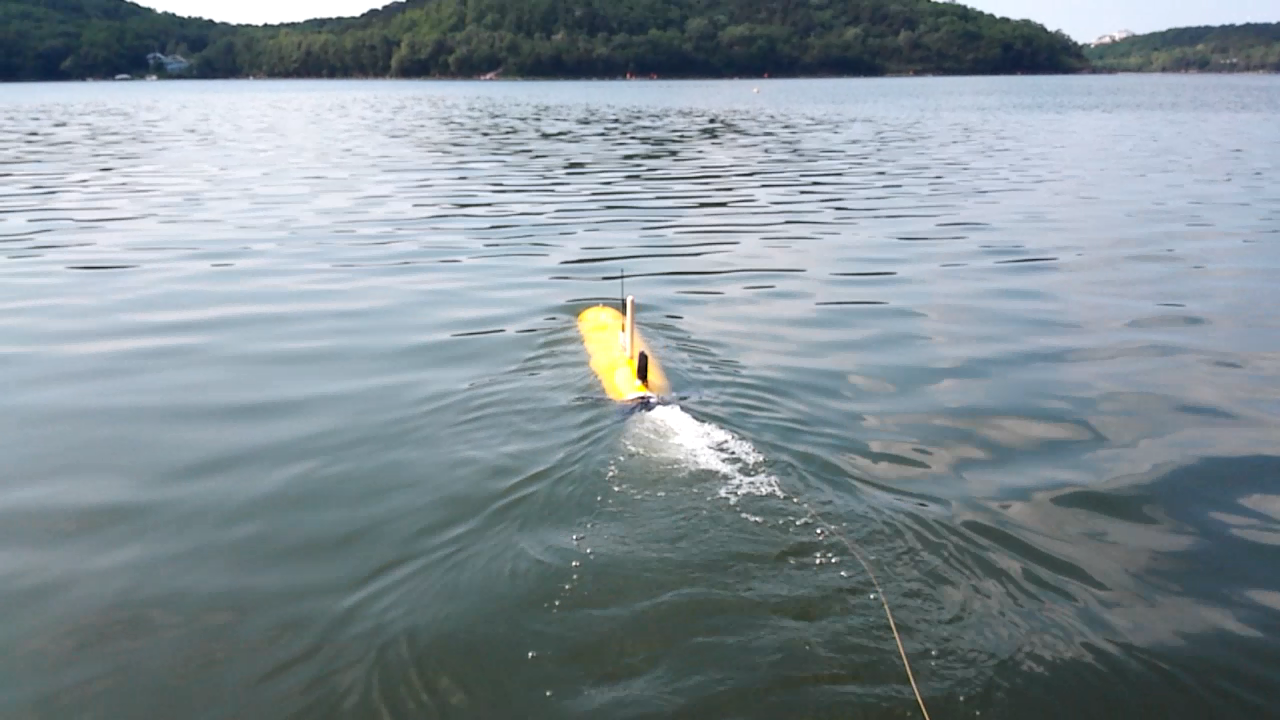}}
    \subfloat[][]{
    \label{fig:refittingauv}
   \includegraphics[width= 0.26\linewidth,height= 0.21\linewidth]{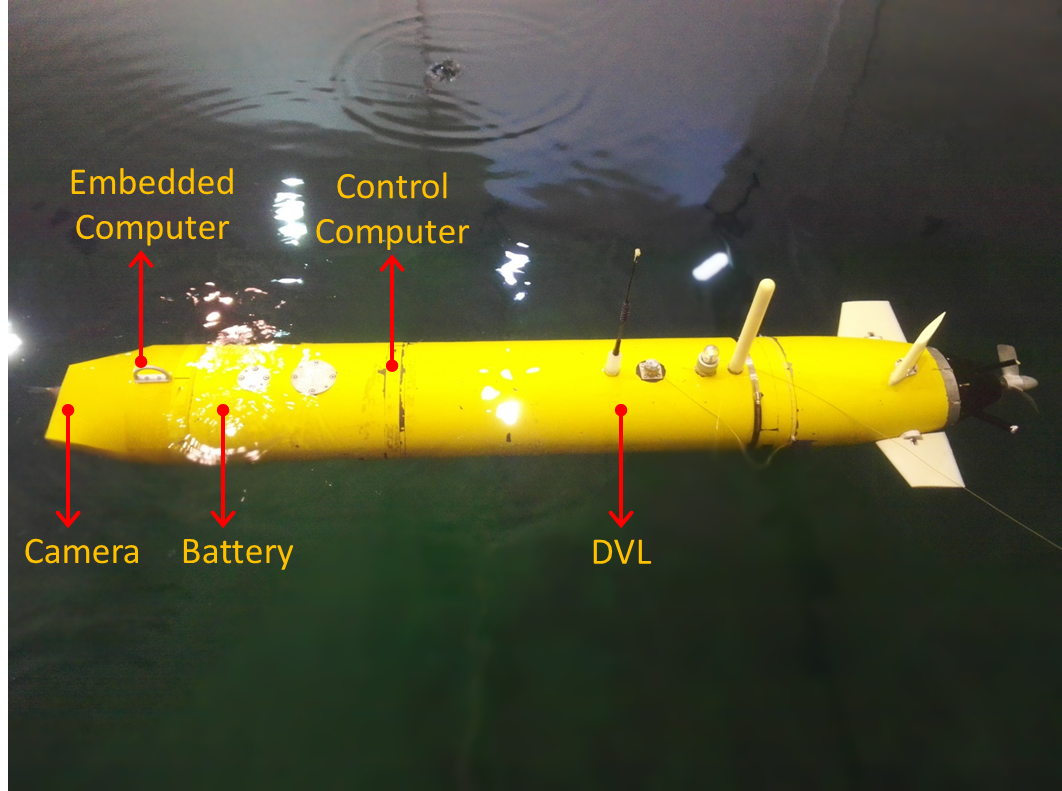}}
    \subfloat[][]{
    \label{fig:auvhead}
   \includegraphics[width= 0.26\linewidth,height= 0.21\linewidth]{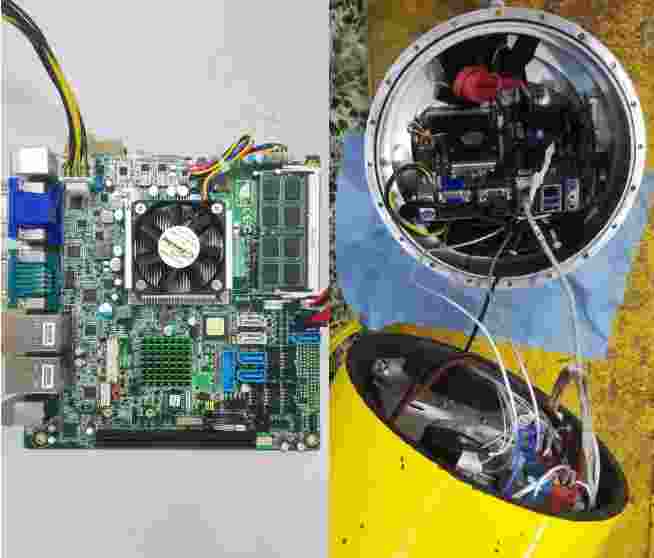}
    }\\
    \subfloat[][]{
    \label{fig:stationwatersurface}
   \includegraphics[width= 0.26\linewidth,height= 0.21\linewidth]{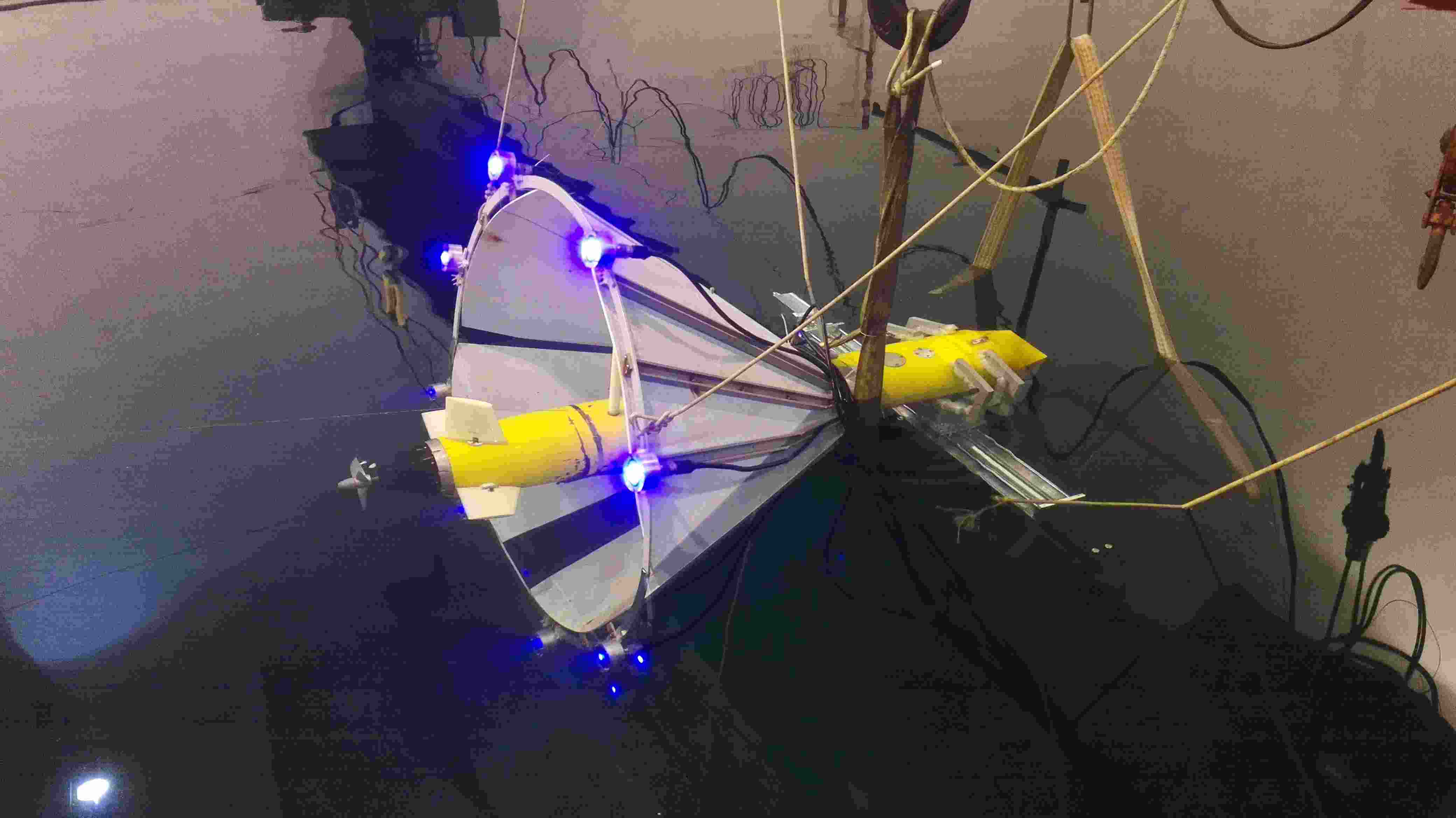}
    }
    \subfloat[][]{
    \label{fig:stationunderwater}
    \includegraphics[width= 0.26\linewidth,height= 0.21\linewidth]{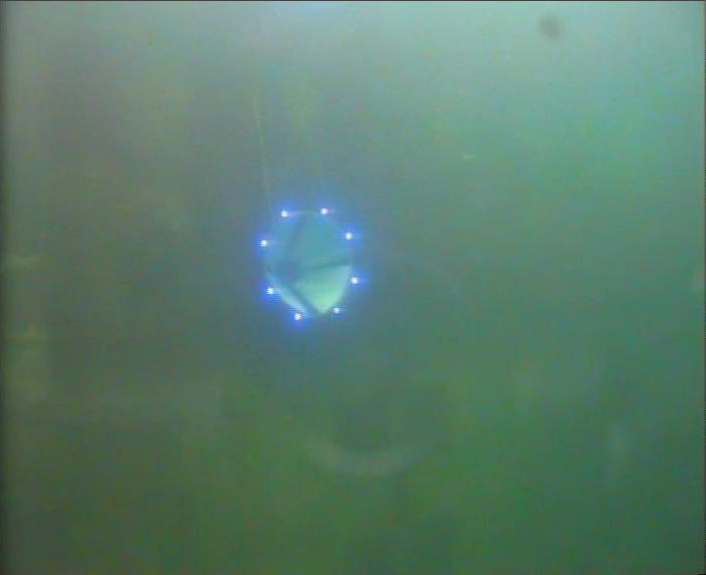}
    }

\caption{Overview of our system used in the experiments. (a) A picture of the SIA-9 in the lake. (b) Illustration of different modules in the SIA-9 after refitting for underwater docking. (c) The embedded computer used for underwater docking. The appearance of our docking station (d) on water surface, and (e) in water.}
\label{fig:sia9}
\end{figure*}

\section{System overview}
\label{systemoverview}
This section introduces an overview of our proposed underwater docking system. 

In order to perform our experiments, we developed a AUV research platform (SIA-9) which is a small torpedo-shaped vehicle as shown in Figure \ref{fig:sia9}. Its specifications are given in Table \ref{tab:auvspecification}. SIA-9 is mainly equipped with Doppler Velocity Log (DVL), Compass, Inertial Measurement Unit, Radio, GPS, a control computer, battery units and motors. 
The software architecture is based on MOOS-IvP \citep{benjamin2009overview} which is used to separate overall capability into distinct modules. Each module acts as a MOOS process which can publish its own information and subscribe others' to and from MOOSDB. MOOSDB is a module used for exchanging information between different MOOS processes, and which is responsible for maintaining the consistency of information. Implementation details of the SIA-9 were given by \cite{jia2016development}.

\begin{table}[h!]
    \centering
    \caption{Specifications of the SIA-9 used in the experiments.}
\begin{tabular}{l|c}
\hline
Item & Value \\
\hline
Diameter & 250 mm \\
Length   & 1576 mm \\
Weights in air & 75 kg \\
Maximum operating depth & 300 m \\
Maximum cruising speed & 3 knots \\
\hline
\end{tabular}
    \label{tab:auvspecification}
\end{table}
In order to use the SIA-9 for our underwater docking task, three modules were installed on the SIA-9. First, a forward-looking RGB monocular camera with frame rate 20 fps was rigidly mounted inside the head of the SIA-9. Second, we installed an embedded computer which is shown in Figure \ref{fig:auvhead} inside the head of SIA-9 to process underwater images. The embedded computer was equipped with Intel Core i7 3.4 Ghz processor, 8 Gb RAM and a 64-bit operating system. It was used to establish communication between the control computer and the camera through LAN and PCIe, respectively. 
Third, the head of the SIA-9 was redesigned to fit the camera and the embedded computer as shown in Figure \ref{fig:refittingauv}.

Our docking station is cone-shaped with outer diameter 1200 mm and inner diameter 300 mm. Eight active landmarks were mounted uniformly around the rim of the docking station as shown in Figure \ref{fig:stationwatersurface} and \ref{fig:stationunderwater}. Blue LED lights with 460 nm wavelength were used as landmarks due to good propagation of blue light in water. 

\section{Underwater docking algorithm}
\label{sec_underwaterdockingalgorithm}
In this section, we provide our underwater docking algorithm consisting of two modules which are used for i) detection of underwater docking stations, and ii) estimation of pose between docking stations and AUVs. The detection module takes underwater images as input, and outputs location of underwater docking stations in 2D images. In other words, it is used to determine whether the docking station is within the field of view of AUVs, and where it is located in the captured image. Pose estimation module computes the relative position and orientation between AUVs and docking stations once the predesignated docking station is detected. AUVs conduct a line tracking task, taking current position as the start point and the position given by the pose estimation module following the detection and pose estimation phases. Line tracking is a common procedure used to operate AUVs, and its analysis is out of scope of this work.
\subsection{Detection of underwater docking station using deep neural networks}
\label{sec_detection}
A robust and credible detection algorithm is highly desirable in practical underwater docking as mentioned in Section \ref{l_intro}. Underwater docking detection suffers from blurring, color shift, reduced contrast, mirror images, non-uniform illumination and noisy luminaries more compared to overwater detection tasks. 
A robust detection can guarantee detection performance in various non-stationary underwater environment while a credible one can improve the docking efficiency. The docking efficiency is crucial for AUVs which are in low battery state, and which will recharge their battery by underwater docking. However, none of them draw enough attention in the previous works, and no detection performance was reported with a detailed analysis. In this work, we propose a convolutional neural network  (CNN), called Docking Neural Network (DoNN), inspired by the YOLO \citep{redmon2016you} aiming at robust and credible detection of docking stations. In this section, we first provide a brief background of CNNs, and then introduce our proposed DoNN for detection of underwater docking stations.

\subsubsection{Background of convolutional neural networks}
\label{CNNfoundation}

A CNN is a deep neural network (DNN) which employs convolution layers as building blocks to model spatial patterns. CNNs draw several key ideas from Hubel and Wiesel's discovery on cat's visual cortex \citep{hubel1959receptive}, and they have been used as one of the most successful methods used to perform robot vision tasks in the recent years \citep{oliveira2017efficient,schwarz2016rgb,levine2016learning}. A CNN is used to estimate a function $f$ defined by  ${{Y}}={f}({X};\boldsymbol{\theta})$, where $X$ and $Y$ is input and output of the CNN. The estimated function $\hat{f}({X};\boldsymbol{\theta})$ is parameterized by the network parameters (weights) $\boldsymbol{\theta}$ of the CNN. In the training phase, the network parameters are estimated by minimizing a loss function $l(\boldsymbol{\theta})$ as
\begin{equation}
\label{eq_thetal}
    \boldsymbol{\theta_L} = \argminB_{\boldsymbol{\theta}} l(\boldsymbol{\theta}).
\end{equation}
A forward propagation follows the training step for prediction of the output ${\hat{Y}}={f}({X};\boldsymbol{\theta})$. A typical CNN layer consists of three basic operations; convolution, nonlinear activation function and pooling. In CNNs, a convolution operation is defined for an input 2D image $\boldsymbol{I}$ by
\begin{equation}
    \tau(x,y) = (\boldsymbol{I} \otimes \boldsymbol{F})(x,y) = \sum_{m,n}\boldsymbol{I}(x+m,y+n)\boldsymbol{F}(m,n),
    \label{eq:featmaptau}
\end{equation} 
where $\otimes$ denotes the convolution operation, $\boldsymbol{F} \in{\mathbb{R}^{m\times n}}$ denotes \textit{filters} or \textit{kernels} and $\tau(x,y)$ is the value of \textit{feature maps} $\tau$ computed at location $(x,y)$. After computation of a convolution step at a location $(x,y)$, the filter $\boldsymbol{F}$ shifts to the next location to perform the next step of the convolution, and the amount shift is controlled by \textit{stride}. 
    
In CNNs, convolution is usually followed by a nonlinear activation function. A commonly used nonlinear activation function is rectified linear unit (ReLU) \citep{nair2010rectified}. Pooling is a form of non-linear down-sampling. It substitutes the input $\boldsymbol{I}(x,y)$ at a location $(x,y)$ with $\boldsymbol{\Psi}(N(x,y))$, where $N(x,y)$ denotes a neighbourhood of $(x,y)$, and $\boldsymbol{\Psi}$ is a pooling function with summary statistics such as \textit{max pooling} \citep{zhou1988computation} which utilized an operation ${\max(a,b) =a, \quad if \quad a>b}$. The neighbourhood $N(x,y)$ of $(x,y)$ can be also viewed as a sliding window $w_p$ centering at $(x,y)$. $w_p$ slides across every possible location $(x,y)$ of $\boldsymbol{I}$ step by step. If max pooling is used, then it takes the largest element within $w_p$ at each step. Stride of pooling controls the shifting units used at each pooling step. The amount of parameters and computational burden are significantly reduced through pooling. Meanwhile pooling provides invariance to small translations of inputs within the receptive field of the corresponding units (neurons) of the CNN.

CNNs are \textit{efficient} in terms of memory requirements and statistical efficiency. The efficiency is obtained by two main features: local connectivity and parameter sharing. Local connectivity means that each neuron is connected only to a local region of its input. Parameter sharing is based on the assumption that different patches of local regions share some collective features, e.g. edges. These two features significantly reduce the amount of parameters of CNNs. 

\subsubsection{Docking Neural Network}
\label{sec:donn}
In this subsection, we introduce our docking neural network (DoNN) proposed for detection of underwater docking stations. Object detection is one of the most important tasks studied in robot vision. Generally speaking, CNN based detection approaches fall into two categories; region proposal based detection and proposal free detection \citep{Kong2016}. In region proposal based detection, such as Fast-RCNN \citep{girshick2015fast} and Faster-RCNN \citep{ren2015faster}, first some candidate regions are proposed by using another neural network, such as region proposal network (RPN) in Faster-RCNN or selective search \citep{uijlings2013selective} in Fast-RCNN. Then, objects are detected in the proposals.
Proposal free detection used by YOLO \citep{redmon2016you} poses detection as a regression problem. Bounding-boxes and their confidence are predicted simultaneously through one pass. DoNN is inspired by YOLO \citep{redmon2016you}. In our proposed method, we redesigned the loss function used in YOLO, in compatible with our datasets which contain one object. 

\begin{figure*}
    \centering
    \includegraphics[width = \linewidth]{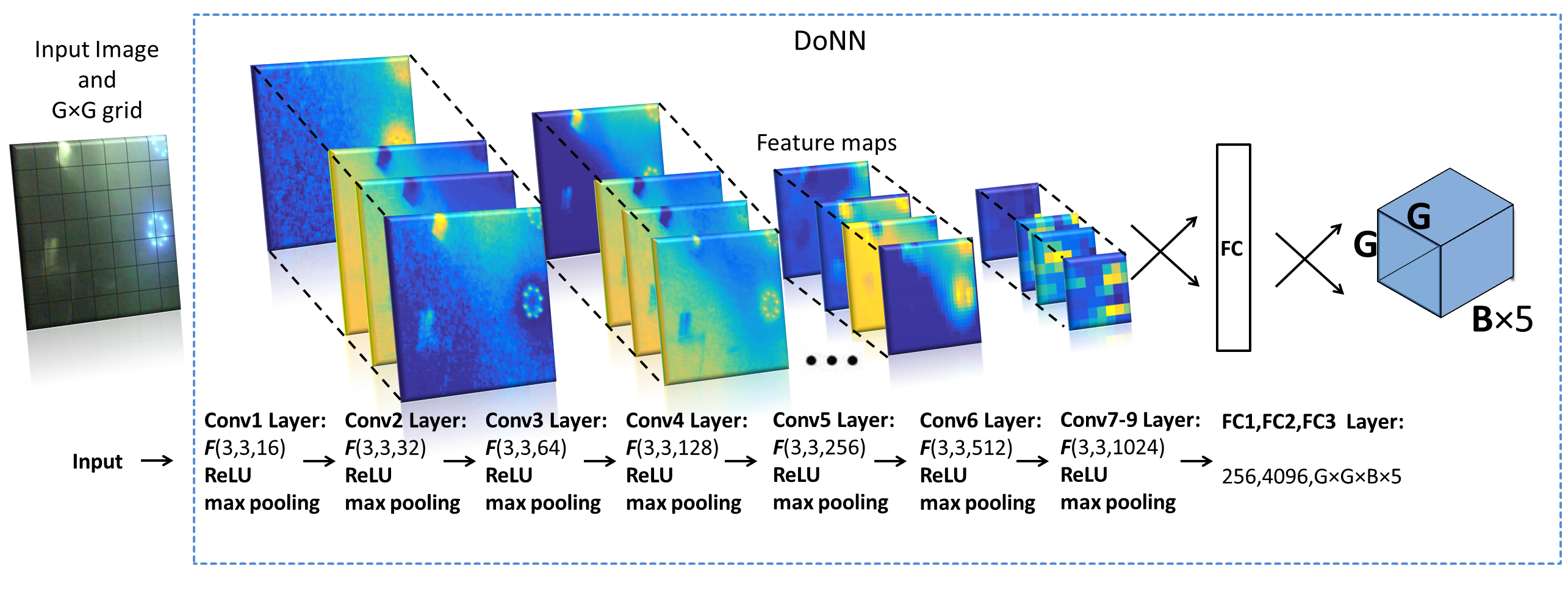}
    \caption{An illustration of the architecture of the proposed DoNN. DoNN consists of nine convolution layers (Conv), seven max pooling layers, and three fully connected layers (FC). \textit{F}(width,height,depth) indicates a convolution filter with the size of width, height and depth. Sliding windows $w_p$ of all max pooling layers have size $2 \times 2$. The stride is set to $1$ in all convolution layers, and it is set to $2$ in all max pooling layers. The number of neurons in the first fully connected layer (FC1), the second fully connected layer (FC2), and the third fully connected layer (FC3) is $256$,$4096$ and $G \times G \times B \times 5$, respectively. DoNN takes three channel images as input, and predicts a $G \times G \times B \times 5$ tensor from which the final detection is obtained.}
    \label{fig:net_archi}
\end{figure*}

DoNN consists of nine convolution layers and seven pooling layers. Its architecture is illustrated in Figure \ref{fig:net_archi}. Detection problem is considered as a regression problem in DoNN. 
In underwater docking tasks, DoNN maps underwater images to location of docking stations in images. DoNN learns feature representations of underwater docking stations from the whole image through minimization of the loss function given in \eqref{eq_loss} in the training phase, and predicts the location of docking stations on unseen underwater images by employing learned feature representations in the inference phase. 

DoNN takes the whole image as input. Input images fed are first divided into $G \times G$ grids, as illustrated in Figure \ref{fig:net_archi}. DoNN predicts the positions and sizes of multiple candidates for the bounding box along with their confidence score. We fix the number of the candidates and denote it by B.  We further explain bounding boxes and associated confidence score as follows.
\begin{enumerate}
\item $B$ bounding boxes: The $b^{th}$ bounding-box of the $i^{th}$ grid is denoted by $\mathcal{B}_{i,b} = (x_{i,b},y_{i,b},w_{i,b},h_{i,b})$, ${i\in \{1, 2, \ldots, G^2\}}$, $b\in{\{1,2,\ldots , B \}}$. The center of $\mathcal{B}_{i,b}$ is located in the $i^{th}$ grid, and $(x_{i,b},y_{i,b})$ is the coordinate of the center of the bounding box $\mathcal{B}_{i,b}$. It is represented by the offsets of the $i^{th}$ grid bound. The terms $w_{i,b}$ and $h_{i,b}$ denote the width and height of the bounding-box, respectively. They are divided by the image width and height to be normalized to the range between $0$ and $1$.

\item $B$ bounding-box confidence score: We denote a confidence score of the $b^{th}$ bounding-box residing in the $i^{th}$ grid by $S_{i,b}$. 
\end{enumerate}

DoNN is trained to minimize the discrepancy between predictions and manually labeled ground truth for each grid. This discrepancy is expressed by the loss function 
\begin{equation}
    \begin{gathered}
    l_{\mathit{DoNN}}(\boldsymbol{\theta}) =  \lambda_{\mathcal{B}}l_{\mathcal{B}}(\boldsymbol{\theta}) +  \lambda_{d}l_{d}(\boldsymbol{\theta})+ \lambda_{\bar{d}}l_{\bar{d}}(\boldsymbol{\theta}).
    \label{eq_loss}    
    \end{gathered}
\end{equation}
We compute each sub-loss function as follows;
\begin{equation}
\label{eq_bboxloss}
 \begin{aligned} 
    l_{\mathcal{B}}(\boldsymbol{\theta}) = \sum_{i=1}^{G^2}\sum_{b=1}^{B}l_{i,b}^{dock}[(x_{i,b}-\hat{x}_{i,b})^2
    +(y_{i,b}-\hat{y}_{i,b})^2]\\
    +\sum_{i=1}^{G^2}\sum_{b=1}^{B}l_{i,b}^{dock}[(\sqrt{w_{i,b}}-\sqrt{\hat{w}_{i,b}})^2+(\sqrt{h_{i,b}}-\sqrt{\hat{h}_{i,b}})^2],
    \end{aligned}
\end{equation}
\begin{equation}
    l_{d}(\boldsymbol{\theta}) = \sum_{i=1}^{G^2}\sum_{b=1}^{B}l_{i,b}^{dock}(S_{i,b}-\hat{S}_{i,b})^2,
    \label{eq_objloss}
\end{equation}
and
\begin{equation}
    l_{\bar{d}}(\boldsymbol{\theta})= \sum_{i=1}^{G^2}\sum_{b=1}^{B}l_{i,b}^{nodock}(S_{i,b}-\hat{S}_{i,b})^2,
    \label{eq_noobjloss}
\end{equation}
where
\begin{itemize}
    \item $l_{\mathcal{B}}(\boldsymbol{\theta})$ penalizes the difference between the predicted bounding-box ${\hat{\mathcal{B}}_{i,b}=(\hat{x}_{i,b},\hat{y}_{i,b},\hat{w}_{i,b},\hat{h}_{i,b})}$ and their ground truth ${\mathcal{B}_{i,b} = (x_{i,b},y_{i,b},w_{i,b},h_{i,b})}$ for each grid cell. Each grid cell has two mutually exclusive states: i) \textit{containing docking stations}, and ii) \textit{not containing docking stations}. Containing docking stations and not containing docking stations represents if the center of docking stations falls into the $i^{th}$ grid or not, respectively. When the $i^{th}$ grid contains docking stations, $\mathcal{B}_{i,b}$ also has two states. One is \textit{responsible for prediction}, and the other is \textit{not responsible for prediction}. $\mathcal{B}_{i,b}$ is \textit{responsible for prediction} when it has the largest IoU \citep{everingham2010pascal} with the ground truth bounding-box among $B$ bounding-boxes predicted in the $i^{th}$ grid. $l_{i,b}^{dock}$ is equal to $1$ if a) the $i^{th}$ grid contains docking stations, and b) $\mathcal{B}_{i,b}$ is responsible for prediction. Otherwise it is $0$.
    \item $l_{d}(\boldsymbol{\theta})$ penalizes confidence score loss for the grids containing docking stations. $S_{i,b}$ and $\hat{S}_{i,b}$ is the ground truth confidence score and predicted confidence score for the $\mathcal{B}_{i,b}$, respectively. $l_{i,b}^{dock}$ is a indicator function which is equal to $1$ if the $i^{th}$ grid contains docking stations, otherwise it is $0$.
    \item $l_{\bar{d}}(\boldsymbol{\theta})$ penalizes confidence score loss for the grids that do not contain docking stations. $l_{i,b}^{nodock}$ is an indicator function indicating appearance of docking stations in the $i^{th}$ grid. $l_{i,b}^{nodock}$ is equal to $1$ if the $i^{th}$ grid does not contain docking stations.
\end{itemize}
The parameters $\lambda_{\mathcal{B}}$, $\lambda_{d}$ and $\lambda_{\bar{d}}$ are used to control the contribution of different parts of the loss function \eqref{eq_loss}. We experimentally analyze how the configuration of these two parameters affect detection performance in detail in Section~\ref{section:lossanalyze}.


DoNN receives an unseen underwater image as input, and outputs a $G \times G \times B \times 5$ tensor for inference by predicting $B$ bounding-boxes (parameterized by $\mathcal{B}_{i,b}$ using $4$ elements), and $B$ confidence scores (parameterized by $S_{i,b}$ using $1$ element) for each grid. The final prediction is computed by
\begin{equation}
\label{eq_bboxpred}
\mathcal{B}_{pred} = \mathcal{B}_{\hat{i},\hat{b}}, \quad \textrm{where} \quad \hat{i},\hat{b}= \argmaxB_{i,b} \hat{S}_{i,b},
\end{equation}
$\hat{S}_{i,b}$ is the predicted confidence score used in \eqref{eq_objloss}, and $\mathcal{B}_{pred}$ is the final predicted bounding-box of the docking station.

\begin{figure*}
    \centering
    \includegraphics[width = \linewidth]{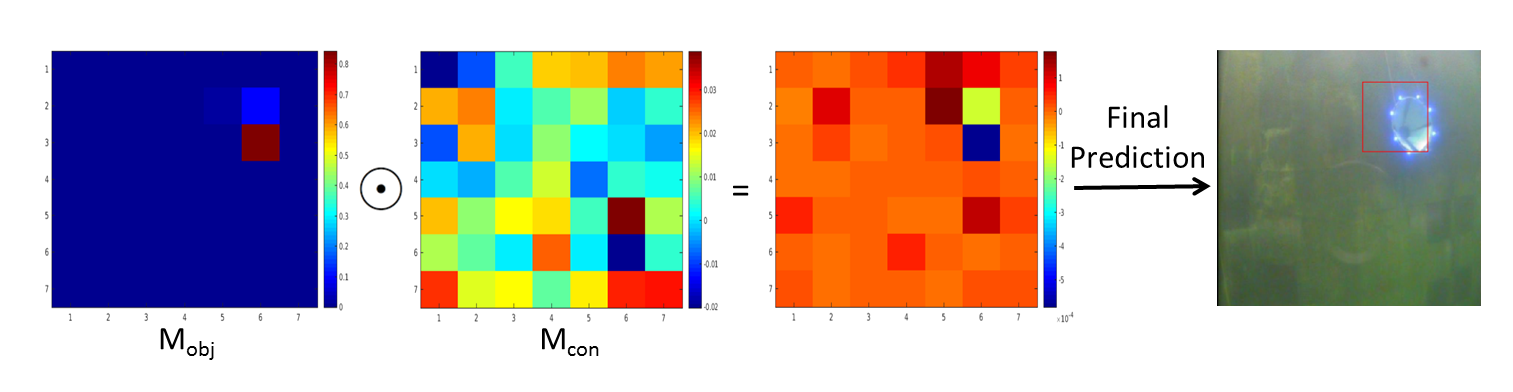}
    \caption{A visualization of $P^i_r(Objectness)$ and $P^i_r(Class_i|Objectness)$ computed using YOLO. We denote  $P^i_r(Objectness)$ and $P^i_r(Class_i|Objectness)$ computed for all grids by $M_{obj}=\{P^i_r(Objectness)| i=1,2,\ldots,G \times G\}$ and $M_{con}=\{P^i_r(Class_i|Objectness)| i=1,2,\ldots,G \times G\}$, respectively. The left two figures show $M_{obj}$ and $M_{con}$, respectively. We use $\odot$ to indicate pixel-wise multiplication of $M_{obj}$ and $M_{con}$. The red bounding-box depicted in the last figure indicates the final prediction provided by YOLO. $M_{obj}$ predicts the location of the docking station correctly, however the final prediction is corrupted after multiplication with $M_{con}$. }
    \label{fig:probability}
\end{figure*}

The major difference between DoNN and YOLO is the loss function. In addition to the loss function \eqref{eq_loss}, \eqref{eq_objloss} and \eqref{eq_noobjloss}, another loss function called \textit{class loss} is used in YOLO. The process of learning in YOLO can be viewed as learning an \textit{objectness} probability and a conditional probability of $Class_c$ for each grid. They are formulated by $P^i_r(Objectness)$ and $P^i_r(Class_c|Objectness)$ for the $i^{th}$ grid, respectively. Then $P^i_r(Objectness)\cdot P^i_r(Class_c|Objectness)$ is used to predict the final class score for each grid. However, in our case, $P^i_r(Class_c|Objectness)$ introduces instability as illustrated in Figure \ref{fig:probability}, due to observation of one target in our task and our relative small datasets. To remedy this problem, we redesign the loss function used in YOLO by estimating only $P_r(dock)$ instead of an objectness probability and a conditional probability. 

\subsection{Pose estimation}

\begin{figure}[h]
    \centering
    \includegraphics[width = \linewidth]{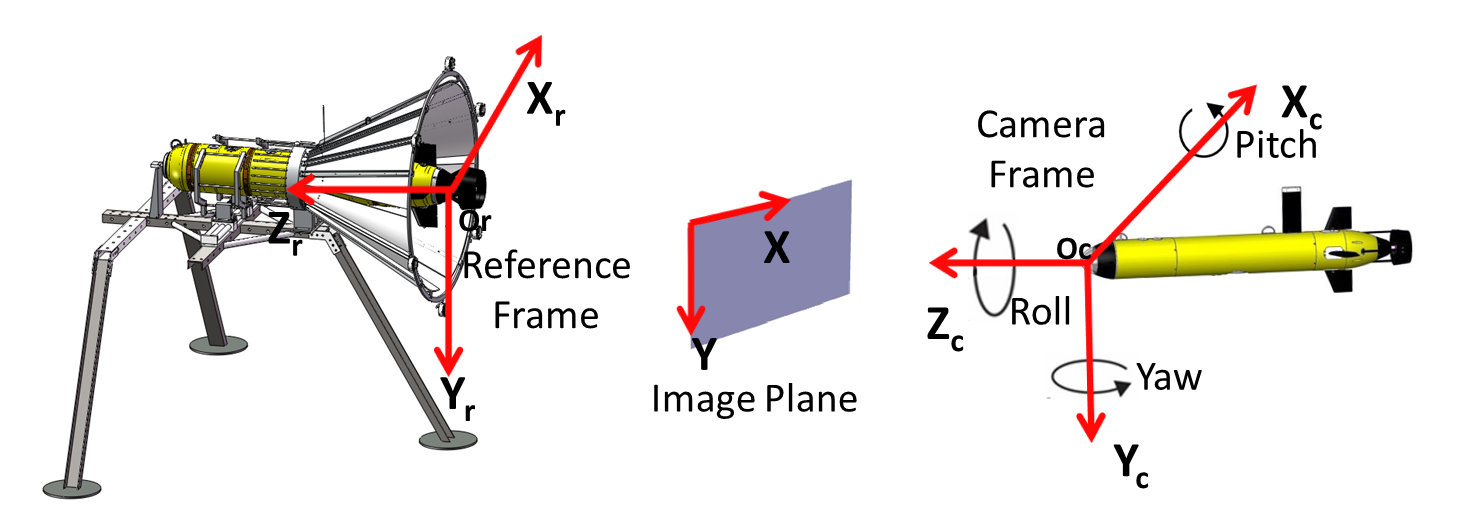}
    \caption{Coordinate frames used in underwater docking. Left: A reference frame. The origin of the reference frame is located at the centroid of the circle formed by landmarks. Middle: An image plane, and an image coordinate frame. Right: A camera frame, and illustrated Euler angles.}
    \label{fig:frames}
\end{figure}

In Section \ref{sec_detection}, we explained how to compute 2D locations $\mathcal{B}_{pred}$ of docking stations in 2D images using the proposed DoNN. In this section, we provide a method which is used to recover the relative 3D position and orientation between docking stations and AUVs from the 2D image patch determined by the estimated 2D location. The 3D position is represented by $X=(x,y,z) \in{\mathbb{R}^3}$, and the orientation is represented by \textit{Euler angles} $(yaw,pitch,roll)$, as illustrated in Figure \ref{fig:frames}. The relative 3D position and orientation between docking stations and AUVs is called \textit{pose}, collectively. The process of pose recovery is called \textit{pose estimation}. Pose estimation requires recovering the pose from the 2D image patch defined by $\mathcal{B}_{pred}$ described in \eqref{eq_bboxpred}. 

Pose estimation is performed if docking stations can be fully observed. We categorize observed docking stations into full observation, and partial observation of docking stations. A full observation of docking stations is a case where all eight landmarks can be observed.  Otherwise, it is called a partial observation. In order to recognize full observations, we segment the image patch determined by $\mathcal{B}_{pred}$ using an adaptive segmentation method proposed by \cite{bradley2007adaptive}. Since other interference, such as noisy luminaries and mirror images, has been addressed during detection, it is needless to worry about the sensitivity of segmentation. After the segmentation, several connected components are available. If the number of connected components is less than eight, then the observation is a partial observation of docking stations. The number of connected components may be greater than eight. which is observed if one light is segmented into more than one connected component. In this case,  we use the k-means algorithm \citep{arthur2007k} to cluster adjacent components and obtain centroids of eight landmarks. At the final step, the centroids are used for pose estimation.

If two coordinate frames are attached to the docking station and the AUV, separately, then we model the geometric relationship (the translation and rotation) between two coordinate frames for pose estimation. In pose estimation, three coordinate frames are used as illustrated in Figure \ref{fig:frames}: i) image coordinate frame, ii) camera coordinate frame and iii) reference coordinate frame. The image coordinate frame is a 2D coordinate system in pixels which is established on the image plane. The camera coordinate and reference coordinate frames are 3D coordinate systems in millimeter. The origin of the camera coordinate resides on the optical center of the camera. We attach a reference coordinate on the center of the circle formed by eight lights. Calculating the pose between AUVs and docking stations is equivalent to determination of the transformation between the camera coordinate frame and reference coordinate frame, since the camera is fixed on the head of AUVs rigidly, and only rigid-body motion is considered.

Next, we explain determination of a transformation between camera and reference coordinate frames. Considering a pinhole camera, the transformation between image coordinate and camera coordinate is computed by 
\begin{equation}
\label{eq_intr}
\begin{bmatrix}
u\\
v\\
1
\end{bmatrix}
=
KX_c
=
\begin{bmatrix}
k_x & k_{\theta} & u_0 & 0\\
0   & k_y        & v_0 & 0\\
0   & 0          & 1   & 0
\end{bmatrix}
\begin{bmatrix}
x_c \\
y_c \\
z_c \\
1
\end{bmatrix},
\end{equation}
where $(u_0,v_0) \in{\mathbb{R}^3}$ is the principal point measured in pixels, $k_{\theta}\in{\mathbb{R}}$ is the skew coefficient, $(u,v)\in{\mathbb{R}^3}$ is the coordinate in the image frame, $(x_c,y_c,z_c)\in{\mathbb{R}^3}$ is the coordinate in the camera frame, $k_x\in{\mathbb{R}}$ and $k_y\in{\mathbb{R}}$ denote the scaling factor converting space metrics to pixel units, and $K\in{\mathbb{R}^{3\times4}}$ is the \textit{intrinsic matrix} of inherent parameters of a camera. It describes the transformation between image frames and camera frames, and can be obtained by camera calibration.
The relationship between the camera frame and reference frame can be described by the extrinsic matrix $E\in{\mathbb{R}^{4\times4}}$ as follows:
\begin{equation}
\label{eq_extr}
\begin{bmatrix}
x_c \\
y_c \\
z_c \\
1
\end{bmatrix}
=
EX_r
=
\begin{bmatrix}
R & T\\
0 & 1
\end{bmatrix}
\begin{bmatrix}
x_r \\
y_r \\
z_r \\
1 
\end{bmatrix}
\end{equation}
where $(x_r,y_r,z_r)\in \mathbb{R}^3$ is the coordinate of the reference coordinate frame, $R \in\mathbb{R}^{3 \times 3}$ is a rotation matrix with constraints $R^TR=RR^T=I$ and $\det(R)=+1$. 
Translation matrix $T \in \mathbb{R}^3$ is computed with respect to the coordinate of the origin point $O_r$ of the reference frame with respect to the camera frame. $O_r$ is the target point used for line tracking mentioned in Section \ref{sec_underwaterdockingalgorithm}.

Computation of $R$ and $T$ using $n$ corresponding points between 2D images points and 3D coordinates is addressed as perspective-n-point (PnP) problems. PnP was first coined by \cite{fischler1981random}, for computation of $R$ and $T$ given a calibrated camera, a set of correspondences between 3D reference points and their 2D images points. At least four correspondences are required to assure computation of a unique solution for $R$ and $T$. Usually two types of methods are used to solve PnP problems: analytical and iterative methods. A popularly used analytical method is Direct Linear Transformation (DLT) \citep{ABDELAZIZ2015103}. DLT calculates $R$ and $T$ by solving 11 entries in linear equations derived by \eqref{eq_intr} and \eqref{eq_extr} from at least six corresponding points. DLT is computationally efficient but suffer from instability in the presence of noise. Iterative methods address PnP problems by minimization of an error criterion, such as re-projection error proposed by \cite{zhang2000flexible}, and collinearity error proposed by \cite{lu2000fast}. Iterative methods are less sensitive to noise, and they are more accurate but  computationally expensive. 

\begin{figure}
    \centering
    \includegraphics[width = \linewidth]{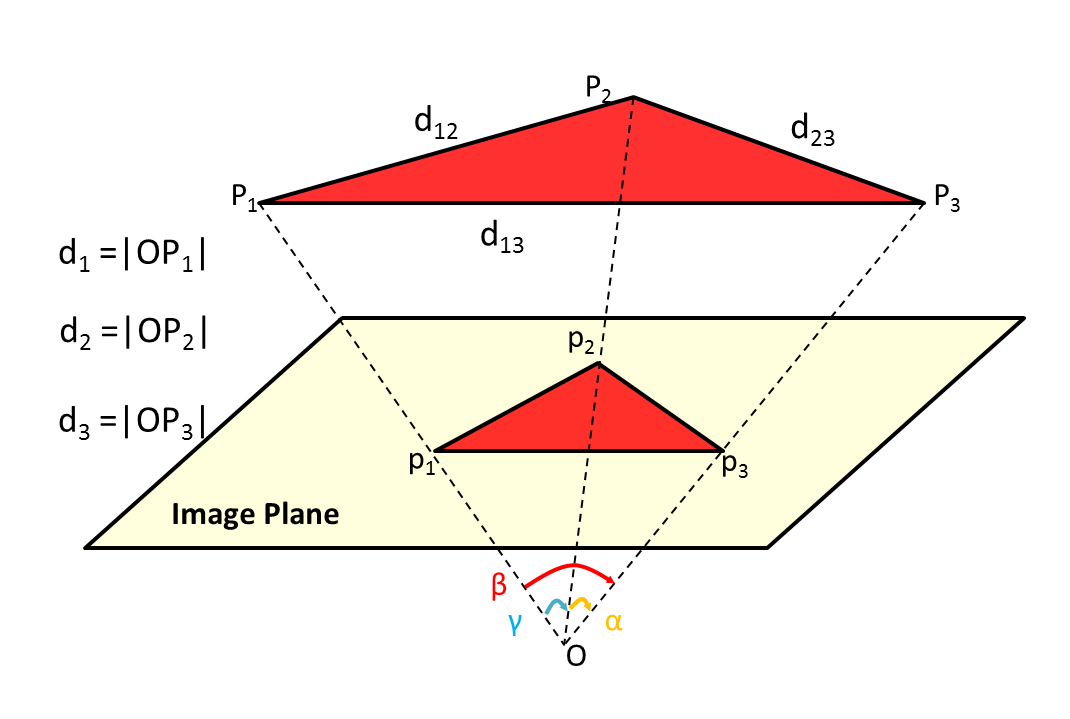}
    \caption{Illustration of the RPnP algorithm. $P_1$,$P_2$ and $P_3$ are three 3D points. $p_1$,$p_2$ and $p_3$ are the corresponding 2D image points located on the image plane. Point $O$ is the optical center of the camera.}
    \label{fig:perspective}
\end{figure}

In our framework, RPnP \citep{li2012robust} is employed to estimate the pose between AUVs and docking stations. In Figure~\ref{fig:perspective}, $P_i$ and $p_i$ denote the $i^{th}$ 3D point and its corresponding image point, respectively. Given $n$ correspondences $P_i \leftrightarrow p_i$, RPnP first split a set of $n$ reference points into $(n-2)$ subsets, each of which contains 3 points. Each subset is illustrated in Figure \ref{fig:perspective}. According to the law of cosines, the following constraints are satisfied for each subset;
\begin{equation}
\label{eq_cosconstrain}
    \begin{array}{ccc}
        d_1^2+d_2^2-2d_1d_2\cos\gamma & = & d_{12}^2 \\
        d_2^2+d_3^2-2d_2d_3\cos\alpha & = & d_{23}^2 \\
        d_1^2+d_3^2-2d_1d_3\cos\beta  & = & d_{13}^2 \\
    \end{array}.
\end{equation}

Then, \eqref{eq_cosconstrain} is converted into a fourth order polynomial for the $j^{th}$ subset by
\begin{equation}
\label{eq_polypnp}
\begin{tabular}{cc}
    $h_j(x) = a_jx^4 + b_jx^3 + c_jx^2 + d_jx + e_j = 0$, \\ $ \forall j=1, 2, \ldots ,n-2$.
\end{tabular}
\end{equation}
Since $(n-2)$ subsets are available, we use $(n-2)$ polynomial equations \eqref{eq_polypnp}, which form a system of nonlinear equations. Rather than solving this nonlinear equation system, RPnP analyzes the local minima of the seventh order polynomial cost function $H$, which is defined by $H=\sum_{j=1}^{n-2}h_j^2(x)$ that has at most 4 minima. The final solution with the least re-projection residual is selected from these minima as the pose estimation result.

The RPnP used in our work has the following properties:
\begin{itemize}
    \item RPnP is accurate and highly efficient. It is a non-iterative solution and achieves as accurate solutions as iterative methods provide with less computational cost. The accuracy of RPnP is 1.5 degree median rotation error, and $ 0.5\%$ median translation error for $n=8$ co-planar points with Gaussian noise $\mathcal{N}(\mu =0 ,\sigma^{2}=9)$ as reported by \cite{li2012robust} in simulations. RPnP consumes less than 1 ms if $n=8$.
    \item RPnP is stable in the co-planar case. Many PnP solutions \citep{oberkampf1996iterative,schweighofer2006robust} suffer from pose ambiguity which results in highly unstable results. Pose ambiguity refers to the fact that orientation cannot be determined uniquely \citep{tanaka2014solution}. Allowing a co-planar arrangement of landmarks can reduce complexity for designing docking stations. \cite{li2012robust} showed that the mean error of rotation and translation converges when number of correspondences is larger than eight. Therefore, eight landmarks are designed in our docking station. 
    \item The computational complexity of RPnP is $O(n)$. Its computational time grows linearly with the number of correspondences. It offers flexibility for increase and decrease of the number of landmarks according to practical requirements. Therefore, we can re-configure the amount of landmarks without substantial increase of the computational cost. 
\end{itemize}

\section{Experimental analyses}
In the experimental analyses, we first analyze robustness and credibility of the proposed DoNN using our datasets. Then, we analyze the accuracy and efficiency of the pose estimation method by ground  and underwater docking experiments. Finally, we give experimental analyses of our integrated underwater docking algorithm incorporating the aforementioned detection and pose estimation methods. Implementation details of the algorithms are given in the supplemental material. 


\subsection{Analysis of detection performance}
\label{sec:performevaluation}

In this section, we first introduce our proposed dataset UDID, and performance measures used to evaluate detection performance. Then, we analyze convergence properties of DoNN for the proposed loss functions. Finally, we compare and analyze detection performance of DoNN, and state-of-the-art YOLO and FasterRCNN methods using the UDID and its deformed variations.

\subsubsection{Our proposed UDID dataset}
In order to evaluate the performance of DoNN, we first set up an underwater docking images dataset (UDID), which is collected in our experimental water pool. The experimental water pool is 15m long, 10m wide and 9m deep with the docking station fixed at 2m deep underwater. It comprises a training set $D_{ltr}$ and a test set $D_{lv}$ as illustrated in Table \ref{table:datasets}. We call images containing docking stations \textit{foreground}, and images not containing docking stations \textit{background}.

As mentioned in Section \ref{l_intro}, real-world underwater images suffer from (1) blurring, (2) color shift, (3) reduced contrast, (4) mirror images, (5) non-uniform illumination and (6) noisy luminaries \citep{bryson2015true,park2009experiments}. Therefore, we also deform the test set $D_{lv}$ using various deformation methods to assess the performance of DoNN in simulated dynamic underwater environments. All images are resized to $448 \times 448$ for training and testing of DoNN, and the images are resized to the original size along with detected bounding-box for pose estimation. 
\begin{table}[!h]
\centering
\caption{Our proposed dataset UDID. The training subset $D_{ltr}$ contains $8252$ foreground images and no background images. The test subset $D_{lv}$ consists of $1128$ foreground images and $1114$ background images.}
\begin{tabular}{p{1cm}|p{1.5cm}|p{1.55cm}|p{1.5cm}|}
\cline{2-4}
& {Foreground Images} & { Background Images }& {Total Images} \\
\cline{1-4}
\multicolumn{1}{|c|}{$D_{ltr}$} & 8252 & 0 & 8252 \\
\hline
\multicolumn{1}{|c|}{$D_{lv}$} & 1128 & 1114 & 2242 \\
\hline
\end{tabular}
\label{table:datasets}
\end{table}

As mentioned in Section \ref{sec:donn}, there are two types of CNN based detection methods. One is region proposal based detection and the other is proposal free detection. In the experimental analyses, we compare our proposed DoNN with Faster-RCNN and YOLO, which are the sate-of-art region proposal based and proposal free detection methods, respectively.
For a comparison, we used the same architecture for design of YOLO and DoNN before employment of the fully connective layers. A Faster-RCNN which employs a ZF network \citep{zeiler2014visualizing} is employed for comparison. Fully connective layers of the Faster-RCNN are updated for our two-class classification for end-to-end training using $D_{ltr}$.

\subsubsection{Performance measures used for evaluation of detection algorithms}

Detection performance is evaluated using receiver operating characteristic (ROC) curve and its area under curve (AUC) \citep{bradley1997use}. ROC is a plot of true positive rate (TPR) against false positive rate (FPR) computed at various threshold settings. TPR and FPR are defined by
\begin{equation}
    TPR = \frac{TP}{TP+FN} \quad {\rm and} \quad FPR = \frac{FP}{FP+TN},
    \label{eq:tpr}
\end{equation}
where $TP$, $FP$, $TN$, $FN$ is the number of true positive, false positive, true negative and false negative samples, respectively. $TP$, $FP$, $TN$ and $FN$ are illustrated by a confusion matrix in Table \ref{table:confusionmatrix}. If a prediction and a true value of a sample is docking station, then the prediction result is evaluated as a true positive (TP). If both the prediction and the true value are non-docking stations, then the prediction is evaluated as a true negative (TN). If the predicted value is a docking station while the true value is non-docking station, then the prediction is evaluated as false positive (FP). False positive means that a docking station is detected when a docking station is not actually there. If the predicted value is non-docking station while the true value is a docking station, then the prediction is false negative (FN). We label the true value of bounding-boxes whose IoU with the ground truth bounding-box exceed $50\%$ as docking stations as suggested by \cite{everingham2010pascal}. Otherwise, they are labeled as non-docking stations.

\begin{table}[!h]
\centering
\caption{Confusion matrix of predicted and true values.}
\begin{tabular}{l|m{2cm}|p{1.3cm}|c|}
\multicolumn{2}{c}{}&\multicolumn{2}{c}{True Values}\\
\cline{3-4}
\multicolumn{2}{c|}{}&\multicolumn{1}{p{1.3cm}|}{Docking Stations}&\multicolumn{1}{p{2cm}|}{Non-docking Stations}\\
\cline{2-4}
\multirow{2}{*}{\shortstack{Predicted \\ Values}}&\multicolumn{1}{m{1.5cm}|}{Docking  Stations} & \multicolumn{1}{c|}{$TP$} & $FP$ \\
\cline{2-4}
& Non-docking stations & \multicolumn{1}{c|}{$FN$} & $TN$ \\
\cline{2-4}
\end{tabular} 
\label{table:confusionmatrix}
\end{table}

The area under the ROC curve is called AUC for short, which can give an insight into the general performance of detection algorithms. AUC is equal to 1 if the maximum performance is achieved. In the following sections, we analyze the performance of DoNN, YOLO and Faster-RCNN in aforementioned test set $D_{lv}$ and its various deformed versions.

\begin{table*}
\centering
\caption{AUC performance of DoNN obtained using different $\lambda_{d}$, $\lambda_{\mathcal{B}}$ and $\lambda_{\bar{d}}$ on $D_{lv}$. $B$ and $G$ are set to $2$ and $7$ in these experiments, respectively. DoNN achieves the best AUC performance for $\lambda_{d}=0.5$, $\lambda_{\bar{d}} = 0.1$, $\lambda_{\mathcal{B}}=3$.}
\subfloat[][Results for $\lambda_d = 0.1$.]{
\begin{tabular}{c|c|c|c|c|c}
\hline
\diagbox {$\lambda_{\bar{d}}$}{AUC}{$\lambda_{\mathcal{B}}$} & 0.1 & 0.5 & 1 & 3 &5 \\
\hline
0.1 & 0.99494 & 0.99856 & 0.99862 & 0.99697 & 0.99564 \\
\hline
0.5 & 0.96462 & 0.99689 & 0.99435 & 0.99410 & 0.99297 \\
\hline
1 & 0.92583 & 0.99509 & 0.99348 & 0.99134 & 0.99067 \\
\hline
3 & 0.87398 & 0.99708 & 0.99166 & 0.98235 & 0.99042 \\
\hline
5 & 0.87772 & 0.99492 & 0.99311 & 0.99137 & 0.99231 \\
\hline
\end{tabular}
}\\
\subfloat[][Results for $\lambda_d = 0.5$.]{
\begin{tabular}{c|c|c|c|c|c}
\hline
\diagbox {$\lambda_{\bar{d}}$}{AUC}{$\lambda_{\mathcal{B}}$} & 0.1 & 0.5 & 1 & 3 &5 \\
\hline
0.1 & 0.99863 & 0.99918 & 0.99878 & \cellcolor{gray}0.99964 & 0.99902 \\
\hline
0.5 & 0.99376 & 0.99847 & 0.99923 & 0.99800 & 0.99847 \\
\hline
1 & 0.98665 & 0.99705 & 0.99822 & 0.99827 & 0.99579 \\
\hline
3 & 0.95976 & 0.99383 & 0.99730 & 0.99806 & 0.99674 \\
\hline
5 & 0.95500 & 0.99367 & 0.98950 & 0.99605 & 0.99648 \\
\hline
\end{tabular}
}\\
\subfloat[][Results for $\lambda_d = 1$.]{
\begin{tabular}{c|c|c|c|c|c}
\hline
\diagbox{$\lambda_{\bar{d}}$}{AUC}{$\lambda_{\mathcal{B}}$} & 0.1 & 0.5 & 1 & 3 &5 \\
\hline
0.1 & 0.99900 & 0.99928 & 0.99930 & 0.99958 & 0.99958 \\
\hline
0.5 & 0.99557 & 0.99907 & 0.99917 & 0.99931 & 0.99916 \\
\hline
1 & 0.98988 & 0.99893 & 0.99892 & 0.99861 & 0.99787 \\
\hline
3 & 0.96917 & 0.99721 & 0.99777 & 0.99789 & 0.99835 \\
\hline
5 & 0.95736 & 0.99383 & 0.99751 & 0.99150 & 0.99697 \\
\hline
\end{tabular}

}\\

\subfloat[][Results for $\lambda_d = 3$.]{
\begin{tabular}{c|c|c|c|c|c}
\hline
\diagbox {$\lambda_{\bar{d}}$}{AUC}{$\lambda_{\mathcal{B}}$} & 0.1 & 0.5 & 1 & 3 &5 \\
\hline
0.1 & 0.99370 & 0.99719 & 0.99907 & 0.99951 & 0.99953 \\
\hline
0.5 & 0.98607 & 0.99888 & 0.99904 & 0.99945 & 0.99895 \\
\hline
1 & 0.98597 & 0.99752 & 0.99905 & 0.99935 & 0.99952 \\
\hline
3 & 0.96551 & 0.99153 & 0.99388 & 0.99526 & 0.99807 \\
\hline
5 & 0.95920 & 0.99428 & 0.99507 & 0.99577 & 0.99434 \\
\hline
\end{tabular}
}\\
\subfloat[][Results for $\lambda_d = 5$.]{
\begin{tabular}{c|c|c|c|c|c}
\hline
\diagbox {$\lambda_{\bar{d}}$}{AUC}{$\lambda_{\mathcal{B}}$} & 0.1 & 0.5 & 1 & 3 &5 \\
\hline
0.1 & 0.98886 & 0.99951 & 0.99920 & 0.99895 & 0.99931 \\
\hline
0.5 & 0.98908 & 0.99577 & 0.99906 & 0.99924 & 0.99961 \\
\hline
1 & 0.98509 & 0.99340 & 0.99916 & 0.99897 & 0.99928 \\
\hline
3 & 0.97131 & 0.99635 & 0.99313 & 0.99463 & 0.99908 \\
\hline
5 & 0.96330 & 0.99158 & 0.99239 & 0.98866 & 0.99797 \\
\hline
\end{tabular}
}
\label{table:hyperparameters}
\end{table*}

\subsubsection{Analysis of hyperparameters of loss functions of DoNN}
\label{section:lossanalyze}

Our proposed DoNN was trained using the loss function given in \eqref{eq_loss}. The loss function contains three parts: i) $l_{\mathcal{B}}(\boldsymbol{\theta})$, ii) $l_{d}(\boldsymbol{\theta})$ and iii) $l_{\bar{d}}(\boldsymbol{\theta})$. Five hyperparameters $\lambda_{\mathcal{B}}$, $\lambda_{\bar{d}}$, $\lambda_{\bar{d}}$, $B$ and $G$ are used during the training phase. 

\begin{figure}
    \hfill
    \includegraphics[width = \linewidth]{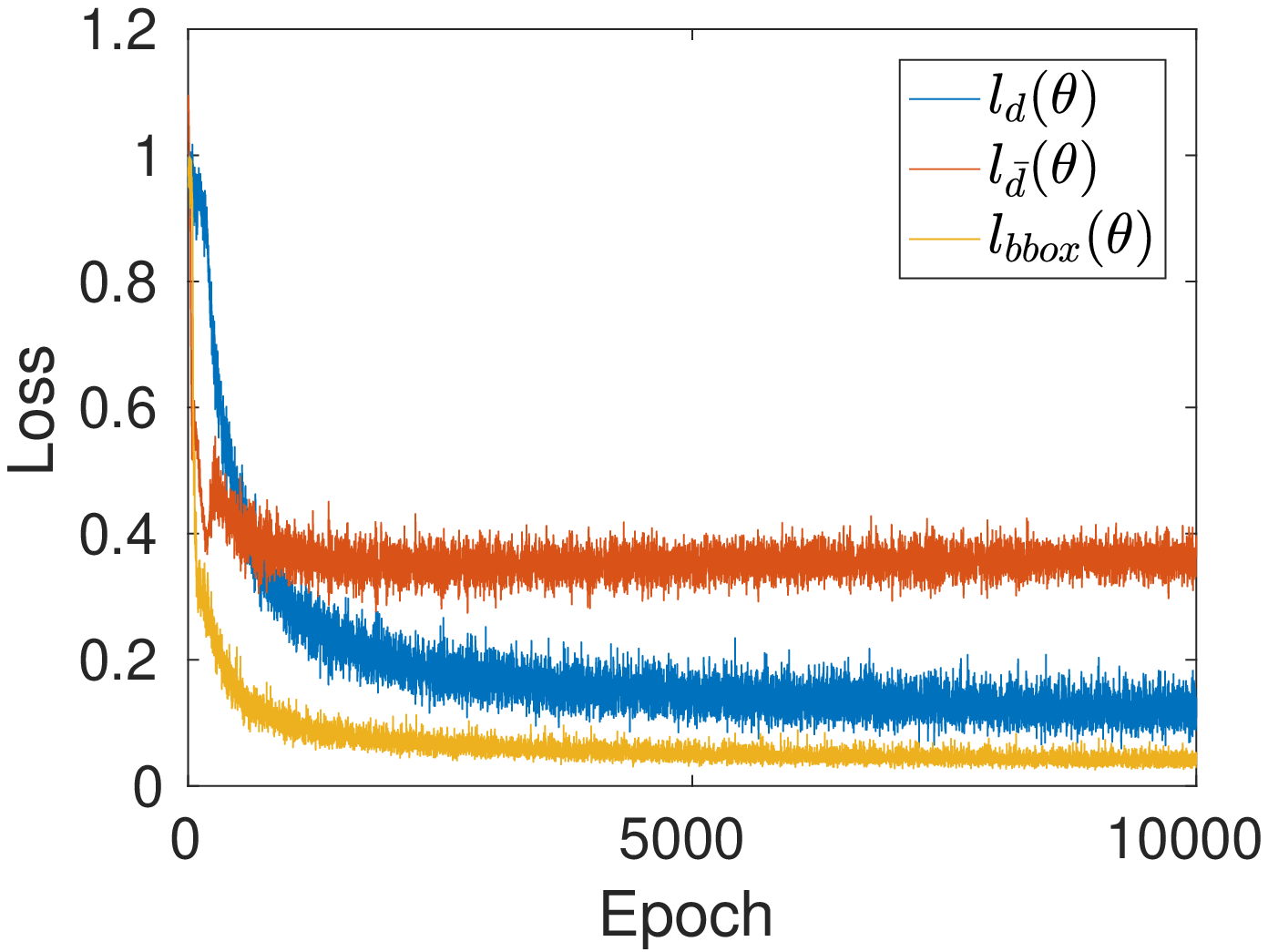}
    \caption{Change of value of loss functions $l_{\mathcal{B}}(\boldsymbol{\theta})$, $l_{d}(\boldsymbol{\theta})$ and $l_{\bar{d}}(\boldsymbol{\theta})$ during training phase.}
    \label{fig:threeloss}
\end{figure}

$\lambda_{\mathcal{B}}$, $\lambda_{\bar{d}}$ and $\lambda_{d}$ are employed to balance their contribution, since three parts of the loss function \eqref{eq_loss} take values from different scales, as shown in Figure \ref{fig:threeloss}. They can also be viewed as three weights acting on three parts of the loss function. We show AUC performance of different settings of these three hyperparameters in Table \ref{table:hyperparameters}. In average, DoNN achieves \textit{acceptable} AUC performance in most of the hyperparameter settings. DoNN provides better performance when the parameters weight $l_{\mathcal{B}}(\boldsymbol{\theta})$ more than $l_{d}(\boldsymbol{\theta})$ and $l_{\bar{d}}(\boldsymbol{\theta})$. DoNN also performs better when less weight is given to $l_{\bar{d}}(\boldsymbol{\theta})$ with $\lambda_{\mathcal{B}}$ and $\lambda_{d}$ fixed. This is observed when the number of grids not containing docking stations is larger than the number of grids containing docking stations. The parameter $l_{\bar{d}}(\boldsymbol{\theta})$ dominates the value of the loss function if equal weights are given to three parts of the loss function. DoNN achieves the best performance for $\lambda_{d}=0.5$, $\lambda_{\bar{d}} = 0.1, \lambda_{\mathcal{B}}=3$. Therefore, we use this setting in the following experiments.

The hyperparameter $B$ controls the number of scored bounding-boxes that are predicted by the DoNN for each grid. The larger $B$ is, the more parameters and computation cost are required. We show that the AUC performance of DoNN at the setting of $\lambda_{d}=0.5$, $\lambda_{\bar{d}} = 0.1$ and $\lambda_{\mathcal{B}}=3$ in Table \ref{table:hyperB}. The AUC performance of DoNN does not benefit a lot from a large $B$. Since the DoNN achieves the best performance for $B=2$, $B$ is set to $2$ in the following experiments.

The hyperparameter $G$ controls the number of grids into which an image is divided. Fine grained grids increase the number of parameters and computational cost. We perform experiments at $G=\{2,4,7,14\}$, since $G$ must be a factor of the width and height of the input image, which are both $448$ in our experiments. The experimental results are shown in Table \ref{table:hyperG}. The best AUC performance is achieved at $G=7$, and $G$ is set to $7$ in the experiments.

\begin{table}
    \centering
    \caption{AUC performance of DoNN obtained using different settings of $B$ on $D_{lv}$. We use $\lambda_{d}=0.5$, $\lambda_{\bar{d}} = 0.1$, $\lambda_{\mathcal{B}}=3$, $G=7$ for experiments given in this table. Best performance is obtained using $B=2$.}
    \begin{tabular}{p{0.5cm} c c c c c}
        \hline
        & $B=1$ & \cellcolor{gray}$B=2$ & $B=3$ & $B=5$ &$B=10$ \\
        \hline
        AUC & 0.99944 & \cellcolor{gray}0.99964 & 0.99962 & 0.99953 & 0.99915 \\
        \hline
    \end{tabular}
    \label{table:hyperB}
\end{table}

\begin{table}
    \centering
    \caption{AUC performance of DoNN obtained using different settings of $G$ on $D_{lv}$. We use $\lambda_{d}=0.5$, $\lambda_{\bar{d}} = 0.1$, $\lambda_{\mathcal{B}}=3$, $B=2$ in experiments of this table. Best performance is obtained using $G=7$.}
    \begin{tabular}{c c c c c}
        \hline
        & $G=2$ & $G=4$ & \cellcolor{gray}$G=7$ &$G=14$ \\
        \hline
        AUC & 0.99900 & 0.99819 & \cellcolor{gray}0.99964 & 0.99890 \\
        \hline
    \end{tabular}
    \label{table:hyperG}
\end{table}

\subsubsection{Comparison of performance of detection algorithms using the original test dataset $D_{lv}$}
\label{sec:detectperformoriginal}

\begin{figure}
    \centering
    \includegraphics[width = 1.05\linewidth]{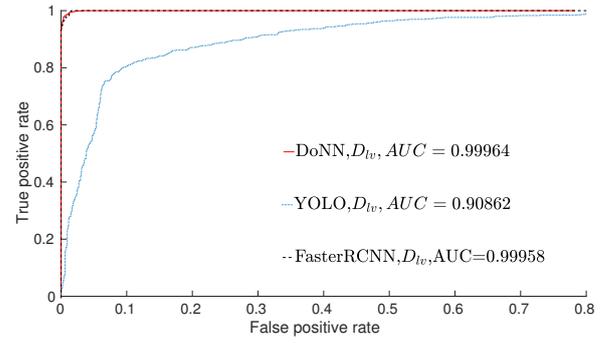}
    \caption{ROC curves and the associated AUCs of DoNN, YOLO, and Faster-RCNN computed using $D_{lv}$.}
    \label{fig:ROCwithouttrans}
\end{figure}

We first analyze detection performance of YOLO, FasterRCNN and DoNN using the original test dataset $D_{lv}$. Figure~\ref{fig:ROCwithouttrans} shows the ROC curve and the associated AUC of three models. The results show that DoNN performs slightly better than Faster-RCNN, and they both outperform YOLO. The AUC of DoNN and Faster-RCNN are $0.99964$ and $0.99958$, respectively, achieving good performance on $D_{lv}$. We will show that Faster-RCNN is not as robust to various deformations of $D_{lv}$ as DoNN in the following analyses.

\begin{figure}
    \centering
    \captionsetup[subfigure]{justification=centering}
    \subfloat[][Input image.]{
        \centering
        \includegraphics[width= 0.3\linewidth,height = 0.179\linewidth]{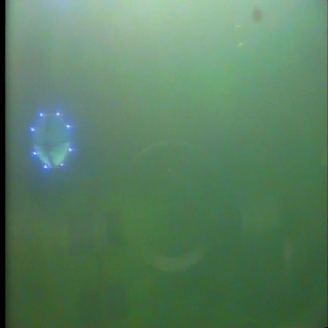}
        }
    \subfloat[][$\mathcal{T}_1$.]{
        \centering
        \includegraphics[width=0.3\linewidth,height = 0.179\linewidth]{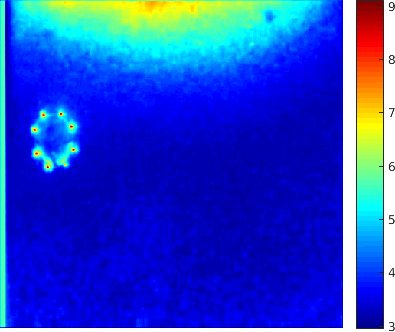}
        }
    \subfloat[][$\mathcal{T}_2$.]{
        \centering
        \includegraphics[width=0.3\linewidth,height = 0.179\linewidth]{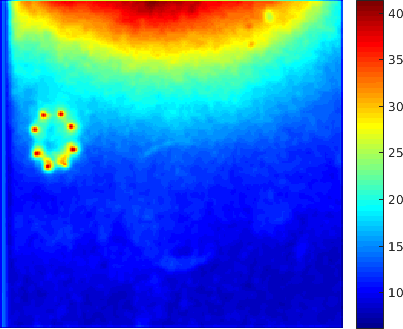}
        }
\\
    \subfloat[][$\mathcal{T}_3$.]{
        \centering
        \includegraphics[width=0.3\linewidth,height = 0.179\linewidth]{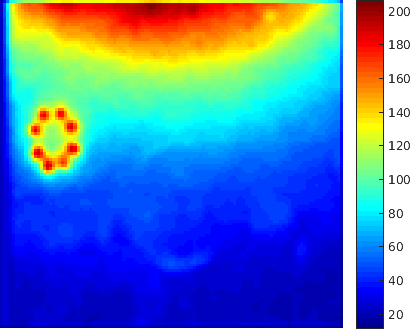}

    }
    \subfloat[][$\mathcal{T}_4$.]{
        \centering
        \includegraphics[width= 0.3\linewidth,height = 0.179\linewidth]{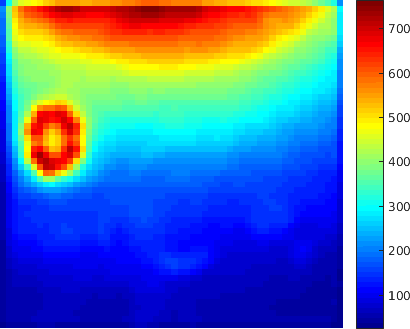}
        }
    \subfloat[][$\mathcal{T}_5$.]{
        \centering
        \includegraphics[width=0.3\linewidth,height = 0.179\linewidth]{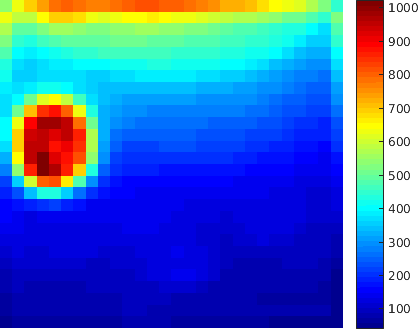}
        }
        \\
    \subfloat[][$\mathcal{T}_6$.]{
        \centering
        \includegraphics[width=0.3\linewidth,height = 0.179\linewidth]{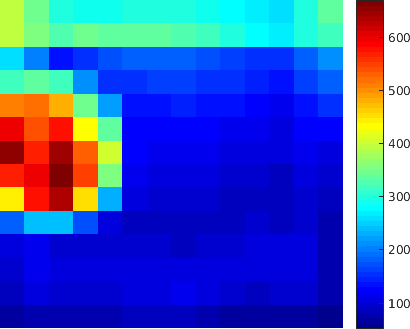}
        }
    \subfloat[][$\mathcal{T}_7$.]{
        \centering
        \includegraphics[width=0.3\linewidth,height = 0.179\linewidth]{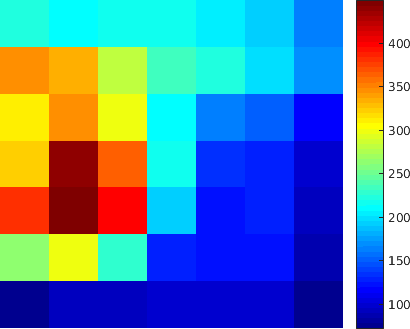}

    }
    \subfloat[][$\mathcal{T}_8$.]{
        \centering
        \includegraphics[width= 0.3\linewidth,height = 0.179\linewidth]{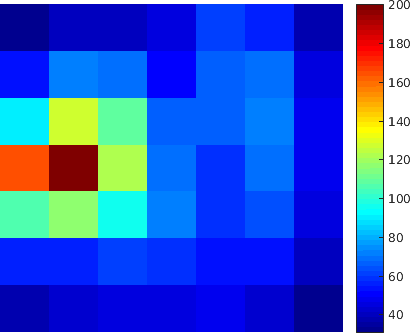}
        }
        \\
    \subfloat[][$\mathcal{T}_9$.]{
        \centering
        \includegraphics[width=0.30\linewidth,height = 0.179\linewidth]{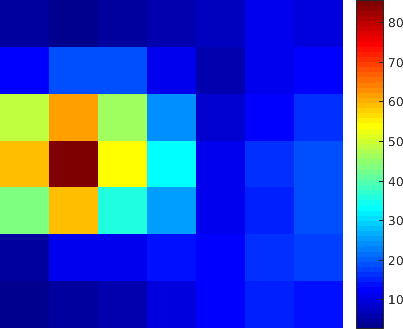}
        }
    \subfloat[][$S$.]{
        \centering
        \includegraphics[width=0.3\linewidth,height = 0.179\linewidth]{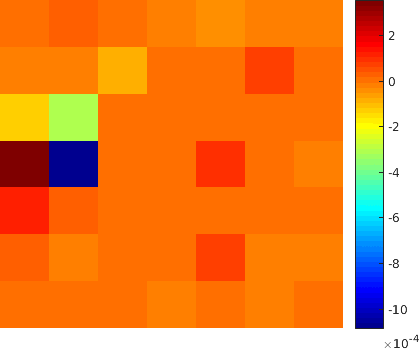}
        }
    \subfloat[][Detection result.]{
        \centering
        \includegraphics[width=0.3\linewidth,height = 0.179\linewidth]{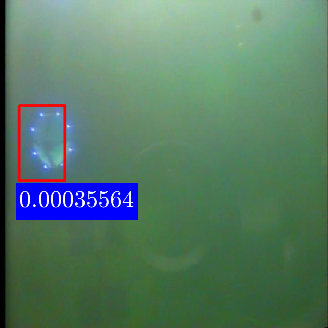}

    }
    \hfill\\
    \caption{Feature maps computed for a false detection predicted by \textit{YOLO} and its detection result on $D_{lv}$. Figures given from left to right, and top to bottom correspond to its input image, feature maps $\mathcal{T}_i$ computed at the $i^{th}$ layer $Conv_i,i = 1,2\ldots,9$, the confidence map and the final detection result.}
    \label{fig:featmapyolowithouttrans}
\end{figure}

\begin{figure}
    \centering
    \captionsetup[subfigure]{justification=centering}
    \subfloat[][Input image.]{
        \centering
        \includegraphics[width= 0.3\linewidth,height = 0.179\linewidth]{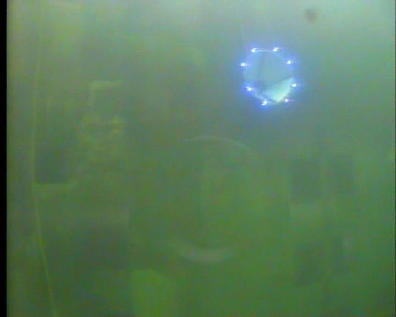}
        }
    \subfloat[][$\mathcal{T}_1$.]{
        \centering
        \label{fig:featmapfrrcnnwithouttrans_t1}
        \includegraphics[width=0.3\linewidth,height = 0.179\linewidth]{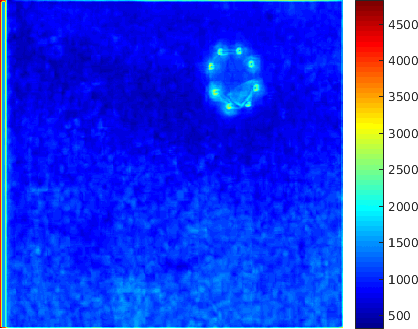}
        }
    \subfloat[][$\mathcal{T}_2$.]{
        \centering
        \includegraphics[width=0.3\linewidth,height = 0.179\linewidth]{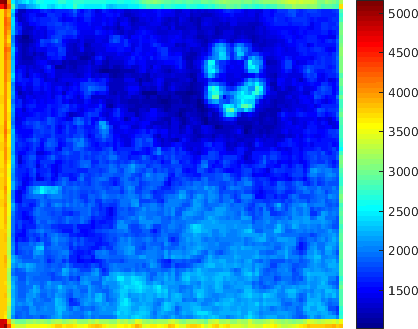}
        }
\\
    \subfloat[][$\mathcal{T}_3$.]{
        \centering
        \includegraphics[width=0.3\linewidth,height = 0.179\linewidth]{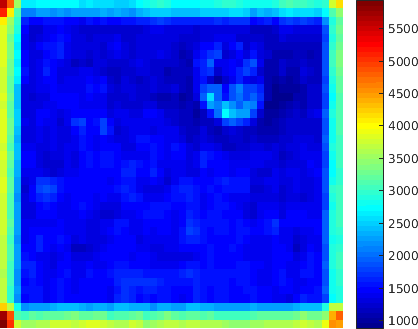}

    }
    \subfloat[][$\mathcal{T}_4$.]{
        \centering
        \includegraphics[width= 0.3\linewidth,height = 0.179\linewidth]{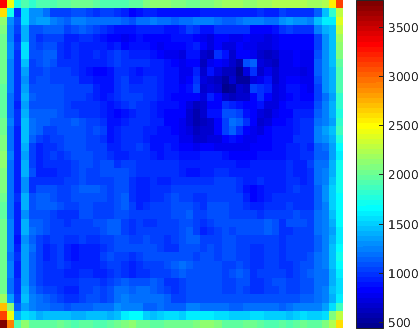}
        }
    \subfloat[][$\mathcal{T}_5$.]{
        \centering
        \includegraphics[width=0.3\linewidth,height = 0.179\linewidth]{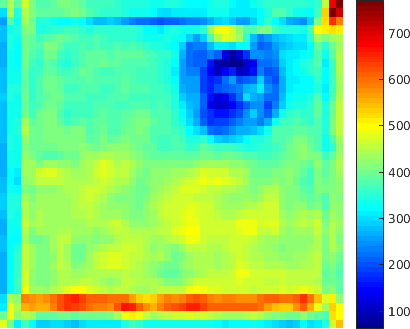}
        }
\\
    \subfloat[][Detection result.]{
        \centering
        \includegraphics[width=0.3\linewidth,height = 0.179\linewidth]{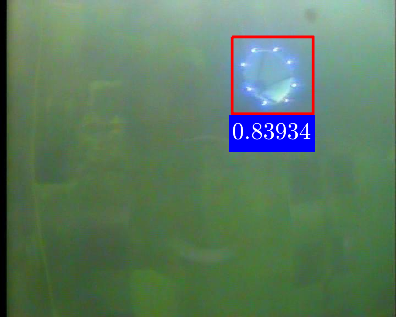}

    }
    \hfill\\
    \caption{Feature maps computed using Faster-RCNN, and its detection result on $D_{lv}$. Figures given from left to right, and top to bottom correspond to its input image, feature maps $\mathcal{T}_i$ computed at the $i^{th}$ layer $Conv_i,i = 1,2\ldots,9$, and the final detection result.}
    \label{fig:featmapfrrcnnwithouttrans}
\end{figure}

\begin{figure}
    \centering
    \captionsetup[subfigure]{justification=centering}
    \subfloat[][Input image.]{
        \centering
        \includegraphics[width= 0.3\linewidth,height = 0.179\linewidth]{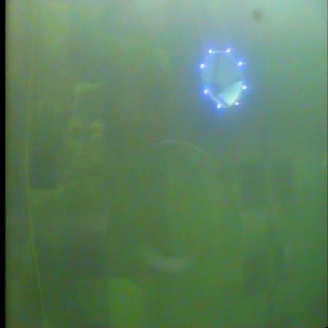}
        }
    \subfloat[][$\mathcal{T}_1$.]{
        \centering
        \includegraphics[width=0.3\linewidth,height = 0.179\linewidth]{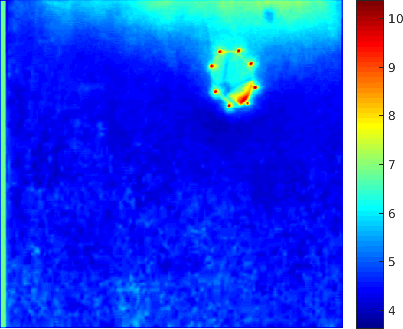}
        }
    \subfloat[][$\mathcal{T}_2$.]{
        \centering
        \includegraphics[width=0.3\linewidth,height = 0.179\linewidth]{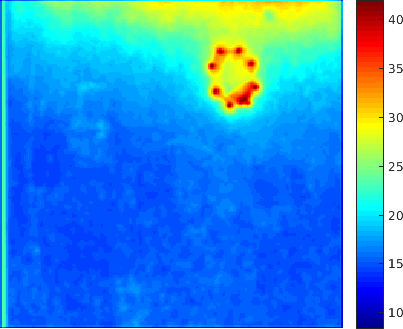}
        }
 \\
    \subfloat[][$\mathcal{T}_3$.]{
        \centering
        \includegraphics[width=0.3\linewidth,height = 0.179\linewidth]{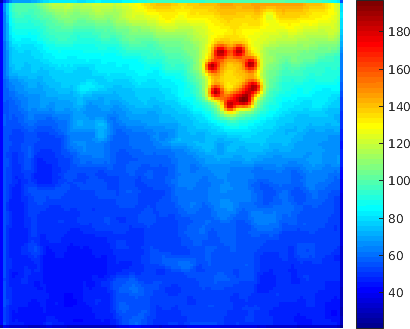}

    }
    \subfloat[][$\mathcal{T}_4$.]{
        \centering
        \includegraphics[width= 0.3\linewidth,height = 0.179\linewidth]{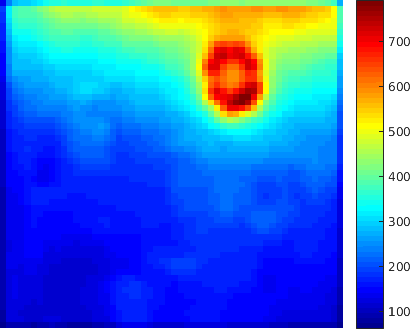}
        }
    \subfloat[][$\mathcal{T}_5$.]{
        \centering
        \label{fig:featmapdonnwithouttrans_t5}
        \includegraphics[width=0.3\linewidth,height = 0.179\linewidth]{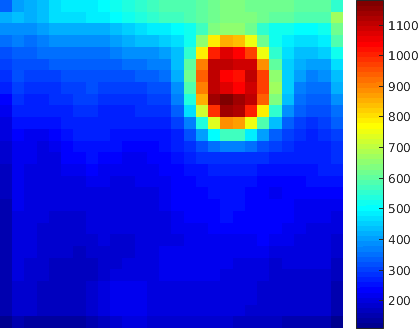}
        }
 \\
    \subfloat[][$\mathcal{T}_6$.]{
        \centering
        \includegraphics[width=0.3\linewidth,height = 0.179\linewidth]{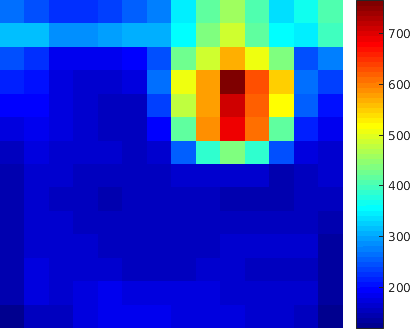}
        }
    \subfloat[][$\mathcal{T}_7$.]{
        \centering
        \includegraphics[width=0.3\linewidth,height = 0.179\linewidth]{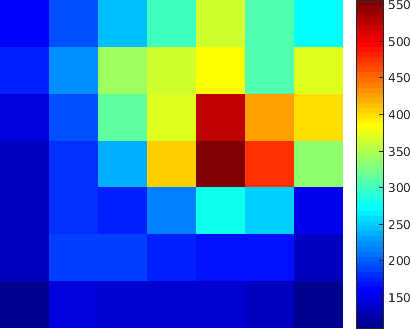}

    }
    \subfloat[][$\mathcal{T}_8$.]{
        \centering
        \includegraphics[width= 0.3\linewidth,height = 0.179\linewidth]{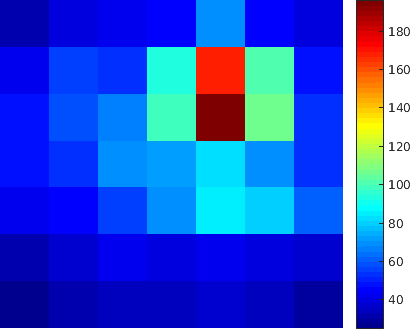}
        }
        \\
    \subfloat[][$\mathcal{T}_9$.]{
        \centering
        \includegraphics[width=0.3\linewidth,height = 0.179\linewidth]{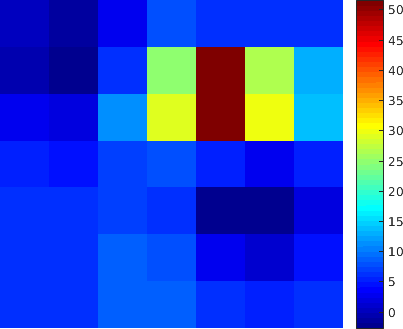}
        }
    \subfloat[][$S$.]{
        \centering
        \includegraphics[width=0.3\linewidth,height = 0.179\linewidth]{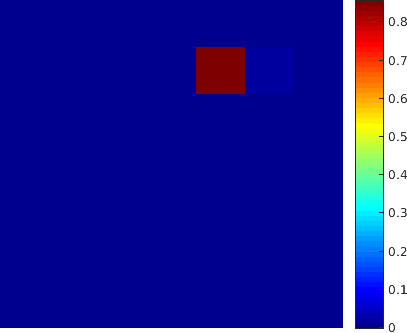}
        }
    \subfloat[][Detection result.]{
        \centering
        \includegraphics[width=0.3\linewidth,height = 0.179\linewidth]{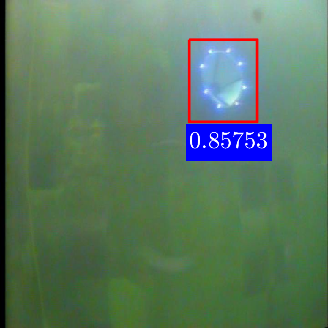}

    }
    \hfill\\
    \caption{Feature maps computed using DoNN, and its detection result on $D_{lv}$. Figures given from left to right, and top to bottom correspond to its input image, feature maps $\mathcal{T}_i$ computed at the $i^{th}$ layer $Conv_i,i = 1,2\ldots,9$, the confidence map, and the final detection result.}
    \label{fig:featmapdonnwithouttrans}
\end{figure}

As a concrete example, we depict feature maps $\mathcal{T}_i$ computed at the $i^{th}$ convolution layer, and the detection result of typical samples for three models in Figure~\ref{fig:featmapyolowithouttrans}, \ref{fig:featmapfrrcnnwithouttrans} and \ref{fig:featmapdonnwithouttrans}. A feature map $\mathcal{T}_i$ is computed by  
\begin{equation}
    \mathcal{T}_i = \sum_{j=1}^{N_i}{\tau_{i,j}(x,y)},
    \label{eq:featmapT}
\end{equation}
where $\tau_{i,j}(x,y)$ denotes the $j^{th}$ feature map of $N_i$ feature maps computed using \eqref{eq:featmaptau} at the $i^{th}$ convolutional layer. After each pooling, $\mathcal{T}_i$ is down-sampled by a factor which is determined by the architecture of the network. 
Each $\mathcal{T}_i$ is coupled with a color-bar which indicates its corresponding color scale. For DoNN and YOLO, an additional confidence map $S = \{ \bar{S}_i \}_{i=1} ^{G^2}$, where $\bar{S}_i =\max(\hat{S}_{i,1},\hat{S}_{i,2},\ldots,\hat{S}_{i,B})$ and $|S|=G^2 $, is depicted in the last but one figure. Each pixel belonging to the set $S$ indicates the predicted confidence score for its corresponding grid. 

Comparing a map $\mathcal{T}_i$ computed using these three models, we conjecture that all three models can be used to learn feature representations of spatial structural patterns of docking stations, which are invariant to change of light to a different extent. In Figure \ref{fig:featmapfrrcnnwithouttrans_t1}, we depict a map $\mathcal{T}_1$ computed using FasterRCNN. We observe that features are activated only in a region containing docking station, although the upper ambient light is stronger than the lower. As for DoNN, effect of ambient light is gradually eliminated as shown in Figure \ref{fig:featmapdonnwithouttrans}. We depicted a map $\mathcal{T}_5$ computed using DoNN in Figure \ref{fig:featmapdonnwithouttrans_t5}. The results show that features are activated only in a region that contains a docking station. Therefore, DoNN is robust to light variance. As a result, the grid corresponding to the docking station obtained the highest confidence score while others obtained almost zero score. Figure \ref{fig:featmapyolowithouttrans} shows a false prediction provided by YOLO. IoU of the prediction is low since estimated conditional class distributions corrupt the final confidence map $S$, as illustrated in Figure \ref{fig:probability}.

Since AUC performance of YOLO is worse compared to FasterRCNN and DoNN, we further analyzed the performance of FasterRCNN and DoNN in the following sections.

\subsubsection{Comparison of performance of detection algorithms using blurred images} 
Blurred underwater images are observed due to scattering of light \citep{bryson2015true}. We analyze change of performance of the detection methods using blurred underwater images in a controlled setting for different blurring patterns. Therefore, we generate blurred images from the UDID by employing Gaussian filters \citep{shapiro2001computer}
\begin{equation}
    I(x,y) = \frac {1}{\sqrt{2\pi}\sigma}e^{-\frac{d^2}{2\sigma^2}}
\end{equation}
with varying standard deviation $\sigma$ to simulate blurring in underwater images, where $d=\sqrt{(x-x_c)^2+(y-y_c)^2}$ is the distance between a pixel $(x,y)$ and a filter center pixel $(x_c,y_c)$. The filter size is set to $2 \times \lceil 2 \times \sigma \rceil +1$. We sample the deviation $\sigma$ uniformly by $\sigma \in [1,10]$. The larger values the $\sigma$ takes, the more blurred images are obtained. The dataset obtained after employment of blurring is called by $D_{bl \sigma}$.

\begin{figure}
    \centering
    \includegraphics[width = 1.0\linewidth, height = 5cm]{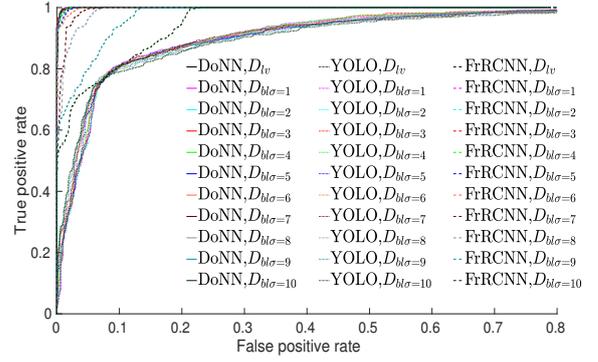}
    \caption{ROC curves of DoNN, YOLO, Faster-RCNN computed for various degrees of blurring. The corresponding AUCs are given in Table \ref{tbl:AUCblur}.}
    \label{fig:ROCblur}
\end{figure}

Table \ref{tbl:blurresfrrcnnanddonn} shows sample blurred images and the corresponding detection results predicted using FasterRCNN and DoNN, respectively. The predicted probability values decrease as $\sigma$ increases. This indicates that both FasterRCNN and DoNN provide less confident results in their prediction by the increase of blurring. FasterRCNN suffers from blurring more than DoNN as observed in Figure~\ref{fig:ROCblur} and Table~\ref{tbl:AUCblur}. The AUC of FasterRCNN degrades from $ 0.99958$ to $ 0.95785$ as $\sigma$ varies from $1$ to $10$ while that of DoNN degrades from $0.99964$ to $0.99948$. DoNN outperforms FasterRCNN in nine out of ten levels of blurring. Fine-grained details of docking stations are lost by the increase of blurring, but by preserving spatial structural patterns of docking stations. We conjecture that DoNN outperforms  FasterRCNN in blurred underwater images since DoNN can be used to learn \textit{better} feature representations of docking stations compared to FasterRCNN. DoNN can still keep a high activation in its feature maps although blurring is very high, as shown in Table \ref{tbl:blurfeatdonn}. However, activation of FasterRCNN in feature maps becomes almost as weak as background in cases of $\sigma=8$ and $\sigma=10$, resulting in incorrect detection, as shown in Table \ref{tbl:blurfeatfrrcnn}.

\begin{table}
\caption{Sample blurred images, and detection results provided by FasterRCNN and DoNN on the set of blurred images.}
\label{tbl:blurresfrrcnnanddonn}
    \begin{tabular}{p{0.6cm} m{.38\linewidth} m{.38\linewidth}}
        \hline
        \multicolumn{1}{c}{$\sigma$} & \multicolumn{1}{c}{FrRCNN} & \multicolumn{1}{c}{DoNN} \\ 
        \hline
        \\[-2ex]
        $1$ &
        \parbox[c]{\linewidth}{\includegraphics[width=\linewidth, height = 0.81 \linewidth]{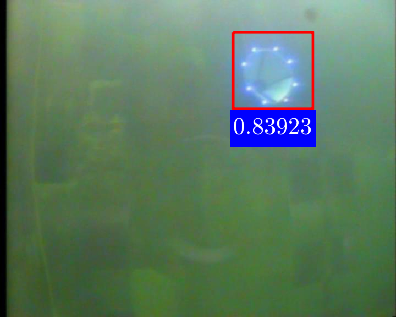}} &
        \parbox[c]{\linewidth}{\includegraphics[width=\linewidth, height = 0.81 \linewidth]{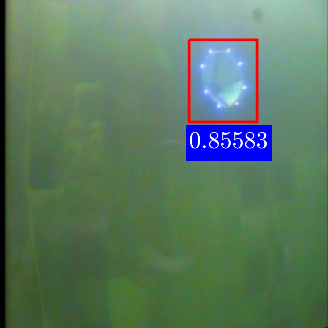}} \\
        \\[-2ex] 
        \hline
        \\[-2ex]
        $2$ &
        \parbox[c]{\linewidth}{\includegraphics[width=\linewidth, height = 0.81 \linewidth]{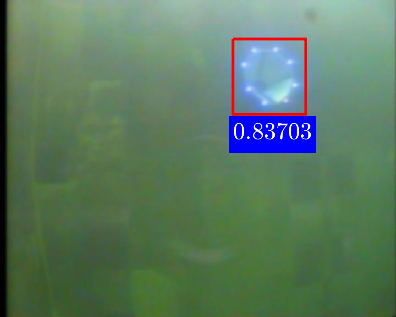}} &
        \parbox[c]{\linewidth}{\includegraphics[width=\linewidth, height = 0.81 \linewidth]{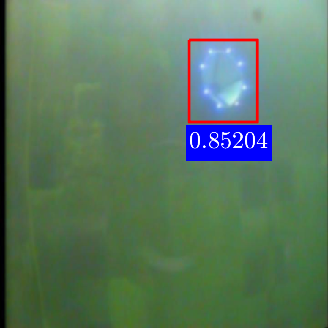}} \\
        \\[-2ex] 
        \hline
        \\[-2ex]
        $4$ &
        \parbox[c]{\linewidth}{\includegraphics[width=\linewidth, height = 0.81 \linewidth]{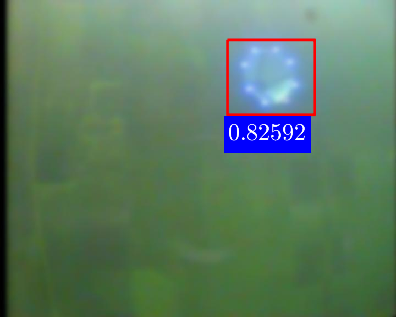}} &
        \parbox[c]{\linewidth}{\includegraphics[width=\linewidth, height = 0.81 \linewidth]{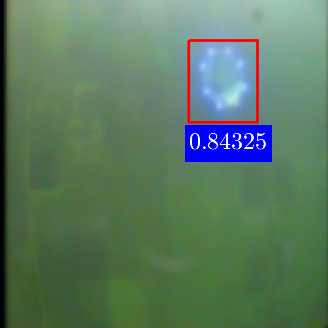}} \\
        \\[-2ex] 
        \hline
        \\[-2ex]
        $6$ &
        \parbox[c]{\linewidth}{\includegraphics[width=\linewidth, height = 0.81 \linewidth]{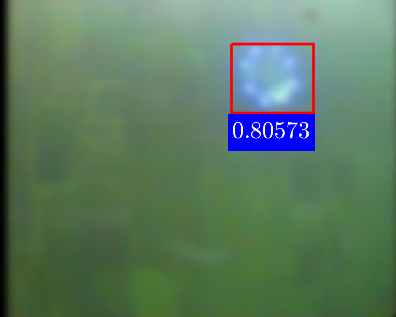}} &
        \parbox[c]{\linewidth}{\includegraphics[width=\linewidth, height = 0.81 \linewidth]{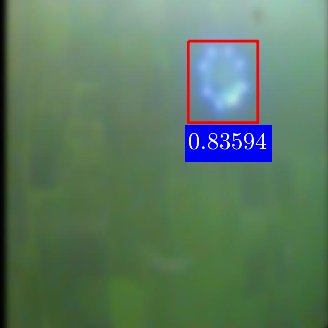}} \\
        \\[-2ex] 
        \hline
        \\[-2ex]
        $8$ &
        \parbox[c]{\linewidth}{\includegraphics[width=\linewidth, height = 0.81 \linewidth]{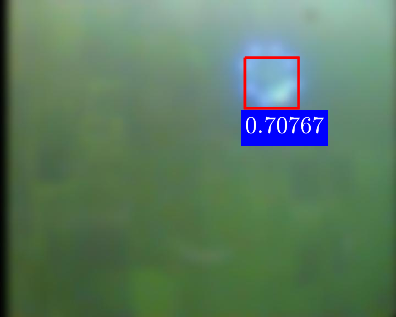}} &
        \parbox[c]{\linewidth}{\includegraphics[width=\linewidth, height = 0.81 \linewidth]{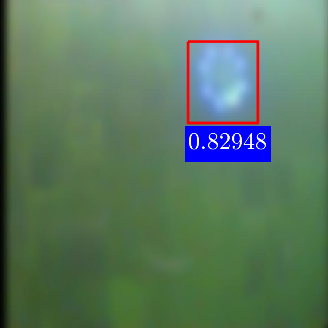}} \\
        \\[-2ex] 
        \hline
        \\[-2ex]
        $10$ &
        \parbox[c]{\linewidth}{\includegraphics[width=\linewidth, height = 0.81 \linewidth]{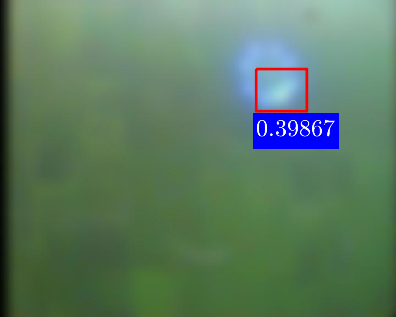}} &
        \parbox[c]{\linewidth}{\includegraphics[width=\linewidth, height = 0.81 \linewidth]{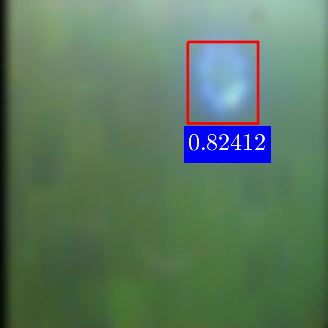}} \\
        \\[-2ex] 
        \hline
    \end{tabular}
\end{table}

\begin{table}
\caption{Feature maps $\mathcal{T}_i$ computed at the $i^{th}$ layer $Conv_i,i = 1,2,5$, and a detection result of FasterRCNN for $\sigma = 8$ and $\sigma = 10$.}
\label{tbl:blurfeatfrrcnn}
    \begin{tabular}{p{0.6cm} m{.38\linewidth} m{.38\linewidth}}
        \hline
        Item & \multicolumn{1}{c}{$\sigma = 8$} & \multicolumn{1}{c}{$\sigma = 10$} \\ 
        \hline
        \\[-2ex]
        $\mathcal{T}_1$ &
        \parbox[c]{\linewidth}{\includegraphics[width=\linewidth, height = 0.81 \linewidth]{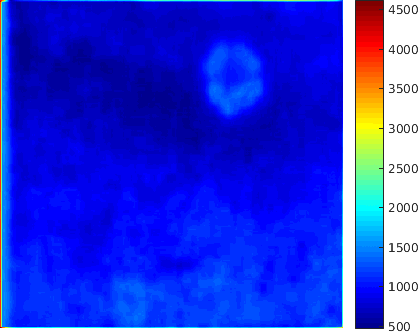}} &
        \parbox[c]{\linewidth}{\includegraphics[width=\linewidth, height = 0.81 \linewidth]{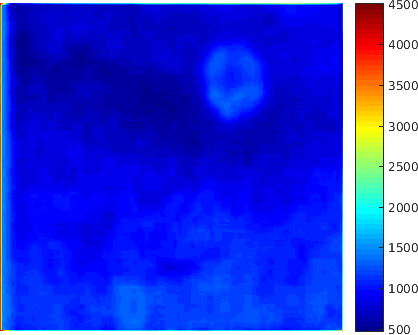}} \\
        \\[-2ex] 
        \hline
        \\[-2ex]
        $\mathcal{T}_2$ &
        \parbox[c]{\linewidth}{\includegraphics[width=\linewidth, height = 0.81 \linewidth]{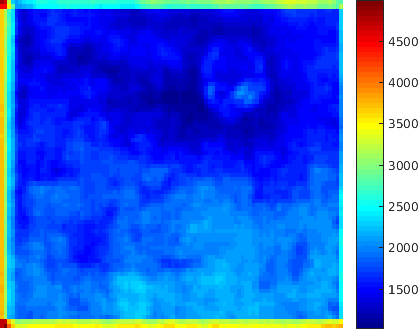}} &
        \parbox[c]{\linewidth}{\includegraphics[width=\linewidth, height = 0.81 \linewidth]{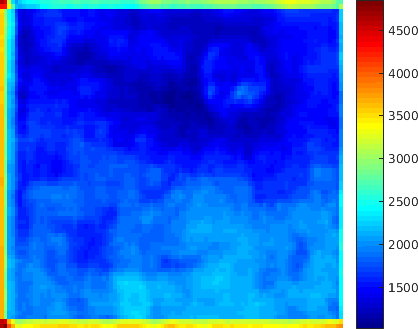}} \\
        \\[-2ex] 
        \hline
        \\[-2ex]
        $\mathcal{T}_5$ &
        \parbox[c]{\linewidth}{\includegraphics[width=\linewidth, height = 0.81 \linewidth]{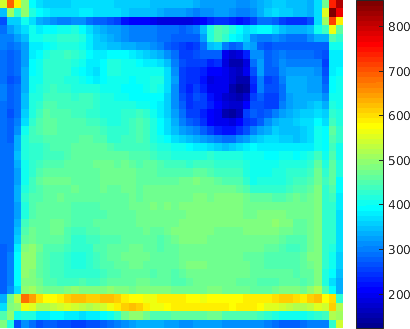}} &
        \parbox[c]{\linewidth}{\includegraphics[width=\linewidth, height = 0.81 \linewidth]{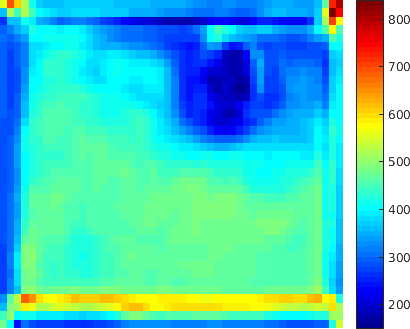}} \\
        \\[-2ex] 
        \hline
        \\[-2ex]
        Det. Result &
        \parbox[c]{\linewidth}{\includegraphics[width=\linewidth, height = 0.81 \linewidth]{result/feat/frrcnn/blur8/detect}} &
        \parbox[c]{\linewidth}{\includegraphics[width=\linewidth, height = 0.81 \linewidth]{result/feat/frrcnn/blur10/detect}} \\
        \\[-2ex] 
        \hline
    \end{tabular}
\end{table}

\begin{table}
\caption{Feature maps $\mathcal{T}_i$ computed at the $i^{th}$ layer $Conv_i,i = 1,5$, confidence map, and a detection result of DoNN for $\sigma = 8$ and $\sigma = 10$.}
\label{tbl:blurfeatdonn}
    \begin{tabular}{p{0.6cm} m{.38\linewidth} m{.38\linewidth}}
        \hline
        Item & \multicolumn{1}{c}{$\sigma = 8$} & \multicolumn{1}{c}{$\sigma = 10$} \\ 
        \hline
        \\[-2ex]
        $\mathcal{T}_1$ &
        \parbox[c]{\linewidth}{\includegraphics[width=\linewidth, height = 0.81 \linewidth]{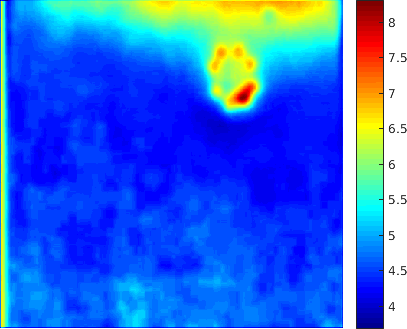}} &
        \parbox[c]{\linewidth}{\includegraphics[width=\linewidth, height = 0.81 \linewidth]{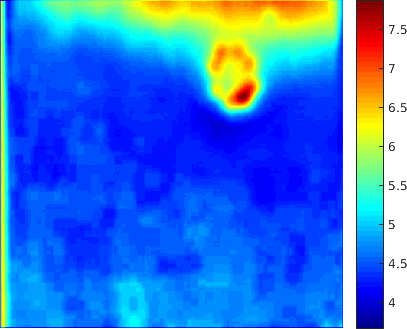}} \\
        \\[-2ex] 
        \hline
        \\[-2ex]
        $\mathcal{T}_5$ &
        \parbox[c]{\linewidth}{\includegraphics[width=\linewidth, height = 0.81 \linewidth]{result/feat/frrcnn/blur8/conv5}} &
        \parbox[c]{\linewidth}{\includegraphics[width=\linewidth, height = 0.81 \linewidth]{result/feat/frrcnn/blur10/conv5}} \\
        \\[-2ex] 
        \hline
        \\[-2ex] 
        $S$ &
        \parbox[c]{\linewidth}{\includegraphics[width=\linewidth, height = 0.81 \linewidth]{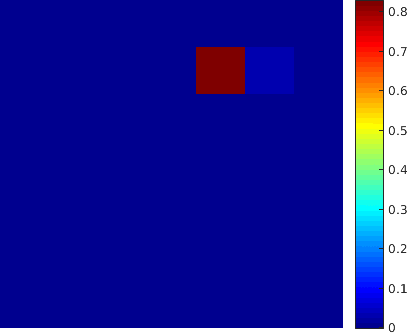}} &
        \parbox[c]{\linewidth}{\includegraphics[width=\linewidth, height = 0.81 \linewidth]{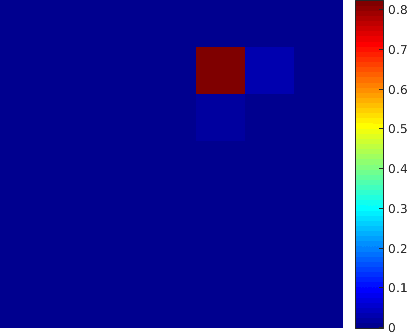}} \\
        \\[-2ex] 
        \hline
        \\[-2ex]
        Det. Result &
        \parbox[c]{\linewidth}{\includegraphics[width=\linewidth, height = 0.81 \linewidth]{result/feat/donn/blur8/detect}} &
        \parbox[c]{\linewidth}{\includegraphics[width=\linewidth, height = 0.81 \linewidth]{result/feat/donn/blur10/detect}} \\
        \\[-2ex] 
        \hline
    \end{tabular}
\end{table}

\begin{table}
\caption{AUC of DoNN, YOLO, Faster-RCNN computed for various degrees of blurring, and the corresponding ROC curves are given in Figure~\ref{fig:ROCblur}.}
\label{tbl:AUCblur}
    \begin{tabular}{c|c|c|c}
    \hline
    \diagbox {$\sigma$}{AUC}{Model} & DoNN & YOLO & FrRCNN \\
    \hline
    $D_{lv}$ & 0.99964&0.90862&0.99958 \\
    \hline
    1 & 0.99964&0.91411&0.99959 \\
    \hline
    2 & 0.99966&0.91025&0.99943 \\
    \hline
    3 & 0.99964&0.91446&0.99971 \\
    \hline
    4 & 0.99965&0.91465&0.99942 \\
    \hline
    5 & 0.99967&0.91147&0.99899 \\
    \hline
    6 & 0.99967&0.91453&0.99854 \\
    \hline
    7 & 0.99953&0.91484&0.99493 \\
    \hline
    8 & 0.99957&0.91549&0.99210 \\
    \hline
    9 & 0.99958&0.91333&0.97791 \\
    \hline
    10& 0.99948&0.90734&0.95785 \\
    \hline
    Average & 0.99961 & 0.91305 & 0.99185\\
    \hline
    \end{tabular}
\end{table}

\subsubsection{Comparison of performance of detection algorithms under color shift} 

Color shift is determined by various factors in underwater environments, such as attenuation. In order to compare performance of three models in underwater images with color shift, underwater images with color shift at different rates of hue, saturation and value shift are created by
\begin{equation}
    L_{out_p} = \lambda_p L_{in_p},  p \in {\{H,S,V\}},
\end{equation}
where $L_{in_p}$ denotes one of HSV components of images in $D_{lv}$. Datasets after hue, saturation and value shift are indicated by $D_{\lambda_h}$, $D_{\lambda_s}$ and $D_{\lambda_v}$ respectively.

\begin{figure}
    \centering
    \captionsetup[subfigure]{justification=centering}
    \subfloat[][Distribution of $\bar{\lambda}_h$]{
        \centering
        \includegraphics[width=0.47\linewidth]{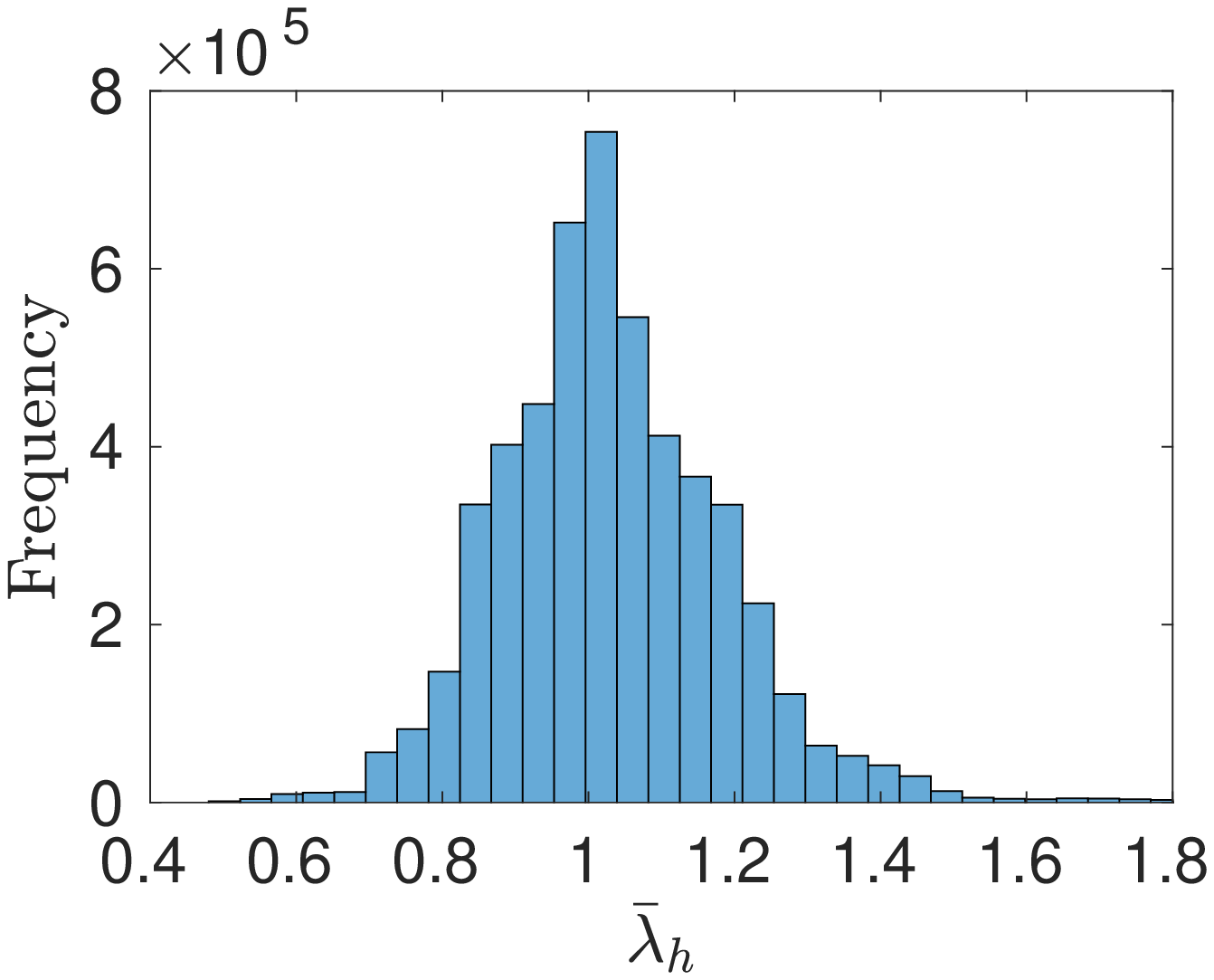}
        \label{fig:hlambdadist}
        }
    \hfill
        \subfloat[][Distribution of $\bar{\lambda}_s$]{
        \centering
        \includegraphics[width=0.47\linewidth]{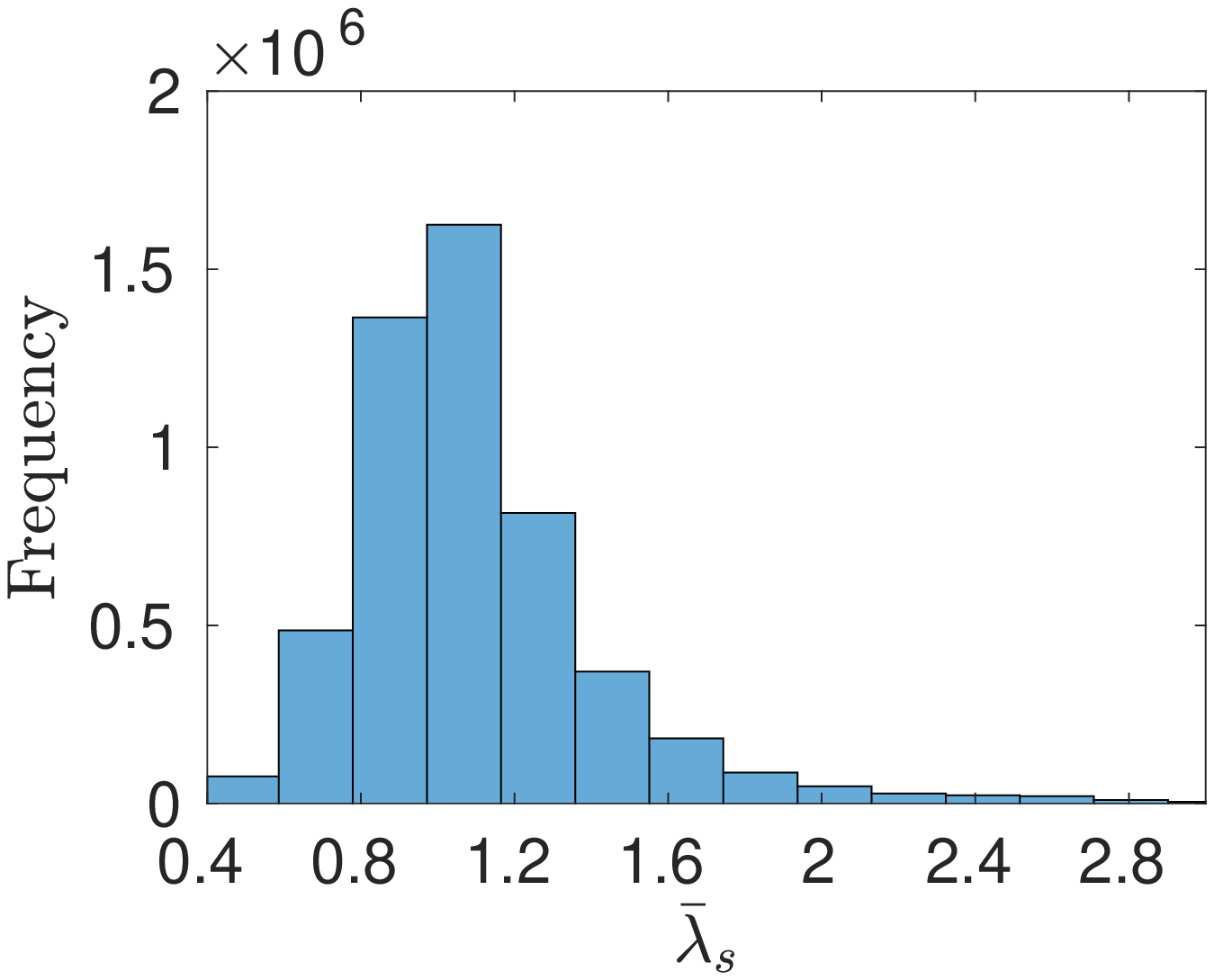}
        \label{fig:slambdadist}
        }
    \hfill
    \\
    \subfloat[][Distribution of $\bar{\lambda}_v$]{
        \centering
        \includegraphics[width=0.47\linewidth]{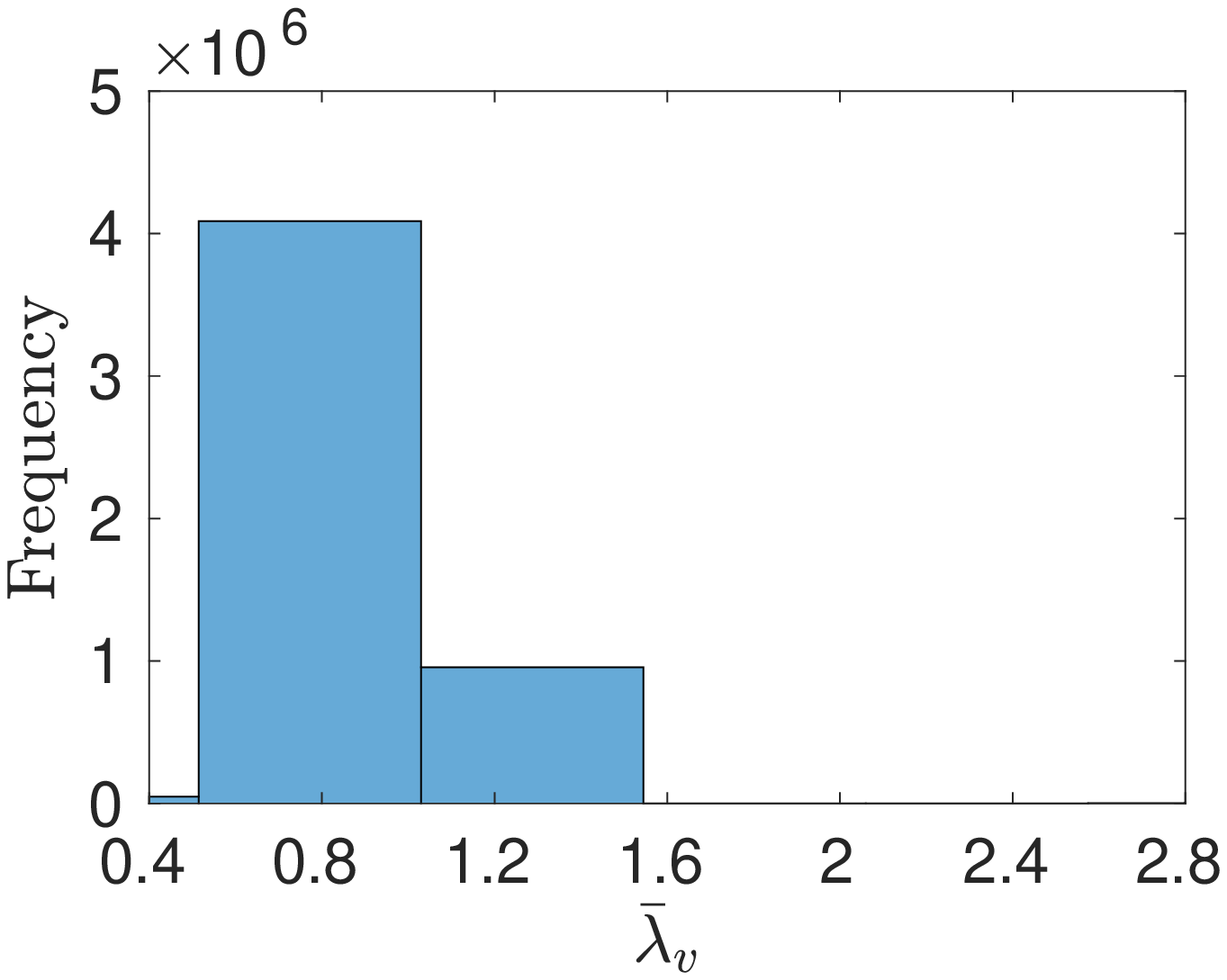}
        \label{fig:vlambdadist}
        }
    \hfill
        \subfloat[][Distribution of $\bar{\gamma}$]{
        \centering
        \includegraphics[width=0.47\linewidth]{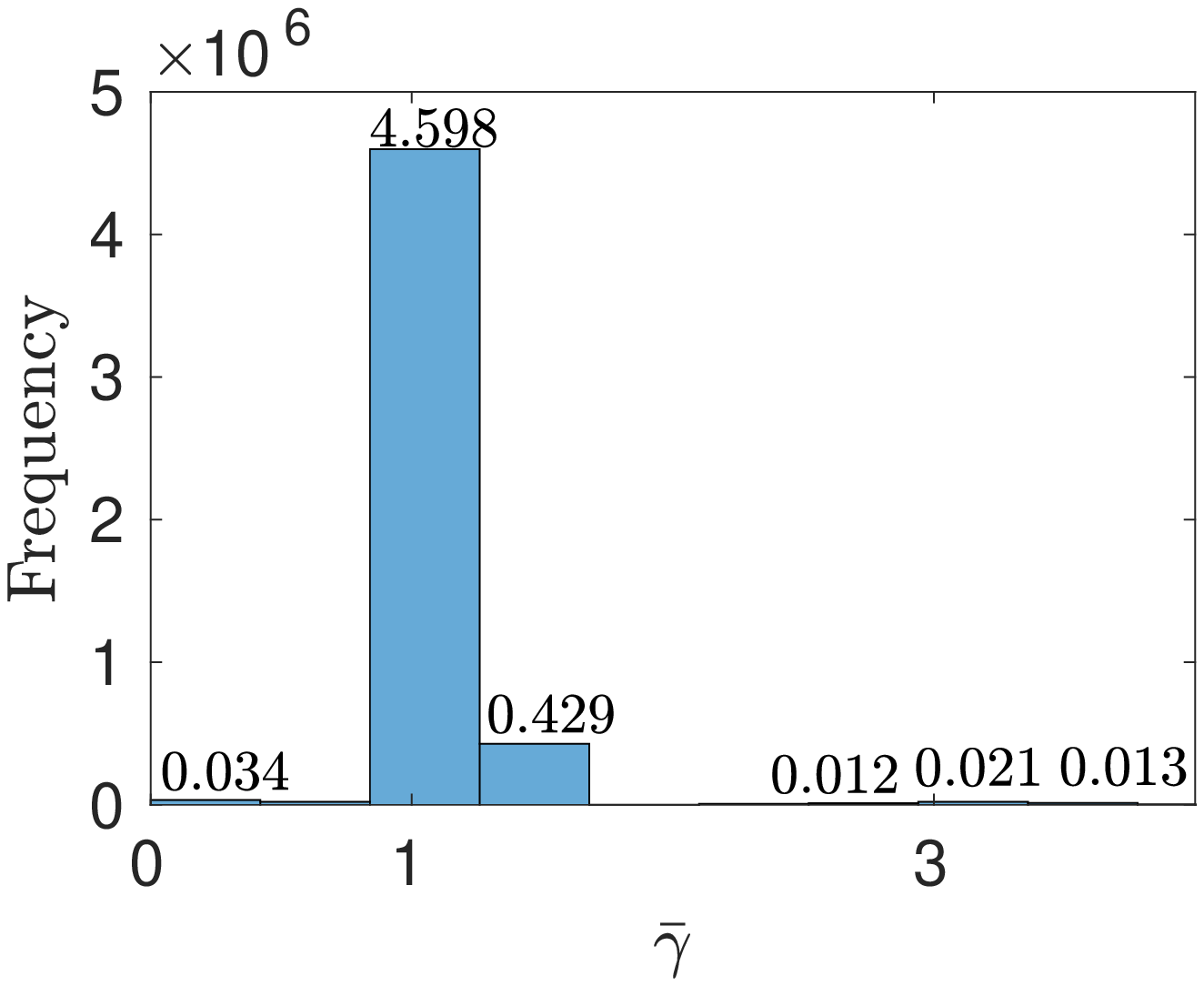}
        \label{fig:gammadist}
        }
    \hfill
    \label{fig:paradist}
    \caption{Distribution of $\bar{\lambda}_h$, $\bar{\lambda}_s$, $\bar{\lambda}_v$ and $\bar{\gamma}$.}
\end{figure}

In order to set a reasonable range for $\lambda_p$, we first compute the distribution of $\lambda_p$ by 
\begin{equation}
    \bar{\lambda}_p = \frac{1}{w \cdot h}\sum_{c=1}^{3}\sum_{x=1}^{w}\sum_{y=1}^{h}\frac{L_{out_p}(x,y)}{L_{in_p}(x,y)},p \in {\{H,S,V\}},
    \label{eq:lambdadist}
\end{equation}
where $\{L_{out_p},L_{in_p}\}$ is an image pair from $D_{lv}$, $(x,y)$ is the image coordinate, $w$ is the width of the image and $h$ is the height of the image. We compute $\bar{\lambda}_p$ for all possible image pairs in $D_{lv}$. The distribution of $\bar{\lambda}_p$ is shown in Figure \ref{fig:hlambdadist}, \ref{fig:slambdadist} and \ref{fig:vlambdadist}. Due to their symmetry, we sample $\lambda_p$ from $[0.5,1]$.

\begin{figure}
    \centering
    \includegraphics[width = \linewidth]{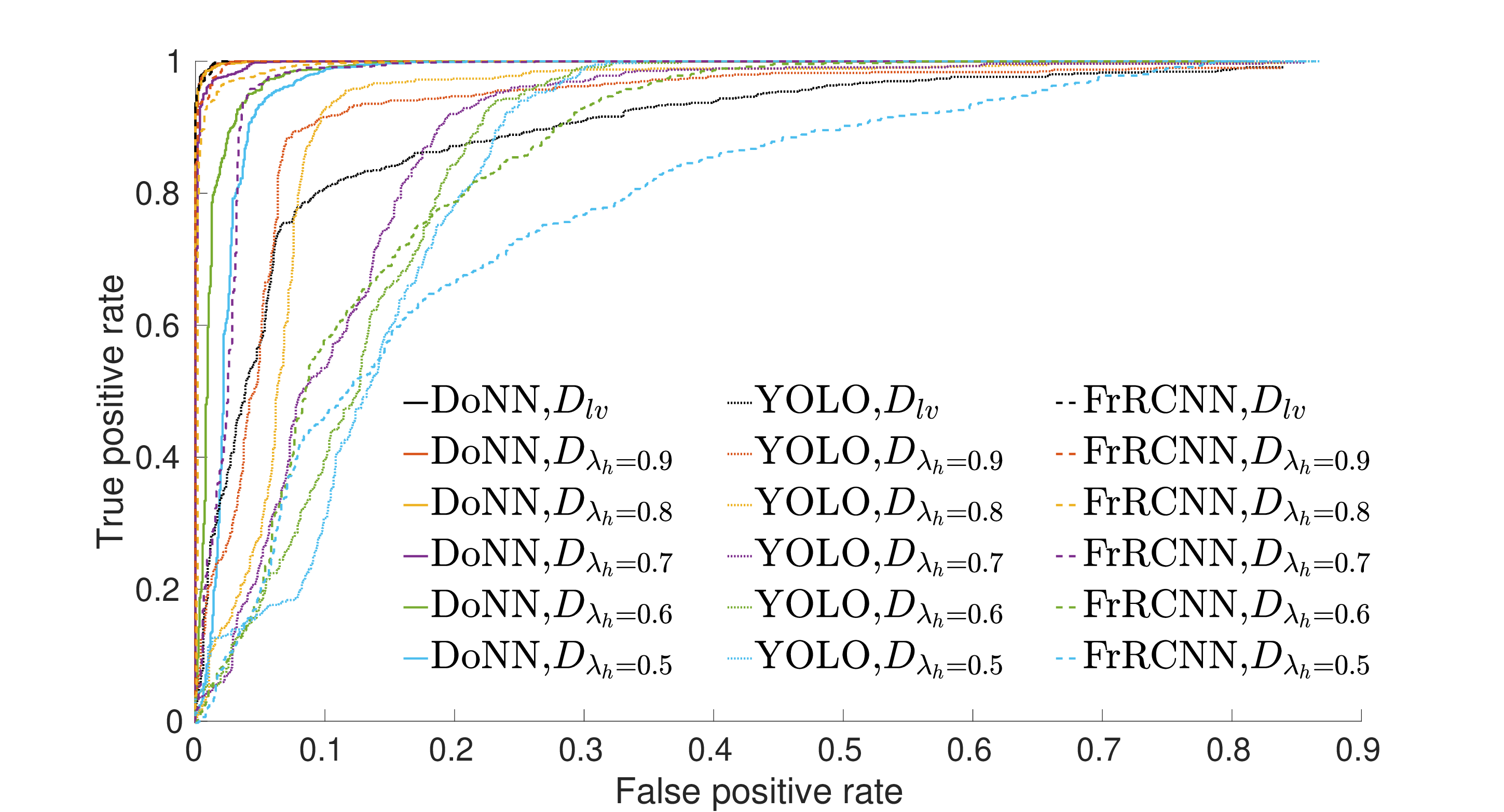}
    \caption{ROC curves of DoNN, YOLO, Faster-RCNN computed under hue shift. Corresponding AUCs are given in Table \ref{tbl:AUChshift}.}
    \label{fig:ROChshift}
\end{figure}

Sample images after hue shift deformation and corresponding detection results of DoNN and FasterRCNN are shown in Table \ref{tbl:hueresfrrcnnanddonn}. In this sample, the increment of hue shift results in less confidence for DoNN while total incorrect detection for FasterRCNN. Figure \ref{fig:ROChshift} and Table \ref{tbl:AUChshift} show the ROC curve and associated AUC of three models in hue shift, respectively. DoNN outperforms FasterRCNN in all cases. DoNN can still achieve an acceptable performance in the extreme case $\lambda_h=0.5$ while FasterRCNN performs poorly. Notable performance difference between these two models occurs at $\lambda_h = 0.5$ and $\lambda_h = 0.6$. Table \ref{tbl:huefeatfrrcnn} shows feature maps of the first, second and fifth convolution layer of FasterRCNN at $\lambda_h = 0.5$ and $\lambda_h = 0.6$. Activation of $\mathcal{T}_1$ and $\mathcal{T}_5$ is very weak in the region of the docking station, giving rise to the final incorrect detection. Table \ref{tbl:huedonnfeat} shows feature maps of the first, fifth convolution layer, confidence map $S$ and detection results of DoNN at $\lambda_h = 0.5$ and $\lambda_h = 0.6$. DoNN remains relatively high activation in the docking station region in $\mathcal{T}_1$ in Table \ref{tbl:huedonnfeat}, resulting in less confident but correct detection of docking stations. The confidence map $S$ of DoNN shows that DoNN feels more uncertain than cases without hue shift. Three neighbouring grids are with similar confidence, but the correct grid overwhelms.

\begin{figure}
    \centering
    \includegraphics[width = \linewidth, height = 5.5cm]{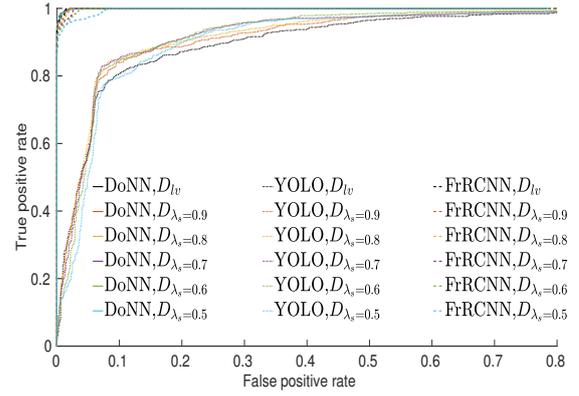}
    \caption{ROC curves of DoNN, YOLO, and Faster-RCNN computed under saturation shift. The corresponding AUCs are given in Table \ref{tbl:AUCsshift}.}
    \label{fig:ROCsshift}
\end{figure}

Saturation indicates the amount of grey in the color. A color is grey when its saturation value is $0$ while is primary color when its saturation value is $1$. Table \ref{tbl:saturresfrrcnnanddonn} shows sample images after saturation shift and associated detection results of FasterRCNN and DoNN. The color becomes closer to grey with the increment of $\lambda_s$. We show ROC curve and associated AUC of three models in Figure \ref{fig:ROCsshift} and Table \ref{tbl:AUCsshift}. As depicted, DoNN outperforms FasterRCNN in all levels of Saturation shift. The AUC of FasterRCNN declines from $0.99958$ to $0.99713$ while DoNN from $0.99964$ to $0.99953$ as $\lambda_s$ varies from $1$ to $0.5$. 

\begin{figure}
    \centering
    \includegraphics[width = \linewidth, height = 5.5cm]{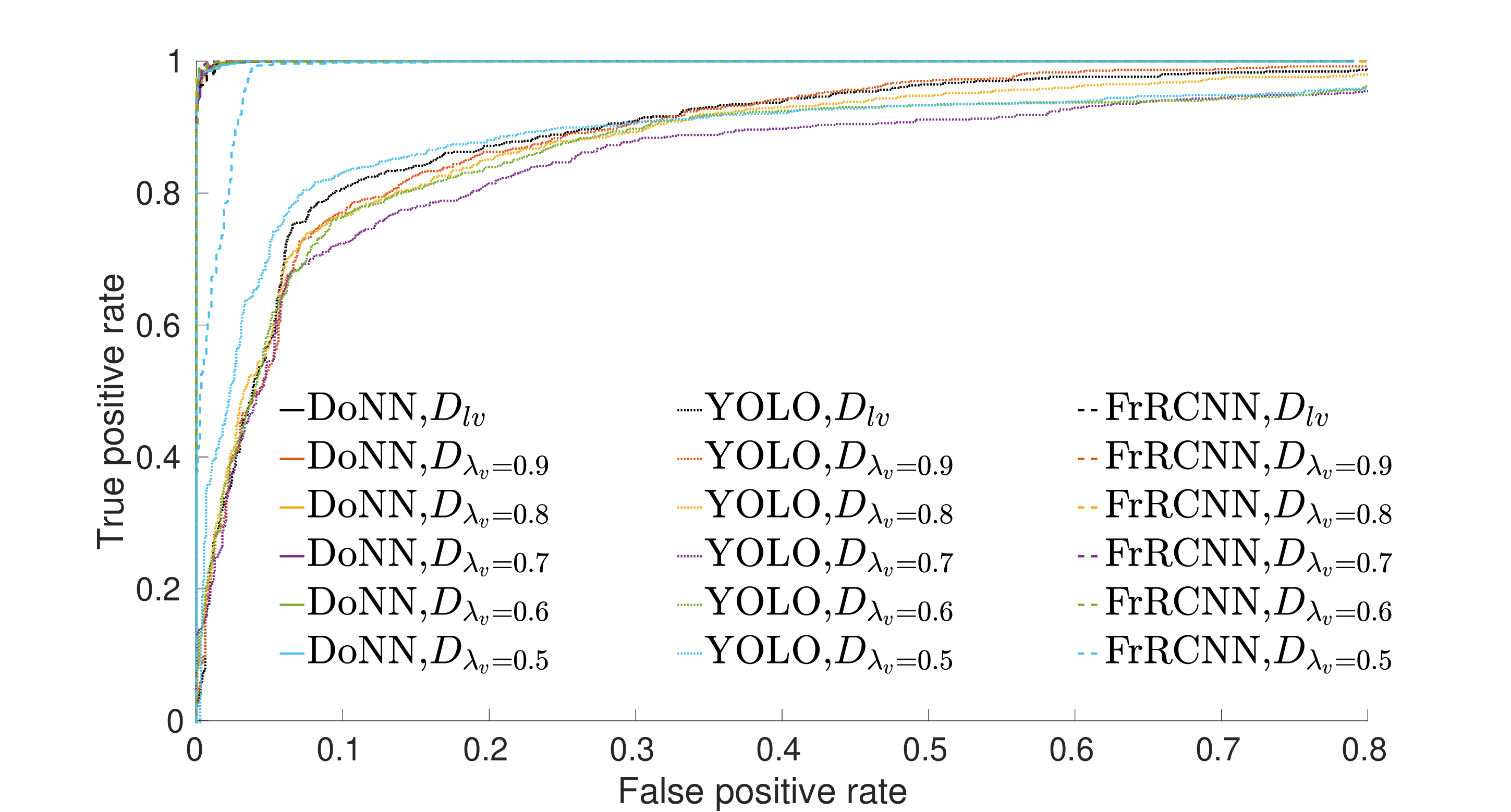}
    \caption{ROC curves of DoNN, YOLO, Faster-RCNN computed under for value shift. Corresponding AUCs are given in Table \ref{tbl:AUCvshift}.}
    \label{fig:ROCvshift}
\end{figure}

Value component describes the brightness of colors. Images become less bright with the increment of $\lambda_v$. $\lambda_v$ with a low value corresponds to a dim underwater environment. As ROC curves and associated AUC shown in Figure \ref{fig:ROCvshift} and Table \ref{tbl:AUCvshift}, DoNN and FasterRCNN are both robust to Value shift. No significant degradation arises in different levels of Value shift. DoNN goes ahead in the case of $\lambda_{v} = 0.5$ and lags slightly behind FasterRCNN in other cases. But DoNN outperforms FasterRCNN in terms of average performance. The average AUC of DoNN is $0.99965$ overall in contrast to $0.99774$ of FasterRCNN. 

To sum up, DoNN is more robust and credible than FasterRCNN in underwater images with color shift.

\begin{table}
\caption{Sample images after hue shift deformation and detection results predicted by FasterRCNN and DoNN.}
\label{tbl:hueresfrrcnnanddonn}
    \begin{tabular}{p{0.6cm} m{.38\linewidth} m{.38\linewidth}}
        \hline
        $\lambda_h$ & \multicolumn{1}{c}{FrRCNN} & \multicolumn{1}{c}{DoNN} \\  
        \hline
        \\[-2ex]
        $1$ &
        \parbox[c]{\linewidth}{\includegraphics[width=\linewidth, height = 0.81 \linewidth]{result/feat/frrcnn/paper_testsets/detect}} &
        \parbox[c]{\linewidth}{\includegraphics[width=\linewidth, height = 0.81 \linewidth]{result/feat/donn/paper_testsets/detect}} \\
        \\[-2ex] 
        \hline
        \\[-2ex]
        $0.9$ &
        \parbox[c]{\linewidth}{\includegraphics[width=\linewidth, height = 0.81 \linewidth]{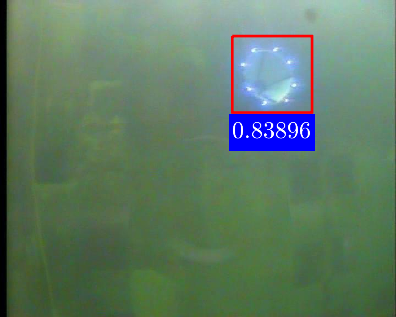}} &
        \parbox[c]{\linewidth}{\includegraphics[width=\linewidth, height = 0.81 \linewidth]{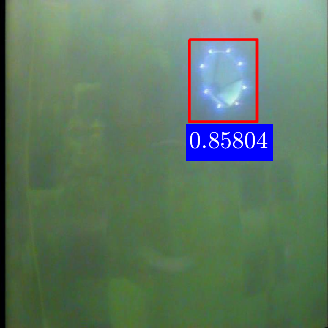}} \\
        \\[-2ex] 
        \hline
        \\[-2ex]
        $0.7$ &
        \parbox[c]{\linewidth}{\includegraphics[width=\linewidth, height = 0.81 \linewidth]{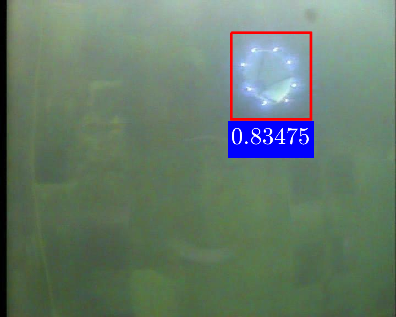}} &
        \parbox[c]{\linewidth}{\includegraphics[width=\linewidth, height = 0.81 \linewidth]{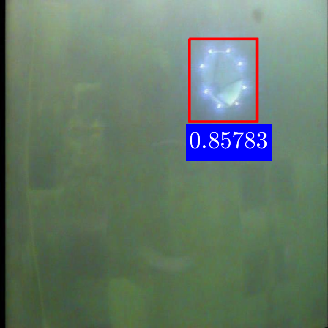}} \\
        \\[-2ex] 
        \hline
        \\[-2ex]
        $0.5$ &
        \parbox[c]{\linewidth}{\includegraphics[width=\linewidth, height = 0.81 \linewidth]{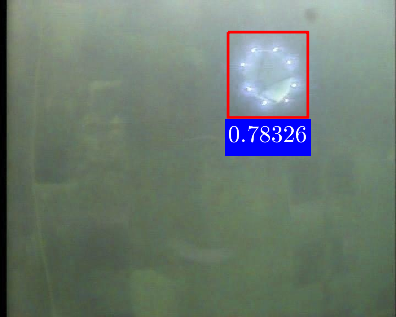}} &
        \parbox[c]{\linewidth}{\includegraphics[width=\linewidth, height = 0.81 \linewidth]{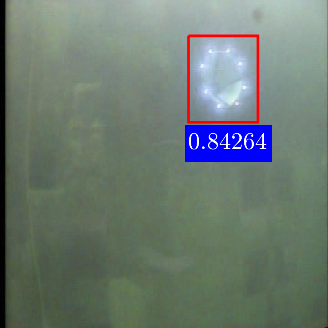}} \\
        \\[-2ex] 
        \hline
    \end{tabular}
\end{table}

\begin{table}
\caption{Sample images after saturation shift deformation and detection results predicted by FasterRCNN and DoNN.}
\label{tbl:saturresfrrcnnanddonn}
    \begin{tabular}{p{0.6cm} m{.38\linewidth} m{.38\linewidth}}
        \hline
        $\lambda_s$ & \multicolumn{1}{c}{FrRCNN} & \multicolumn{1}{c}{DoNN} \\ 
        \hline
        \\[-2ex]
        $1$ &
        \parbox[c]{\linewidth}{\includegraphics[width=\linewidth, height = 0.81 \linewidth]{result/feat/frrcnn/paper_testsets/detect}} &
        \parbox[c]{\linewidth}{\includegraphics[width=\linewidth, height = 0.81 \linewidth]{result/feat/donn/paper_testsets/detect}} \\
        \\[-2ex] 
        \hline
        \\[-2ex]
        $0.9$ &
        \parbox[c]{\linewidth}{\includegraphics[width=\linewidth, height = 0.81 \linewidth]{result/feat/frrcnn/s0.9/detect}} &
        \parbox[c]{\linewidth}{\includegraphics[width=\linewidth, height = 0.81 \linewidth]{result/feat/donn/s0.9/detect}} \\
        \\[-2ex] 
        \hline
        \\[-2ex]
        $0.7$ &
        \parbox[c]{\linewidth}{\includegraphics[width=\linewidth, height = 0.81 \linewidth]{result/feat/frrcnn/s0.7/detect}} &
        \parbox[c]{\linewidth}{\includegraphics[width=\linewidth, height = 0.81 \linewidth]{result/feat/donn/s0.7/detect}} \\
        \\[-2ex] 
        \hline
        \\[-2ex]
        $0.5$ &
        \parbox[c]{\linewidth}{\includegraphics[width=\linewidth, height = 0.81 \linewidth]{result/feat/frrcnn/s0.5/detect}} &
        \parbox[c]{\linewidth}{\includegraphics[width=\linewidth, height = 0.81 \linewidth]{result/feat/donn/s0.5/detect}} \\
        \\[-2ex] 
        \hline
    \end{tabular}
\end{table}

\begin{table}
\caption{AUCs of DoNN, YOLO and Faster-RCNN computed under hue shift, and the corresponding ROC curves are shown in Figure \ref{fig:ROChshift}.}
\label{tbl:AUChshift}
    \begin{tabular}{c|c|c|c}
    \hline
    \diagbox{$\lambda_h$}{AUC}{Model} & DoNN & YOLO & FrRCNN \\
    \hline
    1&    0.99964&0.90862&0.99958 \\
    \hline
    0.9 & 0.99956&0.93559&0.99888 \\
    \hline
    0.8 & 0.99951&0.93065&0.99564 \\ 
    \hline
    0.7 & 0.99824&0.89211&0.97628 \\
    \hline
    0.6 & 0.98667&0.87664&0.87605 \\
    \hline
    0.5 & 0.97405&0.86540&0.80774 \\
    \hline
    Average & 0.99160 & 0.9008 & 0.93092 \\
    \hline
    \end{tabular}
\end{table}

\begin{table}
\caption{AUCs of DoNN, YOLO, Faster-RCNN computed under various levels of saturation shift, and the corresponding to ROC curves are shown in Figure \ref{fig:ROCsshift}.}
\label{tbl:AUCsshift}
    \begin{tabular}{c|c|c|c}
    \hline
    \diagbox{$\lambda_s$}{AUC}{Model} & DoNN & YOLO & FrRCNN \\
    \hline
    1&    0.99964&0.90862&0.99958 \\
    \hline
    0.9 & 0.99961&0.91990&0.99941 \\
    \hline
    0.8 & 0.99961&0.92510&0.99922 \\ 
    \hline
    0.7 & 0.99960&0.92634&0.99891 \\
    \hline
    0.6 & 0.99959&0.92680&0.99851 \\
    \hline
    0.5 & 0.99953&0.91429&0.99713 \\
    \hline
    Average & 0.99959 & 0.92249 & 0.99864 \\
    \hline
    \end{tabular}
\end{table}

\begin{table}
\caption{AUCs of DoNN, YOLO, Faster-RCNN computer for various levels of value shift, and the corresponding ROC curves are shown in Figure \ref{fig:ROCvshift}.}
\label{tbl:AUCvshift}
    \begin{tabular}{c|c|c|c}
    \hline
    \diagbox{$\lambda_v$}{AUC}{Model} & DoNN & YOLO & FrRCNN \\
    \hline
    1&    0.99964&0.90862&0.99958 \\
    \hline
    0.9 & 0.99964&0.90621&0.99981 \\
    \hline
    0.8 & 0.99965&0.89723&0.99988 \\ 
    \hline
    0.7 & 0.99963&0.87105&0.99976 \\
    \hline
    0.6 & 0.99968&0.88644&0.99979 \\
    \hline
    0.5 & 0.99965&0.90317&0.98946 \\
    \hline
    Average & 0.99965 & 0.89282 & 0.99774 \\
    \hline
    \end{tabular}
\end{table}
\begin{table}
\caption{Feature maps $\mathcal{T}_i$ of the $i^{th}$ layer $Conv_i,i = 1,2,5$ and detection results provided by FasterRCNN for $\lambda_h = 0.6$ and $\lambda_h = 0.5$.}
\label{tbl:huefeatfrrcnn}
    \begin{tabular}{p{0.6cm} m{.38\linewidth} m{.38\linewidth}}
        \hline
        Item & \multicolumn{1}{c}{$\lambda_h = 0.6$} & \multicolumn{1}{c}{$\lambda_h = 0.5$} \\ 
        \hline
		\\[-2ex] 
        $\mathcal{T}_1$ &
        \parbox[c]{\linewidth}{\includegraphics[width=  \linewidth, height = 0.81 \linewidth]{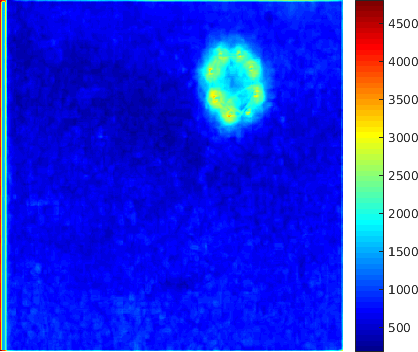}} &
        \parbox[c]{\linewidth}{\includegraphics[width=\linewidth, height = 0.81 \linewidth]{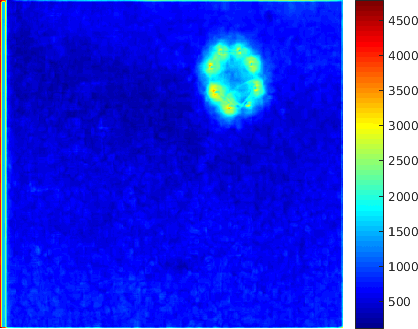}}\\
		\\[-2ex]
        \hline
        \\[-2ex] 
        $\mathcal{T}_2$ &
        \parbox[c]{\linewidth}{\includegraphics[width=\linewidth, height = 0.81 \linewidth]{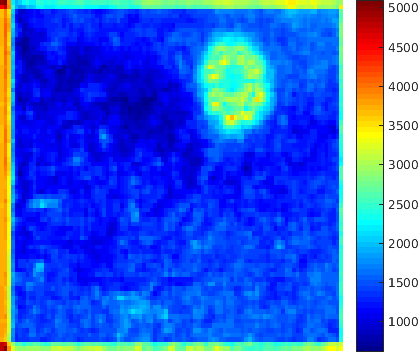}} &
        \parbox[c]{\linewidth}{\includegraphics[width=\linewidth, height = 0.81 \linewidth]{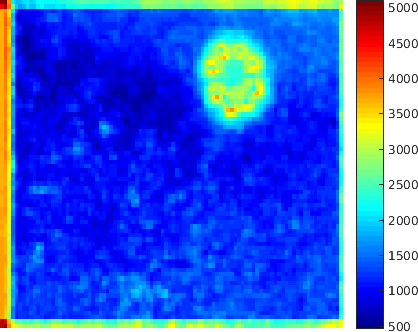}} \\
        \\[-2ex] 
        \hline
        \\[-2ex] 
        $\mathcal{T}_5$ &
        \parbox[c]{\linewidth}{\includegraphics[width=\linewidth, height = 0.81 \linewidth]{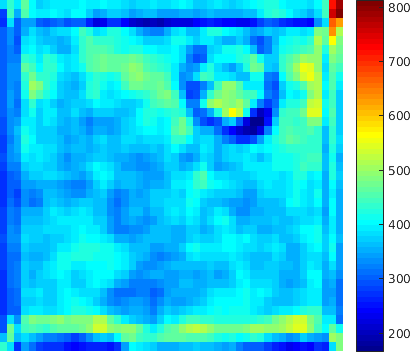}} &
        \parbox[c]{\linewidth}{\includegraphics[width=\linewidth, height = 0.81 \linewidth]{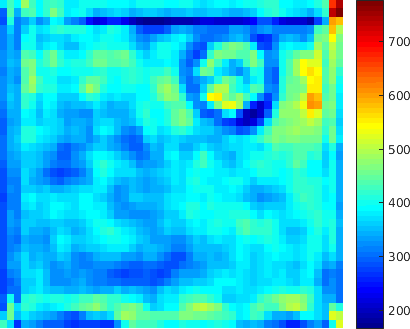}} \\
        \\[-2ex] 
        \hline
        \\[-2ex] 
        Det. Res. &
        \parbox[c]{\linewidth}{\includegraphics[width=\linewidth, height = 0.81 \linewidth]{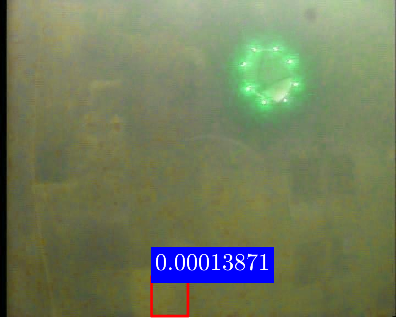}} &
        \parbox[c]{\linewidth}{\includegraphics[width=\linewidth, height = 0.81 \linewidth]{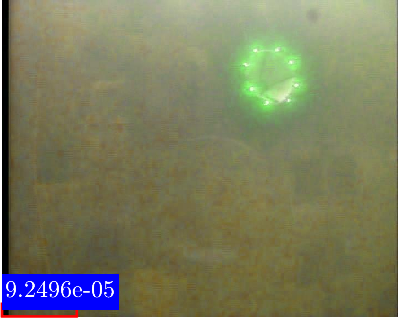}} \\
        \\[-2ex] 
        \hline
        \\[-2ex] 
    \end{tabular}
\end{table}

\begin{table}
\caption{Feature maps $\mathcal{T}_i$ of the $i^{th}$ layer $Conv_i,i = 1,5$, confidence map and detection results provided by DoNN for $\lambda_h = 0.6$ and $\lambda_h = 0.5$.}
\label{tbl:huedonnfeat}
    \begin{tabular}{p{0.6cm} m{.38\linewidth} m{.38\linewidth}}
        \hline
        Item & \multicolumn{1}{c}{$\lambda_h = 0.6$} & \multicolumn{1}{c}{$\lambda_h = 0.5$} \\ 
        \hline
        \\[-2ex]
        $\mathcal{T}_1$ &
        \parbox[c]{\linewidth}{\includegraphics[width=\linewidth, height = 0.81 \linewidth]{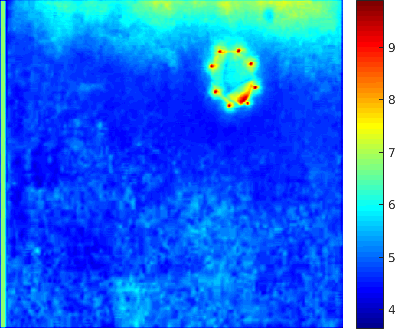}} &
        \parbox[c]{\linewidth}{\includegraphics[width=\linewidth, height = 0.81 \linewidth]{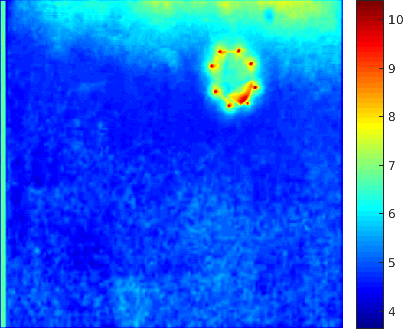}} \\
        \\[-2ex] 
        \hline
        \\[-2ex]
        $\mathcal{T}_5$ &
        \parbox[c]{\linewidth}{\includegraphics[width=\linewidth, height = 0.81 \linewidth]{result/feat/frrcnn/h0.6/conv5}} &
        \parbox[c]{\linewidth}{\includegraphics[width=\linewidth, height = 0.81 \linewidth]{result/feat/frrcnn/h0.5/conv5}} \\
        \\[-2ex] 
        \hline
        \\[-2ex] 
        $S$ &
        \parbox[c]{\linewidth}{\includegraphics[width=\linewidth, height = 0.81 \linewidth]{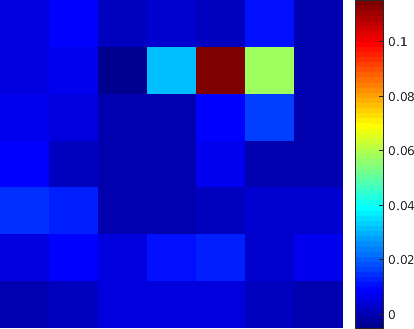}} &
        \parbox[c]{\linewidth}{\includegraphics[width=\linewidth, height = 0.81 \linewidth]{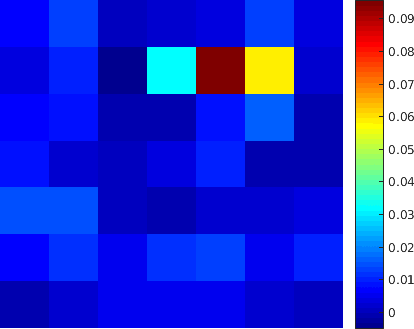}} \\
        \\[-2ex] 
        \hline
        \\[-2ex]
        Det. Res. &
        \parbox[c]{\linewidth}{\includegraphics[width=\linewidth, height = 0.81 \linewidth]{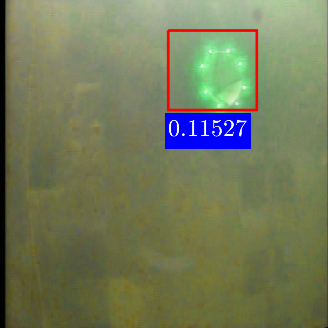}} &
        \parbox[c]{\linewidth}{\includegraphics[width=\linewidth, height = 0.81 \linewidth]{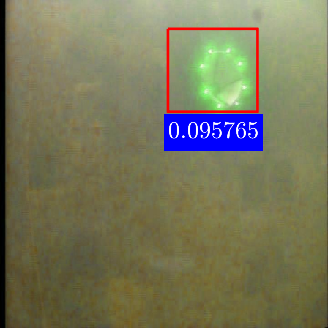}} \\
        \\[-2ex] 
        \hline
    \end{tabular}
\end{table}

\begin{table}
\caption{Feature maps $\mathcal{T}_i$ of the $i^{th}$ layer $Conv_i,i = 1,2,5$ and detection results provided by FasterRCNN for $\lambda_s = 0.6$ and $\lambda_s = 0.5$.}
\label{tbl:saturfeatfrrcnn}
    \begin{tabular}{p{0.6cm} m{.38\linewidth} m{.38\linewidth}}
        \hline
        Item & \multicolumn{1}{c}{$\lambda_s = 0.6$} & \multicolumn{1}{c}{$\lambda_s = 0.5$} \\ 
        \hline
        \\[-2ex]
        $\mathcal{T}_1$ &
        \parbox[c]{\linewidth}{\includegraphics[width=\linewidth, height = 0.81 \linewidth]{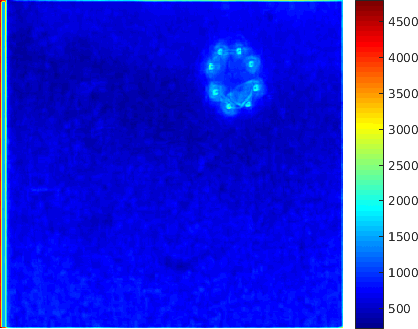}} &
        \parbox[c]{\linewidth}{\includegraphics[width=\linewidth, height = 0.81 \linewidth]{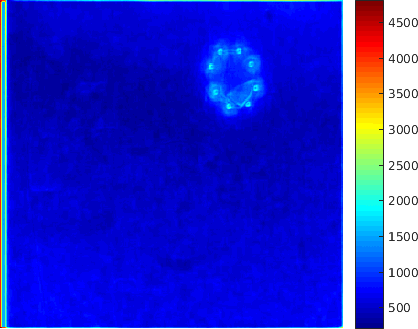}} \\
        \\[-2ex] 
        \hline
        \\[-2ex]
        $\mathcal{T}_2$ &
        \parbox[c]{\linewidth}{\includegraphics[width=\linewidth, height = 0.81 \linewidth]{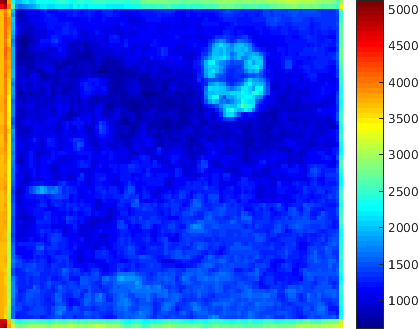}} &
        \parbox[c]{\linewidth}{\includegraphics[width=\linewidth, height = 0.81 \linewidth]{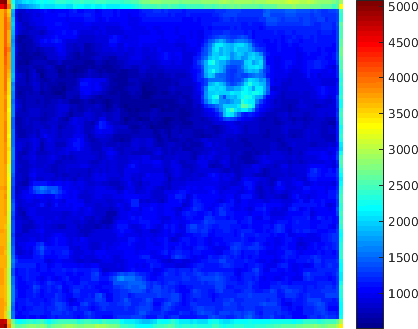}} \\
        \\[-2ex] 
        \hline
        \\[-2ex]
        $\mathcal{T}_5$ &
        \parbox[c]{\linewidth}{\includegraphics[width=\linewidth, height = 0.81 \linewidth]{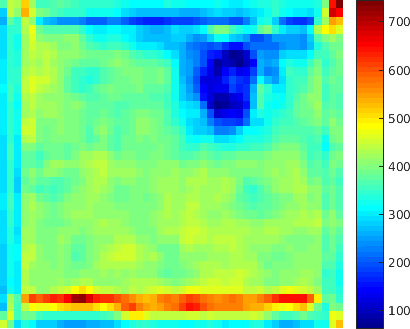}} &
        \parbox[c]{\linewidth}{\includegraphics[width=\linewidth, height = 0.81 \linewidth]{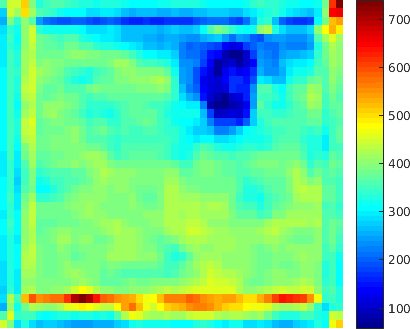}} \\
        \\[-2ex] 
        \hline
        \\[-2ex]
        Det. Res. &
        \parbox[c]{\linewidth}{\includegraphics[width=\linewidth, height = 0.81 \linewidth]{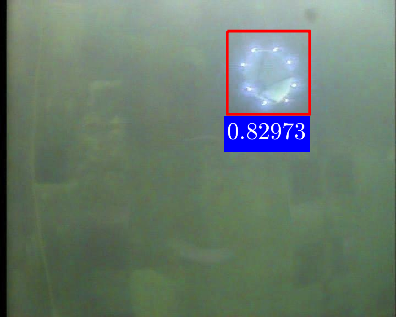}} &
        \parbox[c]{\linewidth}{\includegraphics[width=\linewidth, height = 0.81 \linewidth]{result/feat/frrcnn/s0.5/detect}} \\
        \\[-2ex] 
        \hline
    \end{tabular}
\end{table}

\begin{table}
\caption{Feature maps $\mathcal{T}_i$ of the $i^{th}$ layer $Conv_i,i = 1,5$, confidence map and detection results provided by DoNN for $\lambda_s = 0.6$ and $\lambda_s = 0.5$.}
\label{tbl:saturdonnfeat}
    \begin{tabular}{p{0.6cm} m{.38\linewidth} m{.38\linewidth}}
        \hline
        Item & \multicolumn{1}{c}{$\lambda_s = 0.6$} & \multicolumn{1}{c}{$\lambda_s = 0.5$} \\ 
        \hline
        \\[-2ex]
        $\mathcal{T}_1$ &
        \parbox[c]{\linewidth}{\includegraphics[width=\linewidth, height = 0.81 \linewidth]{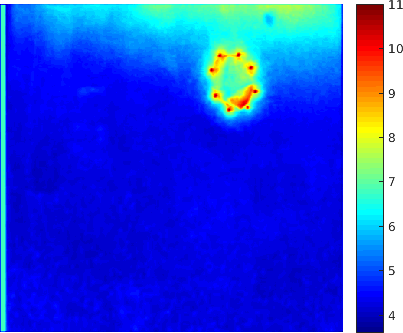}} &
        \parbox[c]{\linewidth}{\includegraphics[width=\linewidth, height = 0.81 \linewidth]{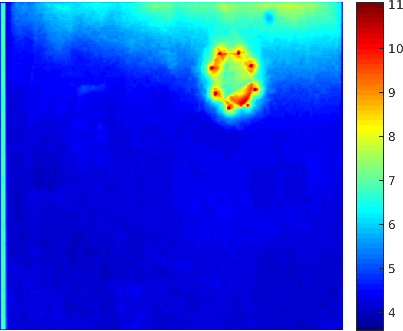}} \\
        \\[-2ex] 
        \hline
        \\[-2ex]
        $\mathcal{T}_5$ &
        \parbox[c]{\linewidth}{\includegraphics[width=\linewidth, height = 0.81 \linewidth]{result/feat/frrcnn/s0.6/conv5}} &
        \parbox[c]{\linewidth}{\includegraphics[width=\linewidth, height = 0.81 \linewidth]{result/feat/frrcnn/s0.5/conv5}} \\
        \\[-2ex] 
        \hline
        \\[-2ex] 
        $S$ &
        \parbox[c]{\linewidth}{\includegraphics[width=\linewidth, height = 0.81 \linewidth]{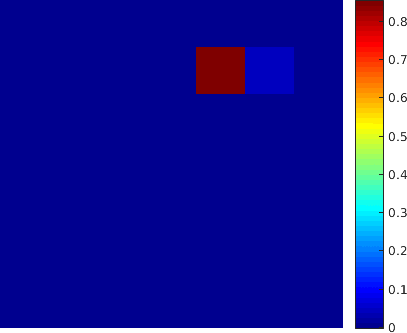}} &
        \parbox[c]{\linewidth}{\includegraphics[width=\linewidth, height = 0.81 \linewidth]{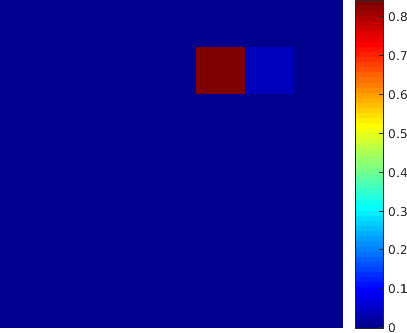}} \\
        \\[-2ex] 
        \hline
        \\[-2ex]
        Det. Res. &
        \parbox[c]{\linewidth}{\includegraphics[width=\linewidth, height = 0.81 \linewidth]{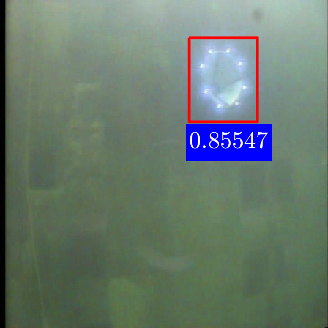}} &
        \parbox[c]{\linewidth}{\includegraphics[width=\linewidth, height = 0.81 \linewidth]{result/feat/donn/s0.5/detect}} \\
        \\[-2ex] 
        \hline
    \end{tabular}
\end{table}

\subsubsection{Comparison of performance of detection algorithms under contrast shift} 

In order to compare the performance of the detection methods in underwater environment under change of contrast, we generate contrast adjustment datasets using gamma transformation 
\begin{equation}
    I_{out} = I_{in}^{\gamma},
\end{equation}
where $I_{out}$, $I_{in}$ and $\gamma$ are input images, output images and the parameter of gamma transformation, respectively. Contrast of images increases as $\gamma<1$ while the contrast decreases as $\gamma>1$. The dataset obtained after contrast deformation is denoted by $D_{ct\gamma}$.

We computed the distribution of $\gamma$, which is shown in Figure~\ref{fig:gammadist}, by 

\begin{equation}
    \bar{\gamma} = \frac{1}{w \cdot h}\sum_{c=1}^{3}\sum_{x=1}^{w}\sum_{y=1}^{h}\frac{logI_{out}(x,y)}{logI_{in}(x,y)},
    \label{eq:gammadis}
\end{equation}
where $w$, $h$  and $c$ are the width, height and channel of $I_{in}$ and $I_{out}$. \{$I_{in}$, $I_{out}$\} is an image pair belonging to $D_{lv}$. $I_{out}(x,y)$ denotes the pixel value of $I_{out}$ at the $(x,y)$ location. According to the distribution of $\gamma$ shown in Figure \ref{fig:gammadist}, we sample $\gamma$ from $\gamma \in [0.2,3.5]$. 

\begin{figure}
    \centering
    \includegraphics[width = \linewidth]{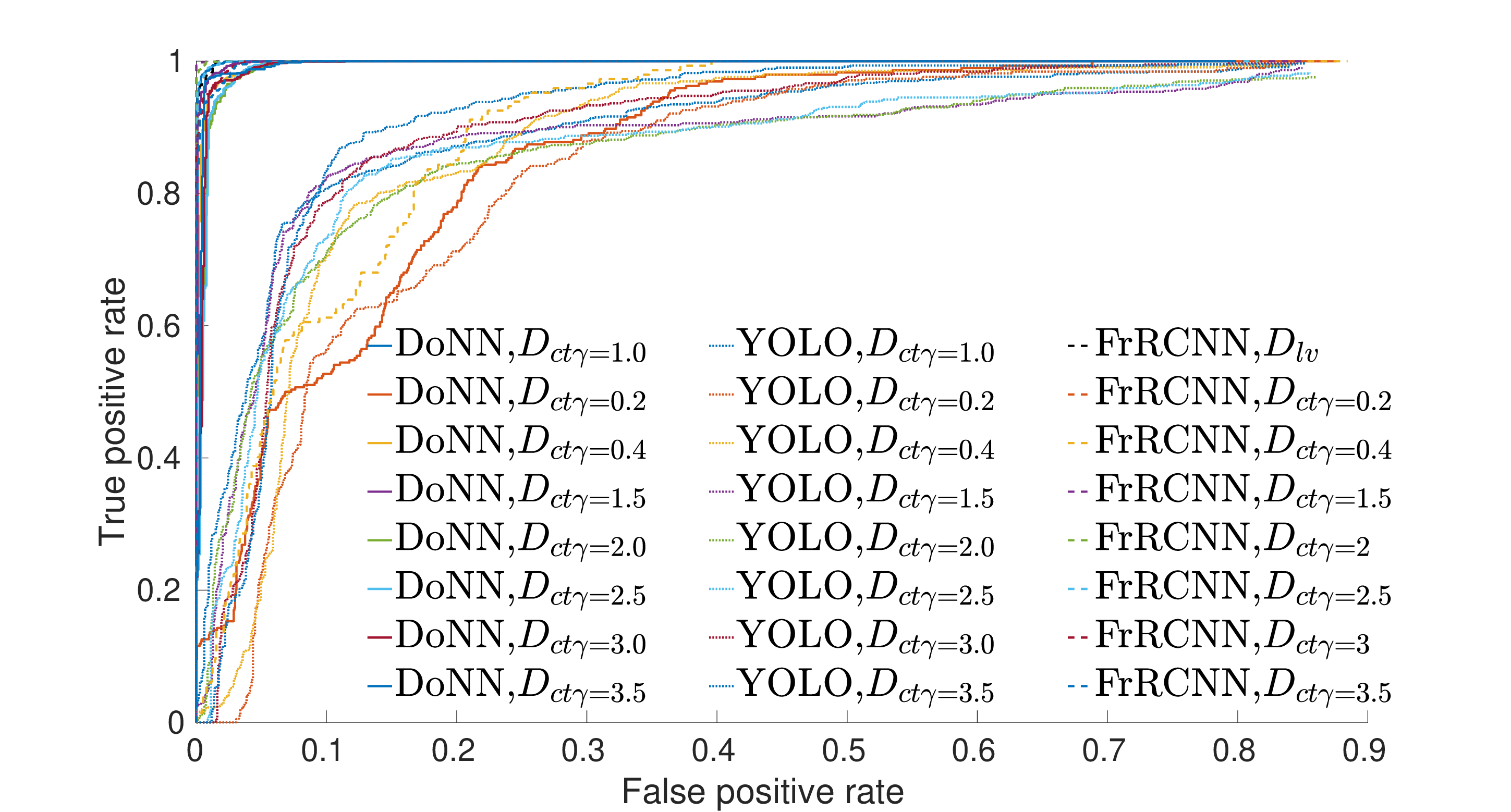}
    \caption{ROC curves of DoNN, YOLO, and Faster-RCNN computer under various contrast conditions. The corresponding AUCs are shown in Table \ref{tbl:AUCcontrast}.}
    \label{fig:ROCcontrast}
\end{figure}

Table \ref{tbl:ctrresfrrcnnanddonn} shows sample images obtained by contrast adjustment of different levels. Intuitively, the docking station is more clearly observed as $\gamma$ increases, and is less clear as $\gamma$ decreases in the image. We give ROC curves and the associated AUCs of three detection algorithms in Figure~\ref{fig:ROCcontrast} and Table~\ref{tbl:AUCcontrast}, respectively. On average, DoNN performs better than FasterRCNN. The average AUC of DoNN is $0.97857$ over all levels of contrast adjustment while that of FasterRCNN is $0.84282$. As $\gamma$ is high, Faster-RCNN outperforms DoNN by a tiny margin, but lags behind significantly if $\gamma$ is low. The AUC of Faster-RCNN becomes $ 0.90320$ if $\gamma= 0.4$, and even more acute with $AUC = 0$, for $\gamma =  0.2$. This means that all the detections fail to surpass the IoU criterion ($50\%$). It is primarily owing to the weak activation of FasterRCNN in feature maps. We compare feature maps $\mathcal{T}_1$ and $\mathcal{T}_5$ of DoNN and FasterRCNN in Table \ref{tbl:contrastfeatfrrcnn} and \ref{tbl:contrastdonnfeat} to examine the performance difference between DoNN and FasterRCNN. For $\gamma = 0.4$, activation of $\mathcal{T}_1$ and $\mathcal{T}_5$ of FasterRCNN becomes quite weak in the docking station region, resulting in the final incorrect detection, as shown in Table \ref{tbl:contrastfeatfrrcnn}. It becomes more acute for $\gamma = 0.2$. Activations computed using $\mathcal{T}_1$ and $\mathcal{T}_5$ are almost as weak as computed in their background in the docking station region, and thus they provide incorrect prediction. However, it provides high activation values, and salient spatial structural patterns in $\mathcal{T}_1$ and $\mathcal{T}_5$ of DoNN for $\gamma = 0.4$ as shown in Table \ref{tbl:contrastdonnfeat}. As a result, relative high confidence is obtained for only one grid in the confidence map $S$ of DoNN. When $\gamma$ is equal to $0.2$, DoNN keeps a distinguishable docking station spatial pattern in its $\mathcal{T}_1$ and $\mathcal{T}_5$, although activation gets weaker than $\gamma = 0.4$. The confidence map $S$ contains more grids with relatively high confidence, but only the correct one overwhelms as shown in Table \ref{tbl:contrastdonnfeat}. 

\begin{table}
\caption{AUCs of DoNN, YOLO, and Faster-RCNN computed under various contrast conditions, and the corresponding ROC curves are shown in Figure \ref{fig:ROCcontrast}.}
\label{tbl:AUCcontrast}
    \begin{tabular}{c|c|c|c}
    \hline
    \diagbox{$\gamma$}{AUC}{Model} & DoNN & YOLO & FrRCNN \\
    \hline
    1&    0.99964&0.90862&0.99958 \\
    \hline
    0.2 & 0.87402&0.84775&0 \\
    \hline
    0.4 & 0.99738&0.88640&0.90320 \\ 
    \hline
    1.5 & 0.99884&0.88991&0.99997 \\
    \hline
    2.0 & 0.99439&0.87390&0.99990 \\
    \hline
    2.5 & 0.99392&0.87943&0.99962 \\
    \hline
    3 & 0.99505&0.90531&0.99873 \\
    \hline
    3.5 & 0.99637&0.91929&0.99833 \\
    \hline
    Average & 0.97857 & 0.88600 & 0.84282 \\
    \hline
    \end{tabular}
\end{table}

\begin{table}
\caption{Sample images obtained by contrast adjustment, and the corresponding detection results provided by FasterRCNN and DoNN.}
\label{tbl:ctrresfrrcnnanddonn}
    \begin{tabular}{p{0.6cm} m{.38\linewidth} m{.38\linewidth}}
        \hline
        $\gamma$ & \multicolumn{1}{c}{FrRCNN} & \multicolumn{1}{c}{DoNN} \\ 
        \hline
        \\[-2ex]
        $0.2$ &
        \parbox[c]{\linewidth}{\includegraphics[width=\linewidth, height = 0.82 \linewidth]{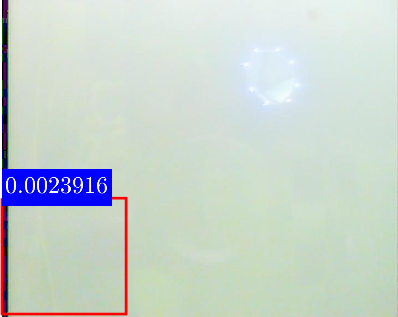}} &
        \parbox[c]{\linewidth}{\includegraphics[width=\linewidth, height = 0.82 \linewidth]{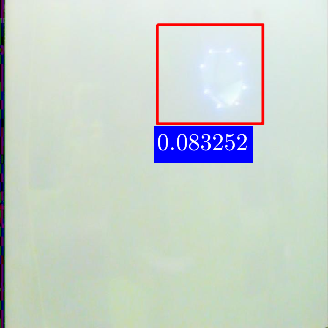}} \\
        \\[-2ex] 
        \hline
        \\[-2ex]
        $0.4$ &
        \parbox[c]{\linewidth}{\includegraphics[width=\linewidth, height = 0.82 \linewidth]{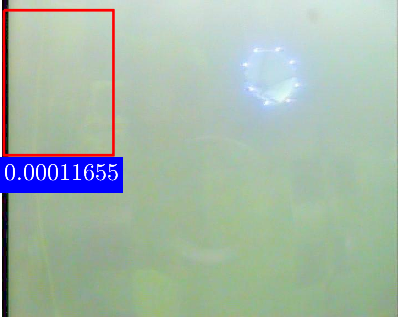}} &
        \parbox[c]{\linewidth}{\includegraphics[width=\linewidth, height = 0.82 \linewidth]{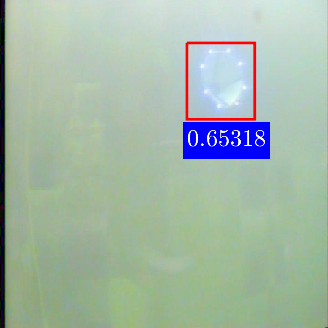}} \\
        \\[-2ex] 
        \hline
        \\[-2ex]
        $0.8$ &
        \parbox[c]{\linewidth}{\includegraphics[width=\linewidth, height = 0.82 \linewidth]{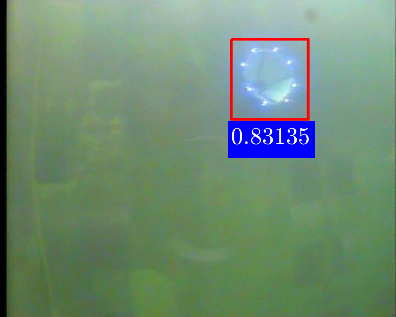}} &
        \parbox[c]{\linewidth}{\includegraphics[width=\linewidth, height = 0.82 \linewidth]{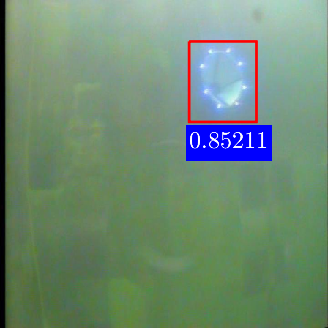}} \\
        \\[-2ex] 
        \hline
        \\[-2ex]
        $1.5$ &
        \parbox[c]{\linewidth}{\includegraphics[width=\linewidth, height = 0.82 \linewidth]{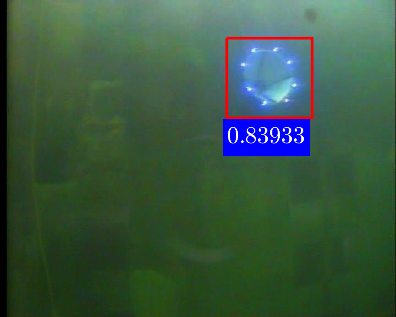}} &
        \parbox[c]{\linewidth}{\includegraphics[width=\linewidth, height = 0.82 \linewidth]{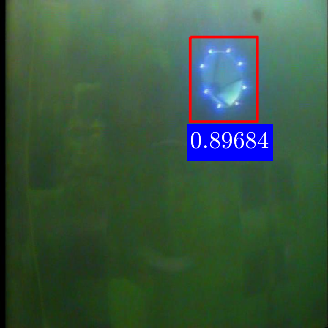}} \\
        \\[-2ex] 
        \hline
        \\[-2ex]
        $2.0$ &
        \parbox[c]{\linewidth}{\includegraphics[width=\linewidth, height = 0.82 \linewidth]{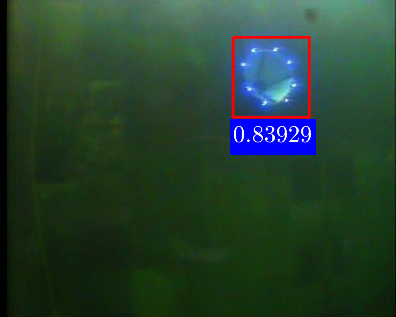}} &
        \parbox[c]{\linewidth}{\includegraphics[width=\linewidth, height = 0.82 \linewidth]{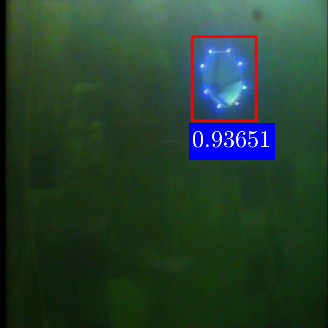}} \\
        \\[-2ex] 
        \hline
        \\[-2ex]
        $2.5$ &
        \parbox[c]{\linewidth}{\includegraphics[width=\linewidth, height = 0.82 \linewidth]{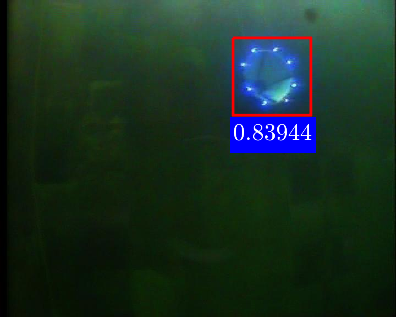}} &
        \parbox[c]{\linewidth}{\includegraphics[width=\linewidth, height = 0.82 \linewidth]{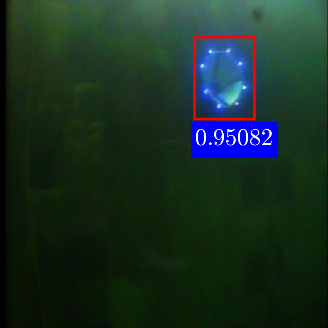}} \\
        \\[-2ex] 
        \hline
        \\[-2ex]
        $3.0$ &
        \parbox[c]{\linewidth}{\includegraphics[width=\linewidth, height = 0.82 \linewidth]{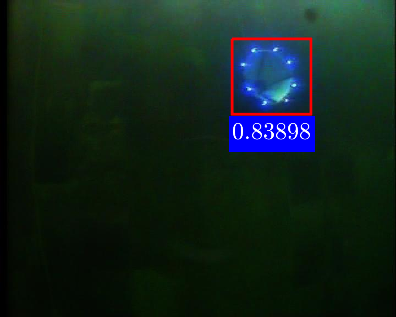}} &
        \parbox[c]{\linewidth}{\includegraphics[width=\linewidth, height = 0.82 \linewidth]{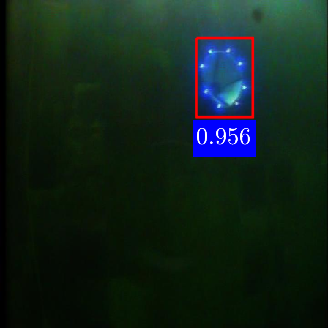}} \\
        \\[-2ex] 
        \hline
        \\[-2ex]
        $3.5$ &
        \parbox[c]{\linewidth}{\includegraphics[width=\linewidth, height = 0.82 \linewidth]{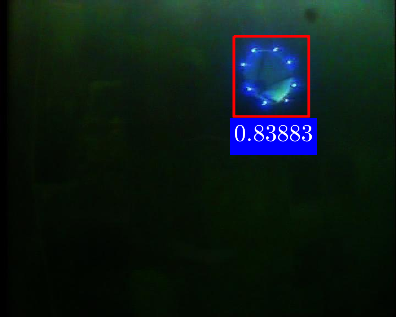}} &
        \parbox[c]{\linewidth}{\includegraphics[width=\linewidth, height = 0.82 \linewidth]{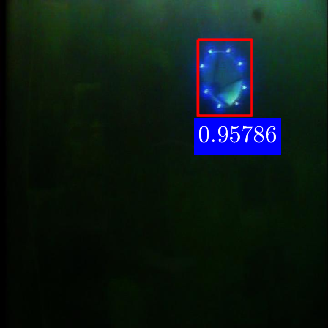}} \\
        \\[-2ex] 
        \hline
    \end{tabular}
\end{table}

\begin{table}
\caption{Feature maps $\mathcal{T}_i$ computed at the $i^{th}$ layer $Conv_i,i = 1,2,5$,  and the detection result provided by FasterRCNN for $\gamma=0.4$ and $\gamma=0.2$.}
\label{tbl:contrastfeatfrrcnn}
    \begin{tabular}{p{0.6cm} m{.38\linewidth} m{.38\linewidth}}
        \hline
        Item & \multicolumn{1}{c}{$\gamma=0.4$} & \multicolumn{1}{c}{$\gamma=0.2$} \\ 
        \hline
        \\[-2ex]
        $\mathcal{T}_1$ &
        \parbox[c]{\linewidth}{\includegraphics[width=\linewidth, height = 0.81 \linewidth]{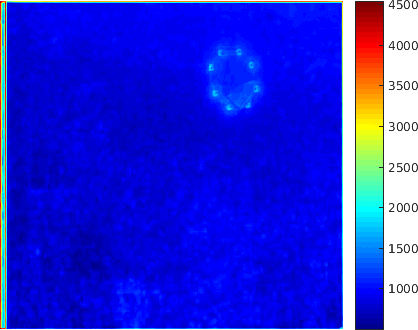}} &
        \parbox[c]{\linewidth}{\includegraphics[width=\linewidth, height = 0.81 \linewidth]{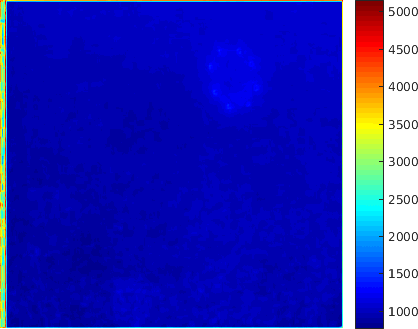}} \\
        \\[-2ex] 
        \hline
        \\[-2ex]
        $\mathcal{T}_2$ &
        \parbox[c]{\linewidth}{\includegraphics[width=\linewidth, height = 0.81 \linewidth]{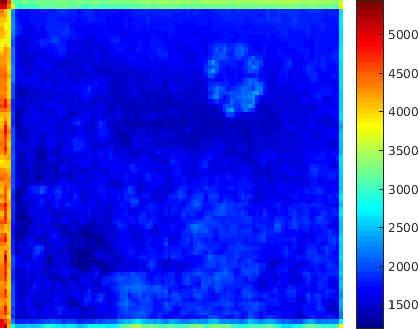}} &
        \parbox[c]{\linewidth}{\includegraphics[width=\linewidth, height = 0.81 \linewidth]{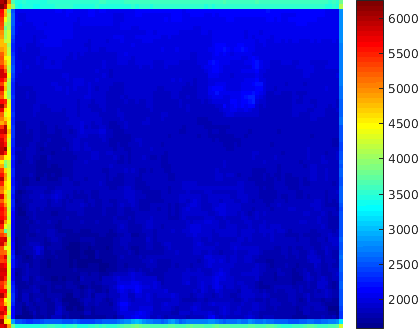}} \\
        \\[-2ex] 
        \hline
        \\[-2ex]
        $\mathcal{T}_5$ &
        \parbox[c]{\linewidth}{\includegraphics[width=\linewidth, height = 0.81 \linewidth]{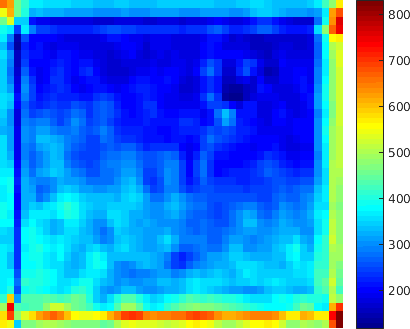}} &
        \parbox[c]{\linewidth}{\includegraphics[width=\linewidth, height = 0.81 \linewidth]{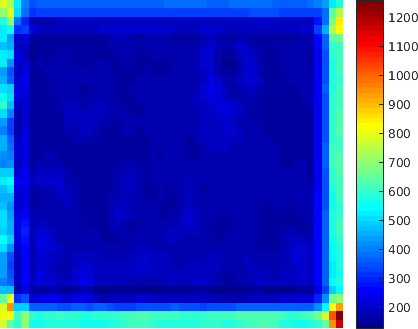}} \\
        \\[-2ex] 
        \hline
        \\[-2ex]
        Det. Res. &
        \parbox[c]{\linewidth}{\includegraphics[width=\linewidth, height = 0.81 \linewidth]{result/feat/frrcnn/contrast0.4/detect}} &
        \parbox[c]{\linewidth}{\includegraphics[width=\linewidth, height = 0.81 \linewidth]{result/feat/frrcnn/contrast0.2/detect}} \\
        \\[-2ex] 
        \hline
    \end{tabular}
\end{table}

\begin{table}
\caption{Feature maps $\mathcal{T}_i$ computed at the $i^{th}$ layer $Conv_i,i = 1,5$, confidence map, and the detection result provided by DoNN for $\gamma=0.4$ and $\gamma=0.2$.}
\label{tbl:contrastdonnfeat}
    \begin{tabular}{p{0.6cm} m{.38\linewidth} m{.38\linewidth}}
        \hline
        Item & \multicolumn{1}{c}{$\gamma=0.4$} & \multicolumn{1}{c}{$\gamma=0.2$} \\ 
        \hline
        \\[-2ex]
        $\mathcal{T}_1$ &
        \parbox[c]{\linewidth}{\includegraphics[width=\linewidth, height = 0.81 \linewidth]{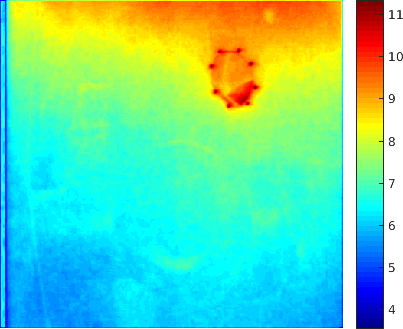}} &
        \parbox[c]{\linewidth}{\includegraphics[width=\linewidth, height = 0.81 \linewidth]{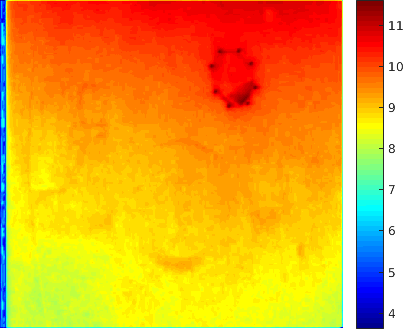}} \\
        \\[-2ex] 
        \hline
        \\[-2ex]
        $\mathcal{T}_5$ &
        \parbox[c]{\linewidth}{\includegraphics[width=\linewidth, height = 0.81 \linewidth]{result/feat/frrcnn/contrast0.4/conv5}} &
        \parbox[c]{\linewidth}{\includegraphics[width=\linewidth, height = 0.81 \linewidth]{result/feat/frrcnn/contrast0.2/conv5}} \\
        \\[-2ex] 
        \hline
        \\[-2ex] 
        $S$ &
        \parbox[c]{\linewidth}{\includegraphics[width=\linewidth, height = 0.81 \linewidth]{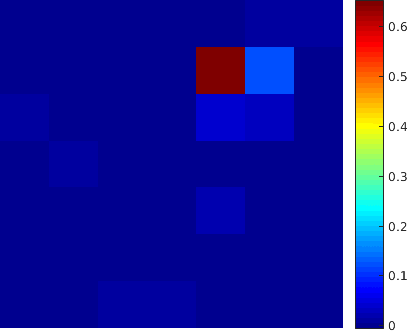}} &
        \parbox[c]{\linewidth}{\includegraphics[width=\linewidth, height = 0.81 \linewidth]{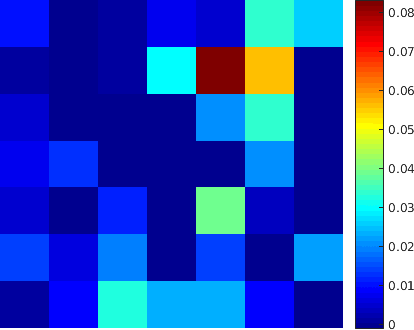}} \\
        \\[-2ex] 
        \hline
        \\[-2ex]
        Det. Res. &
        \parbox[c]{\linewidth}{\includegraphics[width=\linewidth, height = 0.81 \linewidth]{result/feat/donn/contrast0.4/detect}} &
        \parbox[c]{\linewidth}{\includegraphics[width=\linewidth, height = 0.81 \linewidth]{result/feat/donn/contrast0.2/detect}} \\
        \\[-2ex] 
        \hline
    \end{tabular}
\end{table}

\subsubsection{Comparison of performance of detection algorithms for mirror images} 
\begin{figure}
    \centering
    \includegraphics[width = \linewidth]{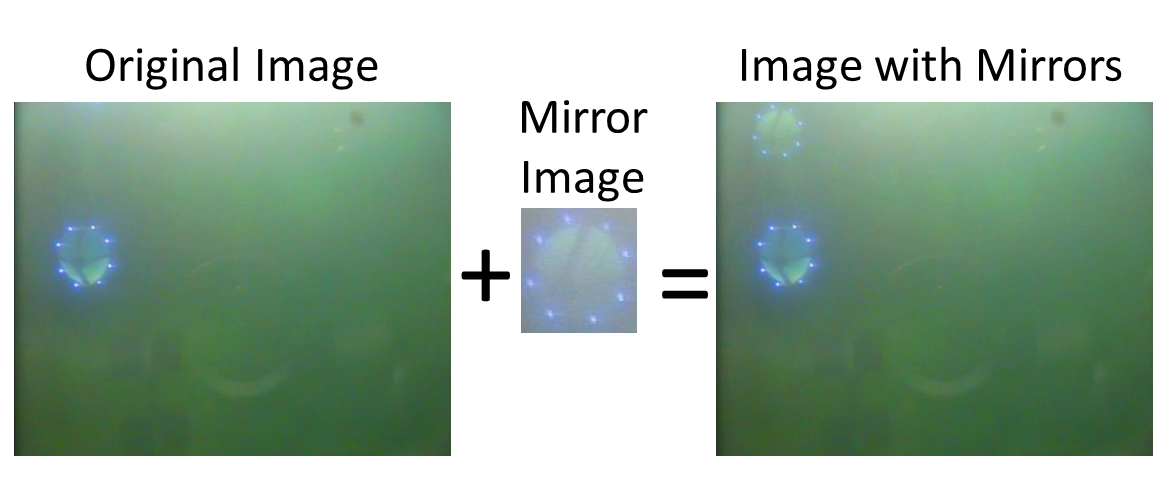}
    \caption{The merging process of synthetic mirror images. A mirror image is merged to the original image on the upper side of the real docking station.}
    \label{fig:mergingmirror}
\end{figure}

\begin{figure}
    \centering
    \captionsetup[subfigure]{justification=centering}
    \subfloat[][$\mathcal{T}_1$]{
        \centering
        \includegraphics[width=0.22\linewidth,height = 0.179\linewidth]{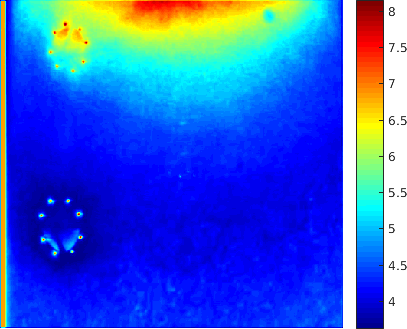}
        }
    \hfill
    \subfloat[][$\mathcal{T}_5$]{
        \centering
        \includegraphics[width=0.22\linewidth,height = 0.179\linewidth]{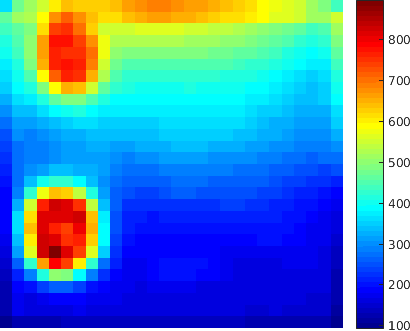}
        }
    \hfill
    \subfloat[][$S$]{
        \centering
        \includegraphics[width=0.22\linewidth,height = 0.179\linewidth]{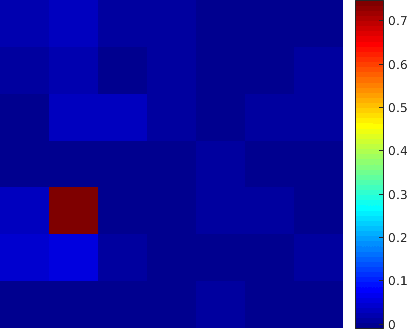}
        }
    \hfill
    \subfloat[][Detection results]{
        \centering
        \label{fig:naturemirror}
        \includegraphics[width=0.22\linewidth,height = 0.179\linewidth]{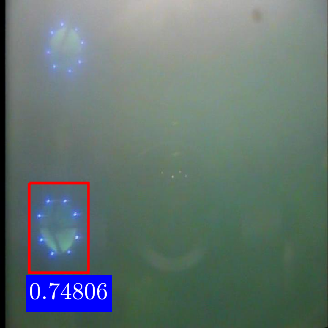}
        
    }
    \hfill\\
    \caption{Feature maps $\mathcal{T}_i$ computed at the $i^{th}$ layer $Conv_i,i = 1,5$, confidence map, and the detection result of DoNN for mirror images.}
    \label{fig:mirrordonnfeat}
\end{figure}

As mentioned in Section \ref{l_intro}, mirror images result from \textit{total internal reflection}, and they are observed when cameras are within the \textit{critical angle}, as shown in Figure \ref{fig:naturemirror}. It poses a nasty problem for detection of docking stations due to its very similar appearance with real docking stations. In order to compare performance of three methods under \textit{total internal reflection}, we establish a dataset $D_{mirr}$ by attaching a mirror image to every foreground image in $D_{lv}$. To this end, image editing \citep{perez2003poisson} is employed to merge a mirror image patch to the upper side of the docking station of original images as real as possible. The merging process is illustrated in Figure \ref{fig:mergingmirror}.

\begin{figure}
    \centering
    \subfloat[][$\mathcal{T}_1$]{
        \centering
        \includegraphics[width=0.22\linewidth,height = 0.179\linewidth]{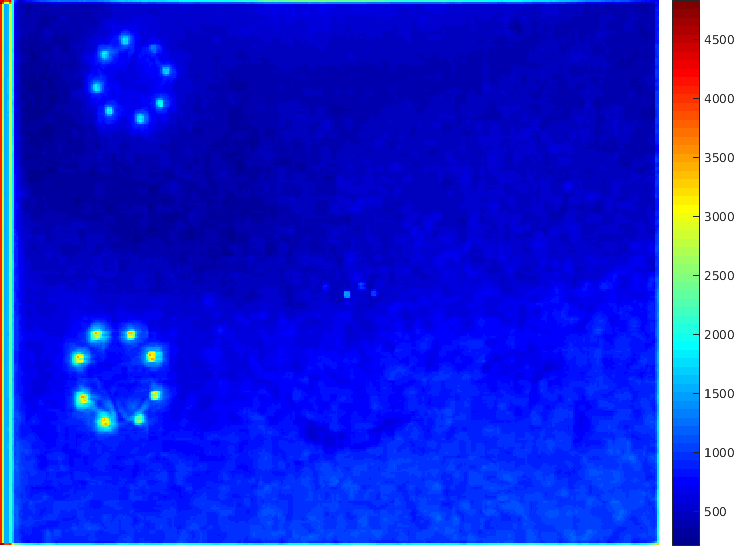}
        }
    \hfill
    \subfloat[][$\mathcal{T}_2$]{
        \centering
        \includegraphics[width=0.22\linewidth,height = 0.179\linewidth]{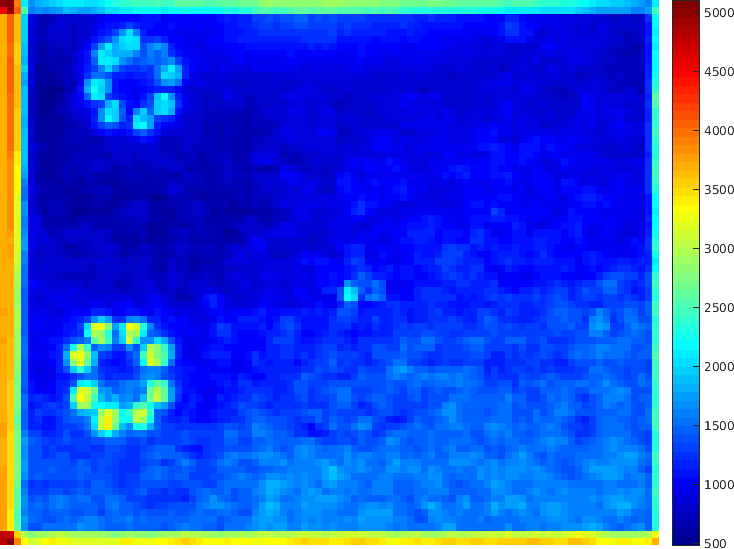}
        }
    \hfill
    \subfloat[][$\mathcal{T}_5$]{
        \centering
        \includegraphics[width=0.22\linewidth,height = 0.179\linewidth]{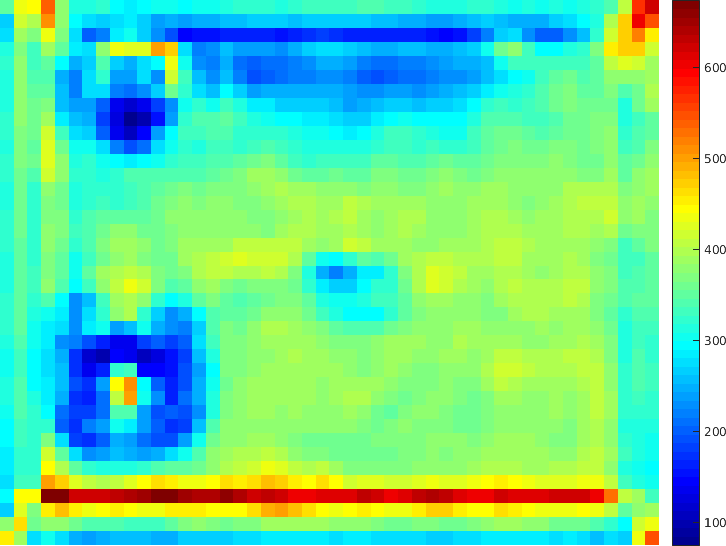}
        }
    \hfill
    \subfloat[][Detection results]{
        \centering
        \includegraphics[width=0.22\linewidth,height = 0.179\linewidth]{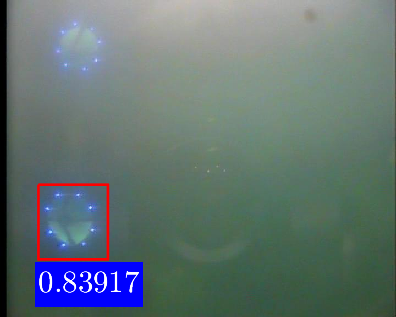}

    }
    \hfill\\
    \caption{Feature maps $\mathcal{T}_i$ computed at the $i^{th}$ layer $Conv_i,i = 1,2,5$, and the corresponding detection result of FasterRCNN for mirror images.}
    \label{fig:mirrorfrrcnnfeat}
\end{figure}

\begin{figure}
    \centering
    \includegraphics[width = \linewidth]{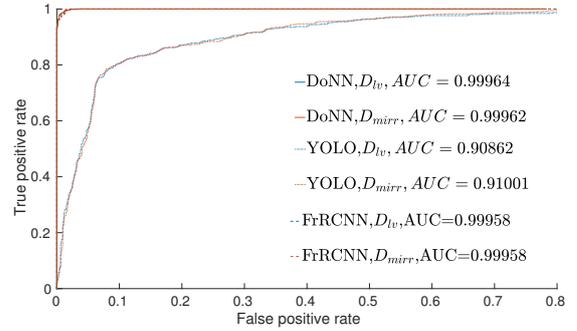}
    \caption{ROC curves of DoNN, YOLO, and Faster-RCNN computed for $D_{mirr}$ which contains mirror images.}
    
    \label{fig:ROCmirror}
\end{figure}

It is shown in Figure \ref{fig:ROCmirror} that all three CNN-based models are able to distinguish real docking stations from their mirror images with no performance degradation. This attributes the success to the learning ability of CNNs. It is also shown in Figure \ref{fig:ROCmirror} that DoNN slightly outperforms FasterRCNN. Figure \ref{fig:mirrorfrrcnnfeat} and \ref{fig:mirrordonnfeat} show feature maps of a natural mirror image generated by DoNN and FasterRCNN, respectively. It is shown in Figure \ref{fig:mirrorfrrcnnfeat} that activations computed for the real docking station computed in  $\mathcal{T}_1$ of FasterRCNN is stronger than those for the mirror docking station. This enables FasterRCNN to discriminate real docking stations from mirror ones. Figure \ref{fig:mirrordonnfeat} shows $\mathcal{T}_1$, $\mathcal{T}_5$ and confidence map $S$ of DoNN. Activation of the mirror docking station is almost as high as the real docking station in $\mathcal{T}_1$, $\mathcal{T}_5$ of DoNN. Even so, confidence map $S$ of DoNN contains only one highly confident grid which is the correct prediction. Next, we will analyze this phenomenon. We conjecture that DoNN can be used to estimate distribution of relative spatial locations between mirror and real docking stations. DoNN learns feature representations of mirror images of docking stations that appear more likely on the upper side rather than real docking stations. 

\begin{figure}
    \centering
    \captionsetup[subfigure]{justification=centering}
    \subfloat[][$\mathcal{T}_1$]{
        \centering
        \includegraphics[width=0.22\linewidth,height = 0.179\linewidth]{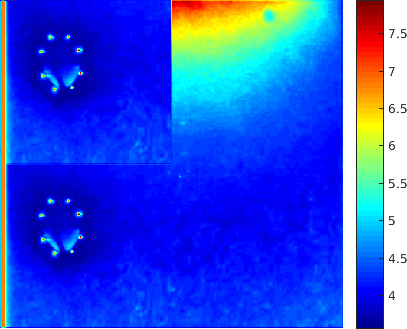}
        }
    \hfill
    \subfloat[][$\mathcal{T}_5$]{
        \centering
        \includegraphics[width=0.22\linewidth,height = 0.179\linewidth]{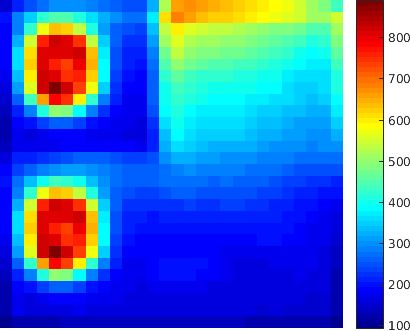}
        }
    \hfill
    \subfloat[][$S$]{
        \centering
        \includegraphics[width=0.22\linewidth,height = 0.179\linewidth]{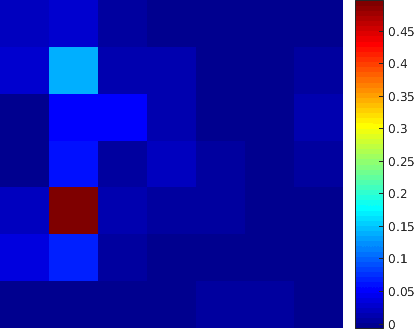}
        }
    \hfill
    \subfloat[][Detection results]{
        \centering
        \label{fig:mirrordonnfeat_replacereal_detectres_res}
        \includegraphics[width=0.22\linewidth,height = 0.179\linewidth]{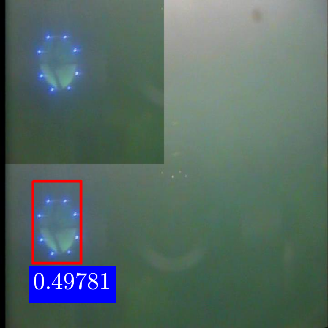}
    \label{fig:mirrordonnfeat_replacereal_detectres}
    }
    
    \hfill\\
    \caption{Feature maps $\mathcal{T}_i$ computed at the $i^{th}$ layer $Conv_i,i = 1,5$, confidence map, and the detection result of DoNN for replacement of the second quadrant of Figure \ref{fig:naturemirror} by its third quadrant.}
    \label{fig:mirrordonnfeat_replacereal}
\end{figure}

\begin{figure}
    \centering
    \captionsetup[subfigure]{justification=centering}
    \subfloat[][$\mathcal{T}_1$]{
        \centering
        \includegraphics[width=0.22\linewidth,height = 0.179\linewidth]{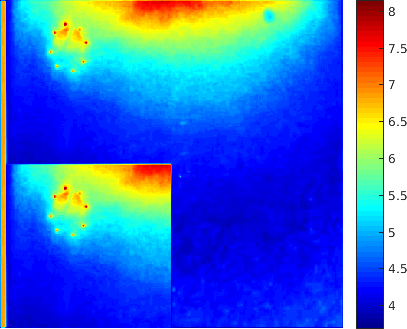}
        }
    \hfill
    \subfloat[][$\mathcal{T}_5$]{
        \centering
        \includegraphics[width=0.22\linewidth,height = 0.179\linewidth]{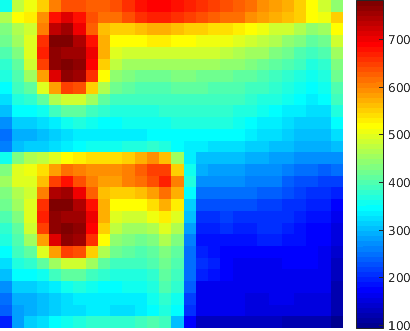}
        }
    \hfill
    \subfloat[][$S$]{
        \centering
        \includegraphics[width=0.22\linewidth,height = 0.179\linewidth]{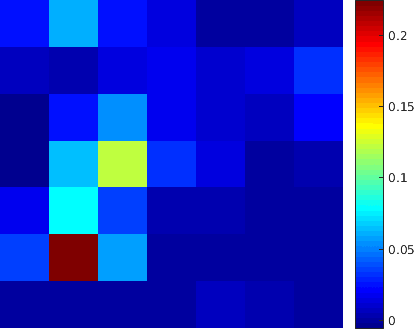}
        }
    \hfill
    \subfloat[][Detection results]{
        \centering
        \label{fig:mirrordonnfeat_replacemirror_res}
        \includegraphics[width=0.22\linewidth,height = 0.179\linewidth]{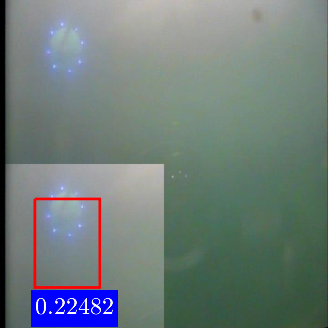}

    }
    \hfill\\
    \caption{Feature maps $\mathcal{T}_i$ computed at the $i^{th}$ layer $Conv_i,i = 1,5$, confidence map, and the detection result of DoNN for replacement of the third quadrant of Figure \ref{fig:naturemirror} by its second quadrant.}
    \label{fig:mirrordonnfeat_replacemirror}
\end{figure}

In the experimental analyses, we have to make sure that it is the relative spatial location or appearance that enables DoNN to distinguish real docking stations from mirror ones. To this end, we carry out two experiments. First, in Figure \ref{fig:naturemirror}, we replace the patch located in the second quadrant of the input image by the patch located in the third quadrant such that docking stations observed in the second and third quadrant share the same appearance, gaining Figure \ref{fig:mirrordonnfeat_replacereal_detectres}. Then, we input the simulated image into DoNN. The image obtained by replacement, its feature maps, confidence map and detection results are shown in Figure \ref{fig:mirrordonnfeat_replacereal}. Due to their same appearance, feature maps computed in the second and the third quadrant are exactly same. DoNN both assigns high confidence to two docking stations, but more to the lower one. In other words, DoNN tends to believe that the lower one is the real docking station. In addition, we compare the confidence of prediction before and after the replacement in Figure \ref{fig:naturemirror} and \ref{fig:mirrordonnfeat_replacereal_detectres}, respectively. The confidence falls from $0.74806$ computed before replacement to $0.49781$ computed after replacement. It reflects that the appearance contributes to the prediction of DoNN as well. Second, we replace the patch located in the third quadrant of the input image given in Figure \ref{fig:naturemirror} by the patch located in the second quadrant of itself, such that all docking stations observed in the image are mirror ones. Then, the obtained image is fed to DoNN. The result is shown in Figure \ref{fig:mirrordonnfeat_replacemirror}. DoNN still provides higher prediction score for the lower one compared to the upper one, although they are both identical mirror images. But the score in Figure \ref{fig:mirrordonnfeat_replacemirror_res} is less than the score in Figure \ref{fig:mirrordonnfeat_replacereal_detectres}. Therefore, we can conclude that DoNN has learned not only feature representations of appearance, but also the distribution of relative spatial locations between real and mirror docking stations.

\subsubsection{Comparison of performance of detection algorithms under non-uniform illumination}

Non-uniform illumination is commonly observed in real underwater images. In order to compare detection performance in an underwater environment as close as to real undersea environment of non-uniform illumination, we apply the non-uniform illumination drawn from a subset of the \textit{Fish4Knowledge} dataset (\textit{luminosity changes}) \citep{kavasidis2014innovative}, to our $D_{lv}$ dataset which was collected in an indoor water pool. The \textit{Fish4Knowledge} dataset was constructed using images captured in a real outdoor undersea environment, and used for detecting targets in noisy underwater environments. The subset \textit{luminosity changes} is specific for underwater luminosity changes. The new dataset obtained after applying undersea non-uniform illumination to $D_{lv}$ is denoted by $D_{nu}$. 

\begin{figure*}
    \hfill
    \includegraphics[width =\linewidth]{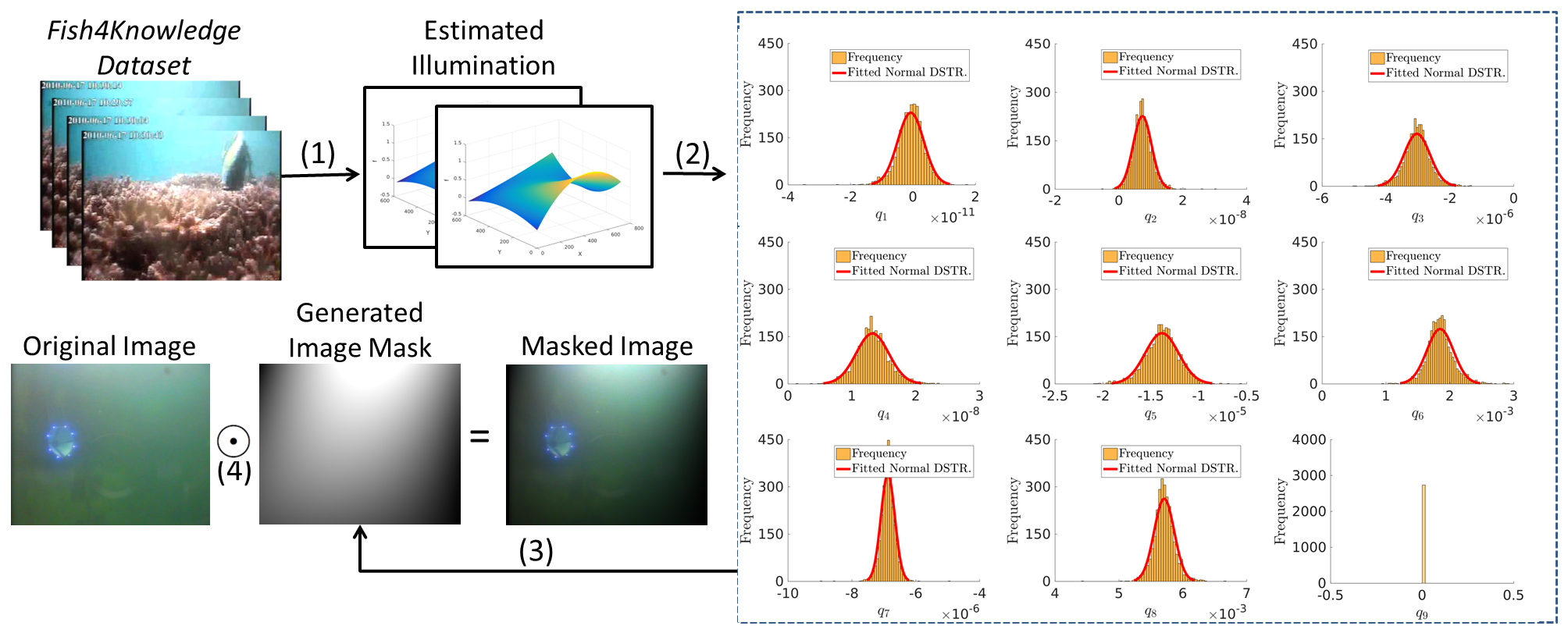}
    \caption{An illustration of our method proposed used for generating non-uniform illumination. (1) Estimation of undersea non-uniform illumination: Our proposed method first takes the value component (V) of images belonging to the \textit{Fish4Knowledge} dataset as input. Next, illumination values are estimated, and polynomial coefficients $Q$ are computed by fitting \eqref{eq_polyequation}. (2) Modeling distribution of coefficients: Distributions of coefficients are estimated by fitting the distributions to Gaussian models $\mathcal{G}$. (3) Generating new samples using non-uniform illumination: First, a set of samples $\hat{Q}$ is drawn from the Gaussian model $\mathcal{G}$ estimated in the step (2). Then, the polynomial function $f\left(x,y;\hat{Q}\right)$ is computed using the drawn samples $\hat{Q}$ in \eqref{eq_polyequation}. (4) Generated illumination values are applied to $D_{lv}$.}
    \label{fig:nonuni}
\end{figure*}

Polynomials are utilized for non-uniform illumination correction in transmission electron microscopy (TEM) images \citep{tasdizen2008non}. In our work, it is used in an opposite way, in order to generate non-uniform illumination. Details of non-uniform illumination generation procedure are shown in Figure \ref{fig:nonuni}, and explained as follows:
\begin{enumerate}
    \item Estimation of undersea non-uniform illumination: In order to estimate undersea non-uniform illumination, we fit a low-order ($m=n=2$) bivariate polynomial to the Value component (HSV color space) of every frame in the \textit{Fish4Knowledge} dataset. The polynomial is represented by
    \begin{equation}
    \label{eq_polyequation}
        \begin{aligned} 
        f\left(x,y;Q\right)=&q_1 x^n y^m+q_2 x^{\left(n-1\right)} y^m+\ldots+q_{n+1} y^m\\   
        & +\ldots +q_{n+2}x^ny^{\left(m-1\right)}+\\
        & +q_{n+3}x^{\left(n-1\right)}y^{\left(m-1\right)}+\\
        & +\ldots+q_{2\left(n+1\right)}y^{\left(m-1\right)}+\ldots+  \\ &q_{m\left(n+1\right)+1}x^n+q_{m\left(n+1\right)+2}x^{\left(n-1\right)}\\
        &+\ldots+q_{\left(n+1\right)\left(m+1\right)},
        \end{aligned} 
    \end{equation}
where $Q$ stands for the set of parameters. After fitting, $Q_i=(q_{i,1},\ldots,q_{i,j},\ldots,q_{i,(n+1)(m+1)})\in{\mathbb{R}}^{{(n+1)(m+1)}}$ is obtained for the $i^{th}$ frame.
    
    \item Modeling distribution of coefficients: Suppose that $q_j (j = (1,2, \ldots , (n+1)(m+1)))$ are independent random variables. Then, $q_{i,j}$ is viewed as the $i^{th}$ sample drawn from the distribution of $q_j$. The distribution of $q_{j}$ is shown in the histogram given in Figure \ref{fig:nonuni}. Obviously, the distribution of  $q_j$ forms a Gaussian shape. It is fitted by a Gaussian Distribution $\mathcal{G}_j(\mu_j ,\sigma_j^{2})$, as depicted in Figure \ref{fig:nonuni}.
    
    \item Generation of new underwater images using non-uniform illumination: Generation of new illumination involves drawing samples from the obtained distribution $\mathcal{G}_j(\mu_j ,\sigma_j^{2})(j = (1, \ldots , (n+1)(m+1)))$, and an evaluation of polynomial function \eqref{eq_polyequation}. For each image $I_{o} \in D_{lv}$, a set of samples $\hat{Q}$ is drawn from $\mathcal{G}_j(\mu_j ,\sigma_j^{2})$, and a new non-uniform illumination $I_\delta$ is generated.
    
    \item Applying generated illumination to $D_{lv}$: Since an image can be modeled as a multiplicative effect \citep{tasdizen2008non}, the value component (V) $I_{nuv}$ of newly generated image $I_{nu}$ with non-uniform illumination can be obtained by
    \begin{equation}
        I_{nuv} = I_{ov} \odot I_{\delta}
    \end{equation}
where $I_{ov}$ is the value (V) component of $I_o$ in HSV color space, and $\odot$ indicates pixel-wise multiplication.
\end{enumerate}

\begin{figure}
    \centering
    \includegraphics[width = \linewidth]{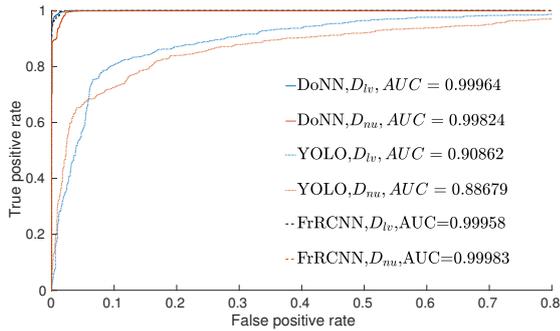}
    \caption{ROC curves of DoNN, YOLO, and Faster-RCNN obtained for undersea nonuniform illumination.}
    \label{fig:ROCnonuniform}
\end{figure}

\begin{figure}
    \centering
    \captionsetup[subfigure]{justification=centering}
    \subfloat[][$\mathcal{T}_1$]{
        \centering
        \includegraphics[width=0.22\linewidth,height = 0.179\linewidth]{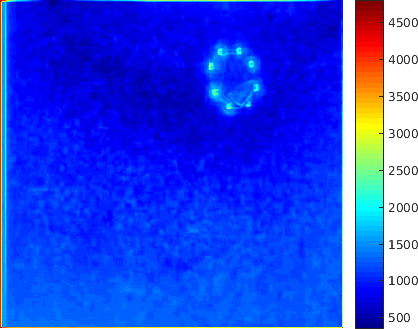}
        }
    \hfill
    \subfloat[][$\mathcal{T}_2$]{
        \centering
        \includegraphics[width=0.22\linewidth,height = 0.179\linewidth]{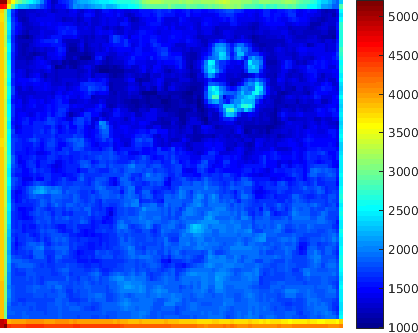}
        }
    \hfill
    \subfloat[][$\mathcal{T}_5$]{
        \centering
        \includegraphics[width=0.22\linewidth,height = 0.179\linewidth]{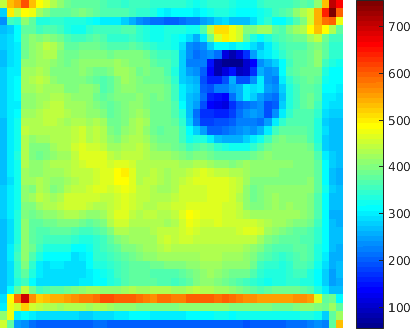}
        }
    \hfill
    \subfloat[][Detection results]{
        \centering
        \includegraphics[width=0.22\linewidth,height = 0.179\linewidth]{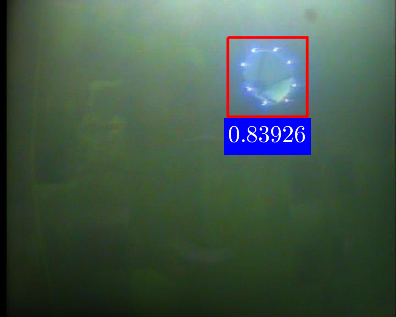}

    }
    \hfill\\
    \caption{Feature maps $\mathcal{T}_i$ computed at the $i^{th}$ layer $Conv_i,i = 1,2,5$, and detection result of FasterRCNN obtained for non-uniform illumination.}
    \label{fig:nonuniformfrrcnnfeat}
\end{figure}

\begin{figure}
    \centering
    \captionsetup[subfigure]{justification=centering}
    \subfloat[][$\mathcal{T}_1$]{
        \centering
        \includegraphics[width=0.22\linewidth,height = 0.179\linewidth]{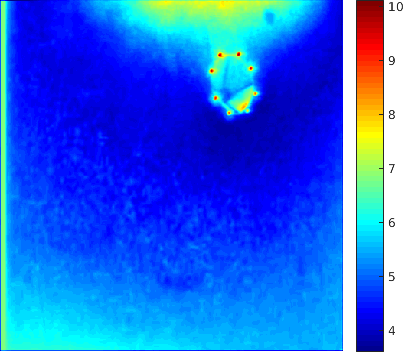}
        }
    \hfill
    \subfloat[][$\mathcal{T}_5$]{
        \centering
        \includegraphics[width=0.22\linewidth,height = 0.179\linewidth]{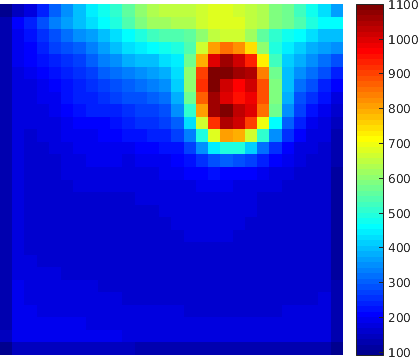}
        }
    \hfill
    \subfloat[][$S$]{
        \centering
        \includegraphics[width=0.22\linewidth,height = 0.179\linewidth]{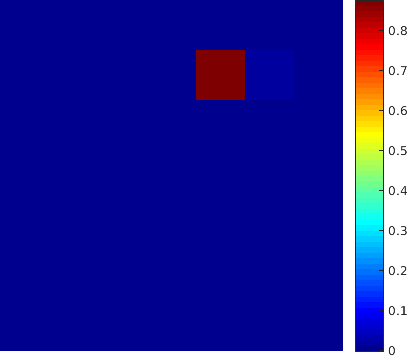}
        }
    \hfill
    \subfloat[][Detection results]{
        \centering
        \includegraphics[width=0.22\linewidth,height = 0.179\linewidth]{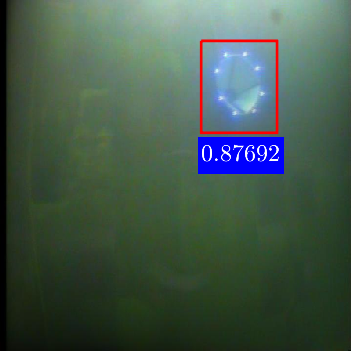}

    }
    \hfill\\
    \caption{Feature maps $\mathcal{T}_i$ computed at the $i^{th}$ layer $Conv_i,i = 1,5$, confidence map, and detection result of DoNN obtained for non-uniform illumination.}
    \label{fig:nonuniformdonnfeat}
\end{figure}

Figure \ref{fig:ROCnonuniform} shows ROC curves and the associated AUCs of three models. It is observed that both FasterRCNN and DoNN are robust to non-uniform illumination. Their corresponding feature maps are shown in Figure \ref{fig:nonuniformfrrcnnfeat} and \ref{fig:featmapfrrcnnwithouttrans}. The feature map $\mathcal{T}_1$ is generated as if there is no non-uniform illumination. Comparing $\mathcal{T}_5$ without and with non-uniform transformation in Figure \ref{fig:nonuniformdonnfeat} and \ref{fig:featmapdonnwithouttrans}, we observed that the maps $\mathcal{T}_5$ are almost identical. Therefore, non-uniform illumination does not affect the final prediction. The notation $\odot$ indicates pixel-wise multiplication.

\subsubsection{Comparison of performance of detection algorithms under noisy luminaries} 

\begin{figure}
    \centering
    \includegraphics[width = \linewidth]{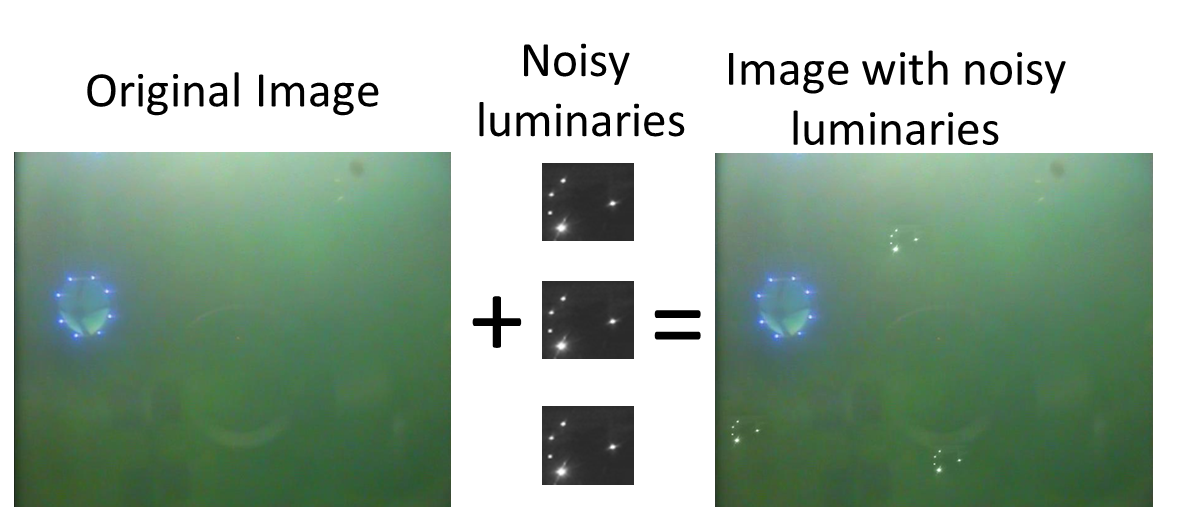}
    \caption{The merging process of synthetic noisy luminaries images. Three noisy luminaries are merged at random locations to the original image.}
    \label{fig:mergingnoisy}
\end{figure}

Similar to synthesis of mirror images, noisy luminaries which are similar to docking stations in terms of brightness and structure are also merged to every foreground image in $D_{lv}$ by using image editing \citep{perez2003poisson}, forming a dataset $D_{nlum}$. The noisy luminaries image is pinned three times on three random locations of the original images. We show the merging process of noisy luminaries in Figure \ref{fig:mergingnoisy}.

\begin{figure}
    \centering
    \includegraphics[width = \linewidth]{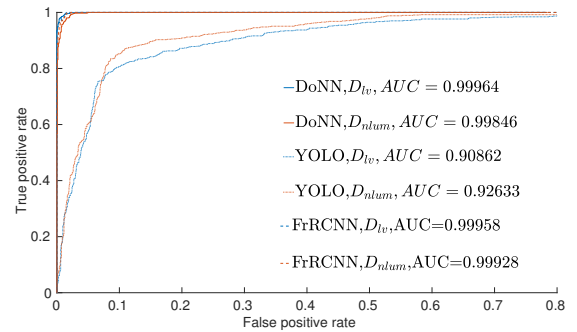}
    \caption{ROC curves of DoNN, and YOLO, Faster-RCNN computed for noisy luminaries.}
    \label{fig:ROCnoisylumin}
\end{figure}

\begin{figure}
    \centering
    \captionsetup[subfigure]{justification=centering}
    \subfloat[][$\mathcal{T}_1$]{
        \centering
        \includegraphics[width=0.22\linewidth,height = 0.179\linewidth]{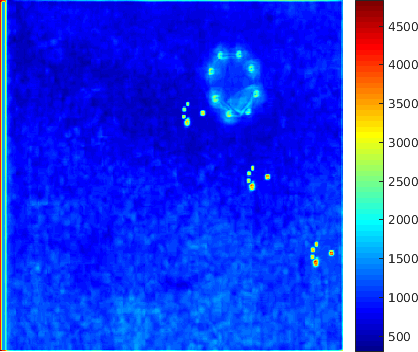}
        }
    \hfill
    \subfloat[][$\mathcal{T}_2$]{
        \centering
        \includegraphics[width=0.22\linewidth,height = 0.179\linewidth]{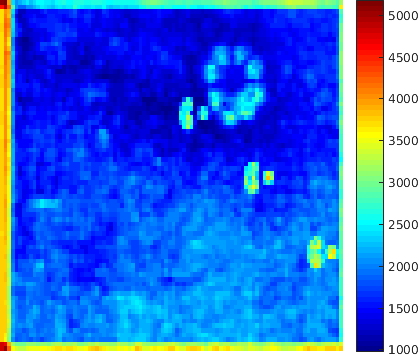}
        }
    \hfill
    \subfloat[][$\mathcal{T}_5$]{
        \centering
        \includegraphics[width=0.22\linewidth,height = 0.179\linewidth]{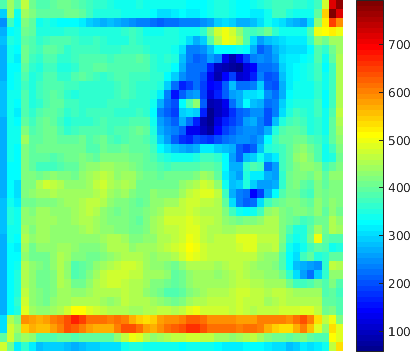}
        }
    \hfill
    \subfloat[][Detection results]{
        \centering
        \includegraphics[width=0.22\linewidth,height = 0.179\linewidth]{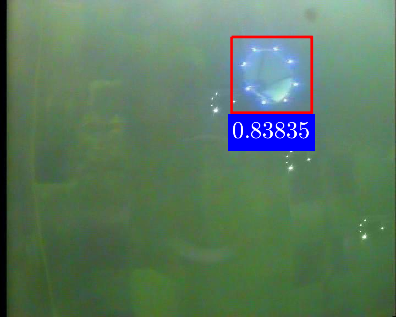}

    }
    \hfill\\
    \caption{Feature maps $\mathcal{T}_i$ computed at the $i^{th}$ layer $Conv_i,i = 1,2,5$, and the detection result of FasterRCNN obtained for noisy luminaries.}
    \label{fig:noisyluminfrrcnnfeat}
\end{figure}

\begin{figure}
    \centering
    \captionsetup[subfigure]{justification=centering}
    \subfloat[][$\mathcal{T}_1$]{
        \centering
        \includegraphics[width=0.22\linewidth,height = 0.179\linewidth]{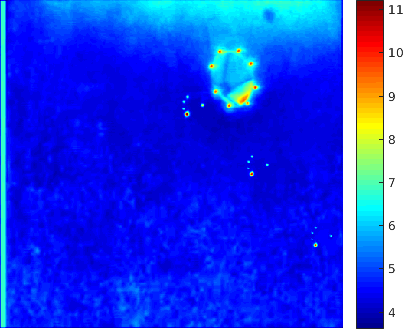}
        }
    \hfill
    \subfloat[][$\mathcal{T}_5$]{
        \centering
        \includegraphics[width=0.22\linewidth,height = 0.179\linewidth]{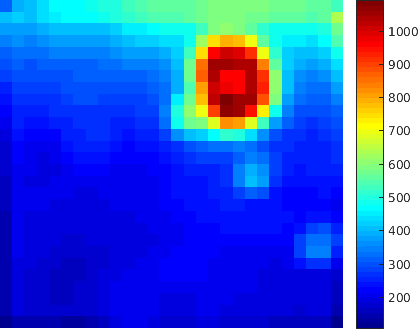}
        }
    \hfill
    \subfloat[][$S$]{
        \centering
        \includegraphics[width=0.22\linewidth,height = 0.179\linewidth]{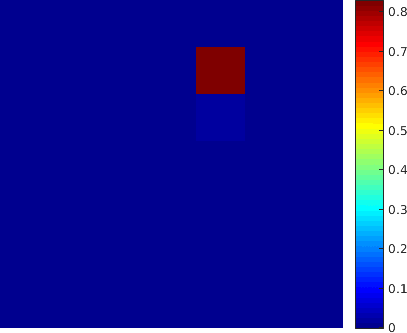}
        }
    \hfill
    \subfloat[][Detection results]{
        \centering
        \includegraphics[width=0.22\linewidth,height = 0.179\linewidth]{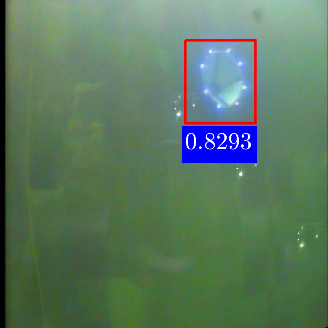}

    }
    \hfill\\
    \caption{Feature maps $\mathcal{T}_i$ computed at the $i^{th}$ layer $Conv_i,i = 1,5$, confidence map, and the detection result of DoNN obtained for noisy luminaries.}
    \label{fig:noisylumindonnfeat}
\end{figure}

We provide ROC curves and the corresponding AUCs of three models in Figure \ref{fig:ROCnoisylumin}. The negative effect of noisy luminaries on three models is very tiny, although noisy luminaries are as bright as docking stations, showing strong robustness of CNN based methods to noisy luminaries. DoNN achieves an acceptable performance where AUC is $0.99846$ in the presence of random synthetic noisy luminaries. It lags behind FasterRCNN whose AUC is $0.99928$. Figure \ref{fig:noisyluminfrrcnnfeat} shows feature maps and the detection result of a sample image predicted by FasterRCNN. The region of noisy luminaries is activated stronger than the docking station region in the feature map $\mathcal{T}_1$ of FasterRCNN. However, the region of the docking station overwhelms in the feature map $\mathcal{T}_5$ of FasterRCNN. Figure \ref{fig:noisylumindonnfeat} shows feature maps, confidence maps and the detection result of a sample image predicted by DoNN. Activation of noisy luminaries is as strong as the docking station in the map $\mathcal{T}_1$ of DoNN, but vanishes in the map $\mathcal{T}_5$. We conjecture that it is the learned feature representation of spatial structural patterns that enables FasterRCNN and DoNN to avoid suffering from noisy luminaries. 

\subsubsection{Comparison of performance of detection algorithms for nonlinear combination of deformations} So far we have analyzed the performance of three detection methods for different deformations. Among them, hue shift and contrast shift affect the performance difference between DoNN and FasterRCNN mostly as shown above. Hence, the performance of algorithms is explored for images that are deformed by combination of hue and contrast shift in this section. We sample $\lambda_h$ and the $\gamma$ in contrast deformation from $\{ 0.5, 0.7, 0.9\}$ and $\{  0.2,  0.4,  0.6,  0.8,1,1.2,1.4,1.6,1.8,2\}$, respectively. The performance of DoNN, FasterRCNN and YOLO computed for combination of transformations is shown in Table \ref{tab:auccombine}.  DoNN outperforms FasterRCNN by a large margin in all cases except two cases, and slightly lags behind  for $\lambda_h = 0.9,\gamma = {1.8,2.0}$. Therefore we draw the conclusion that DoNN works better than FasterRCNN and YOLO in non-stationary underwater environments generally.

\begin{table}[h!]
    \centering
    \caption{AUCs of DoNN, FasterRCNN and YOLO computed for combination of hue and contrast shift.}
\begin{tabular}{c|c|c|c}
\hline
Deformation & DoNN & FrRCNN & YOLO \\
\hline
$\lambda_h = 0.5$,$\gamma = 0.2$	 & \textbf{0.82383}	& 0.00000	& 0.00000 \\
$\lambda_h = 0.5$, $\gamma = 0.4$	 & \textbf{0.91931}	& 0.00000	& 0.79135 \\
$\lambda_h = 0.5$, $\gamma = 0.6$	 & \textbf{0.93689}	& 0.83695	& 0.85912 \\
$\lambda_h = 0.5$, $\gamma = 0.8$	 & \textbf{0.95685}	& 0.80759	& 0.81419 \\
$\lambda_h = 0.5$, $\gamma = 1.0$	 & \textbf{0.97405}	& 0.80774	& 0.86540 \\
$\lambda_h = 0.5$, $\gamma = 1.2$	 & \textbf{0.93893}	& 0.86421	& 0.86557 \\
$\lambda_h = 0.5$, $\gamma = 1.4$	 & \textbf{0.92623}	& 0.86660	& 0.86074 \\
$\lambda_h = 0.5$, $\gamma = 1.6$	 & \textbf{0.92196}	& 0.87586	& 0.87612 \\
$\lambda_h = 0.5$, $\gamma = 1.8$	 & \textbf{0.91799}	& 0.87833	& 0.89935 \\
$\lambda_h = 0.5$, $\gamma = 2.0$	 & \textbf{0.91227}	& 0.86694	& 0.90924 \\
$\lambda_h = 0.7$, $\gamma = 0.2$	 & \textbf{0.85802}	& 0.00000	& 0.72217 \\
$\lambda_h = 0.7$, $\gamma = 0.4$	 & \textbf{0.94511}	& 0.00000	& 0.82594 \\
$\lambda_h = 0.7$, $\gamma = 0.6$	 & \textbf{0.99237}	& 0.85504	& 0.86705 \\
$\lambda_h = 0.7$, $\gamma = 0.8$	 & \textbf{0.99381}	& 0.92334	& 0.88413 \\
$\lambda_h = 0.7$, $\gamma = 1.0$	 & \textbf{0.99824}	& 0.97628	& 0.89211 \\
$\lambda_h = 0.7$, $\gamma = 1.2$	 & \textbf{0.99776}	& 0.96343	& 0.88061 \\
$\lambda_h = 0.7$, $\gamma = 1.4$	 & \textbf{0.99740}	& 0.95966	& 0.89431 \\
$\lambda_h = 0.7$, $\gamma = 1.6$	 & \textbf{0.99743}	& 0.94142	& 0.91749 \\
$\lambda_h = 0.7$, $\gamma = 1.8$	 & \textbf{0.99530}	& 0.93769	& 0.93151 \\
$\lambda_h = 0.7$, $\gamma = 2.0$	 & \textbf{0.99276}	& 0.93607	& 0.93536 \\
$\lambda_h = 0.9$, $\gamma = 0.2$	 & 0.83497	& 0.00000	& \textbf{0.87129} \\
$\lambda_h = 0.9$, $\gamma = 0.4$	 & \textbf{0.99431}	& 0.92728	& 0.87093 \\
$\lambda_h = 0.9$, $\gamma = 0.6$	 & \textbf{0.99829}	& 0.96821	& 0.91910 \\
$\lambda_h = 0.9$, $\gamma = 0.8$	 & \textbf{0.99862}	& 0.99262	& 0.92616 \\
$\lambda_h = 0.9$, $\gamma = 1.0$	 & \textbf{0.99956}	& 0.99888	& 0.93559 \\
$\lambda_h = 0.9$, $\gamma = 1.2$	 & \textbf{0.99976}	& 0.99871	& 0.91672 \\
$\lambda_h = 0.9$, $\gamma = 1.4$	 & \textbf{0.99930}	& 0.99902	& 0.92118 \\
$\lambda_h = 0.9$, $\gamma = 1.6$	 & \textbf{0.99918}	& 0.99837	& 0.92003 \\
$\lambda_h = 0.9$, $\gamma = 1.8$	 & 0.99726	& \textbf{0.99782}	& 0.92106 \\
$\lambda_h = 0.9$, $\gamma = 2.0$	 & 0.99658	& \textbf{0.99678}	& 0.92229 \\
\hline
\end{tabular}
    \label{tab:auccombine}
\end{table}


In conclusion, DoNN is quite robust to deformation of images observed in real-world underwater environments. This is attributed to its success for learning of feature representations and estimation of relative spatial distributions of spatial structural patterns. DoNN achieves acceptable performance in cases of extreme hue shift and extreme contrast shift. We can conclude that DoNN outperforms YOLO and FasterRCNN in terms of credibility and robustness.  

\subsection{Analysis of the pose estimation algorithm}
In this section, we first show results on ground experiments to assess the accuracy and robustness of our pose estimation method to noise. Next, we examine the effectiveness of our pose estimation method by underwater docking experiments. Finally, we show experimental results using an integrated underwater docking framework.

\subsubsection{Ground experiments}
\begin{figure}[h!]
\centering
\captionsetup[subfigure]{justification=centering}
    \subfloat[][]{
    \centering
    \label{fig:infrasa}
    \includegraphics[width= 0.4\linewidth]{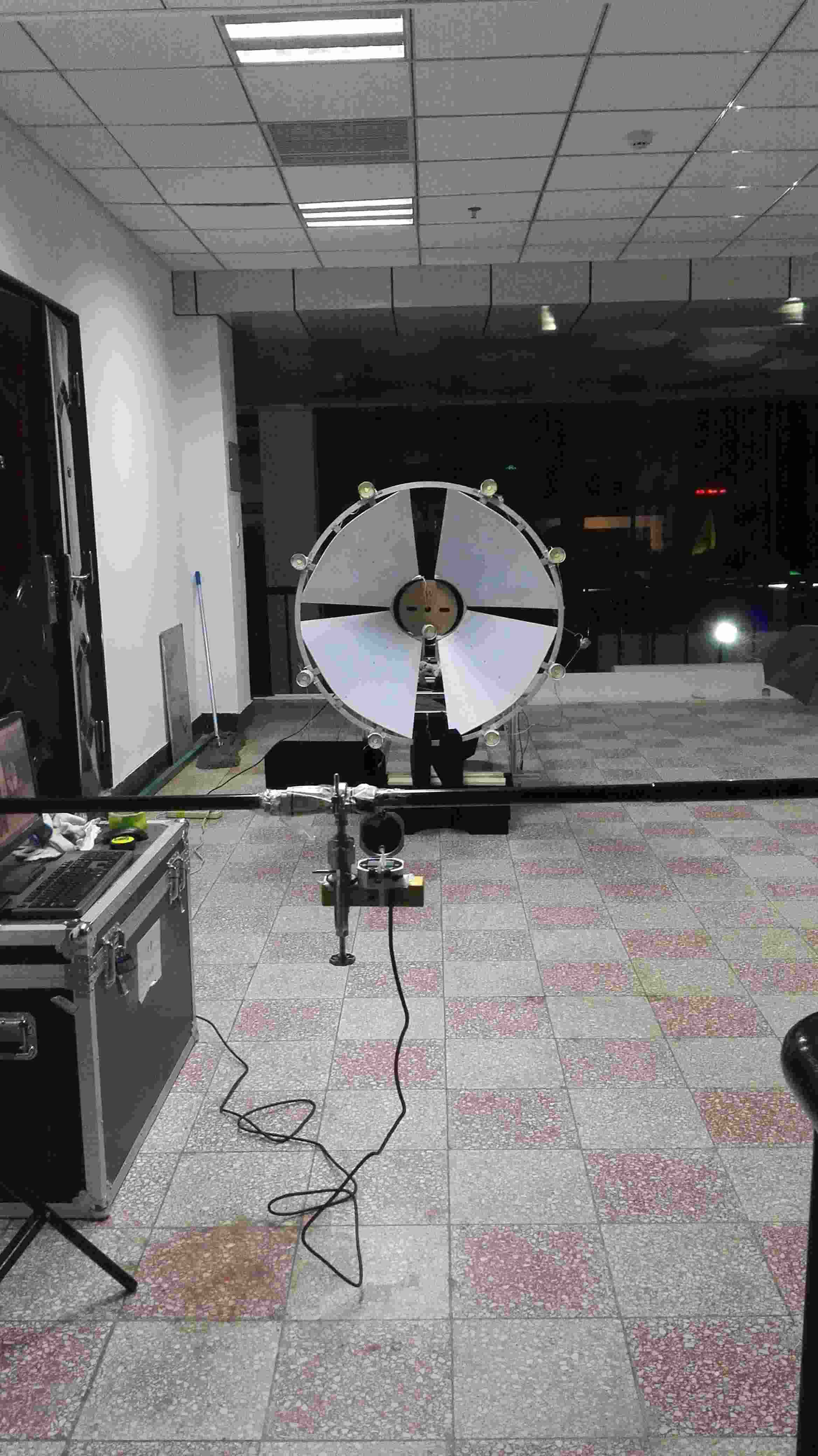}
    }
    \subfloat[][]{
    \centering
    \label{fig:infrasb}
    \includegraphics[width= 0.4\linewidth]{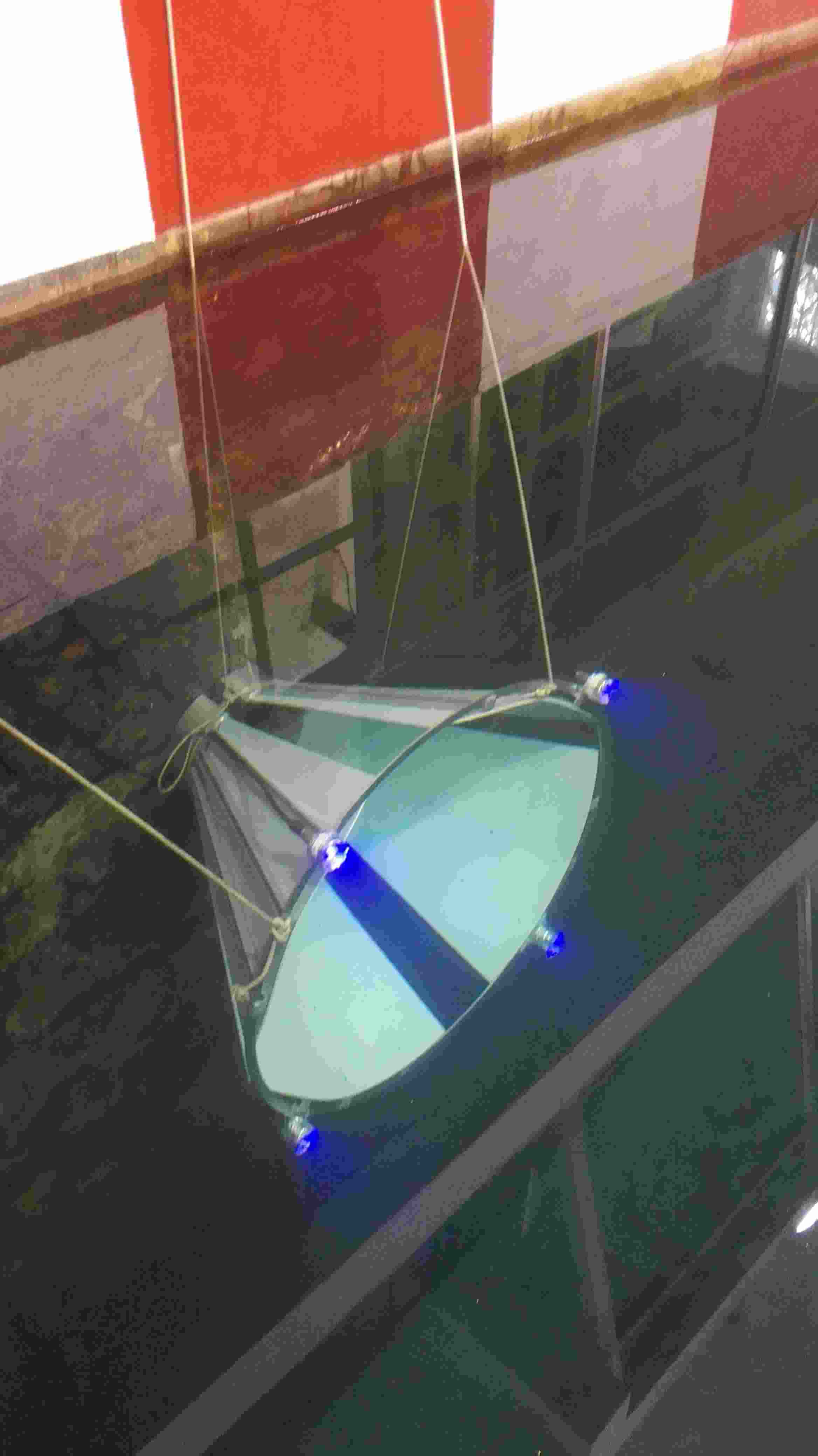}
    }
\caption{Infrastructures used for ground experiments and the underwater docking station located in water.}
\label{fig:infras}
\end{figure}

Since it is not feasible to obtain the underwater ground truth of the relative pose, we validate the accuracy and robustness of our pose estimation method to noise using ground experiments. Specifically in the experiments, we moved the camera around the docking station to capture its images with various pose from distance 3m -- 5m with a stationary docking station, as shown in Figure \ref{fig:infrasa}. Then, we manually labeled coordinates of the landmarks in 2D images. Finally, poses were computed by using our pose estimation method. In order to validate the robustness in presence of noise, we added different levels of Gaussian noise to the manually obtained coordinates of the landmarks. We provided mean estimation results averaged over 1000 trials for each noise level. The ground truth of orientation is obtained using an electronic compass which rigidly bounds together with the camera. In our ground experiments, half of eight landmarks are white LED, and the other half are blue LED. Same results can be obtained as the configuration of all blue LED, owing to manually labeled image points in ground experiments.

Table \ref{tab:groundtest} shows the pose estimation results and their ground truth. Three levels of Gaussian noise are added: standard deviation $\sigma = 0$, $\sigma = 3$ and $\sigma = 5$. Without adding any noise, the average error of predicted orientation and position are $1.970^{\circ}$ and $5.927$ mm respectively. As $\sigma$ increases to $3$, orientation error and position error increase by $0.096$ degree and $0.708$ millimeter respectively. As noise level becomes $\sigma = 5$, orientation error and position error increase by $0.383$ degree and $3.505$ millimeter, comparing to the case of $\sigma = 0$. Therefore, we can draw the conclusion that our pose estimation algorithm is accurate and robust in the presence of noise. We will show that it is also fast in Section \ref{sect:integrated}.

\begin{figure}[h!]
\captionsetup[subfigure]{justification=centering}
    \subfloat[][]{
    \centering
    \includegraphics[width= 0.23\linewidth]{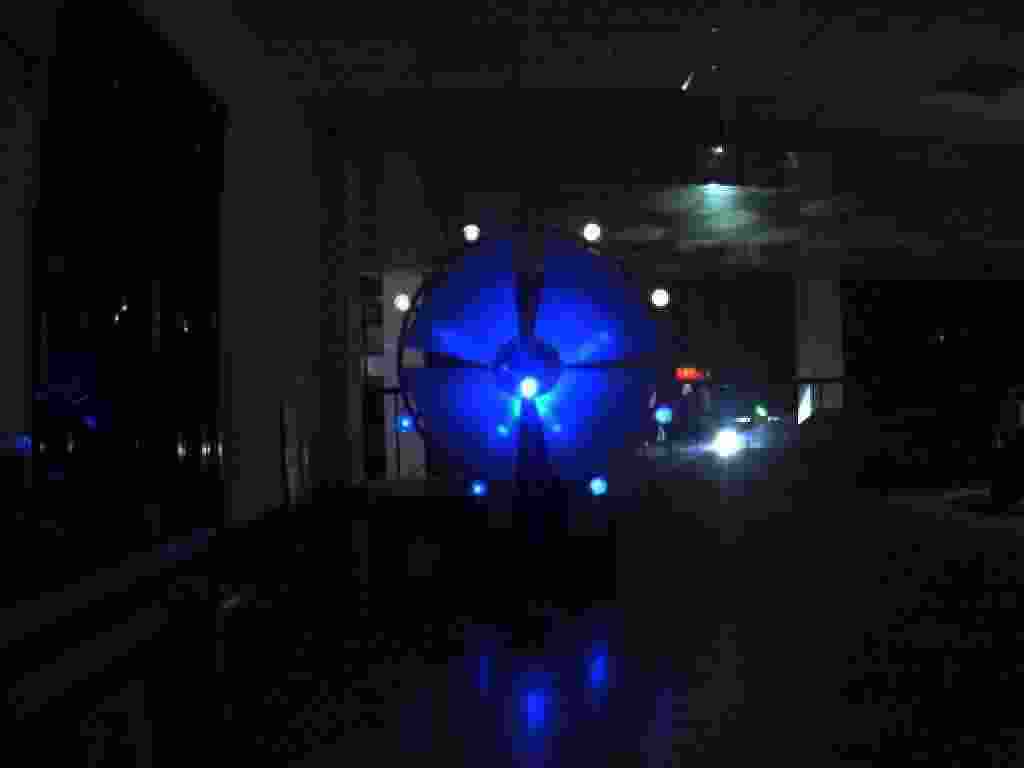}
    }
    \subfloat[][]{
    \centering
    \includegraphics[width= 0.23\linewidth]{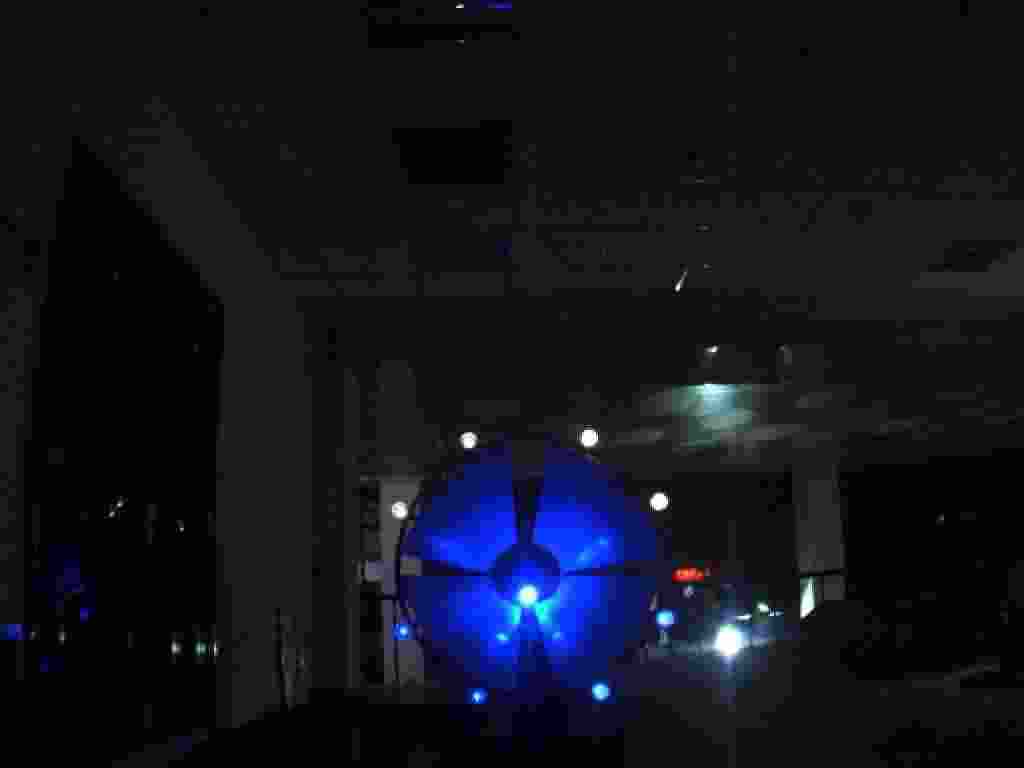}
    }
    \subfloat[][]{
    \centering
    \includegraphics[width= 0.23\linewidth]{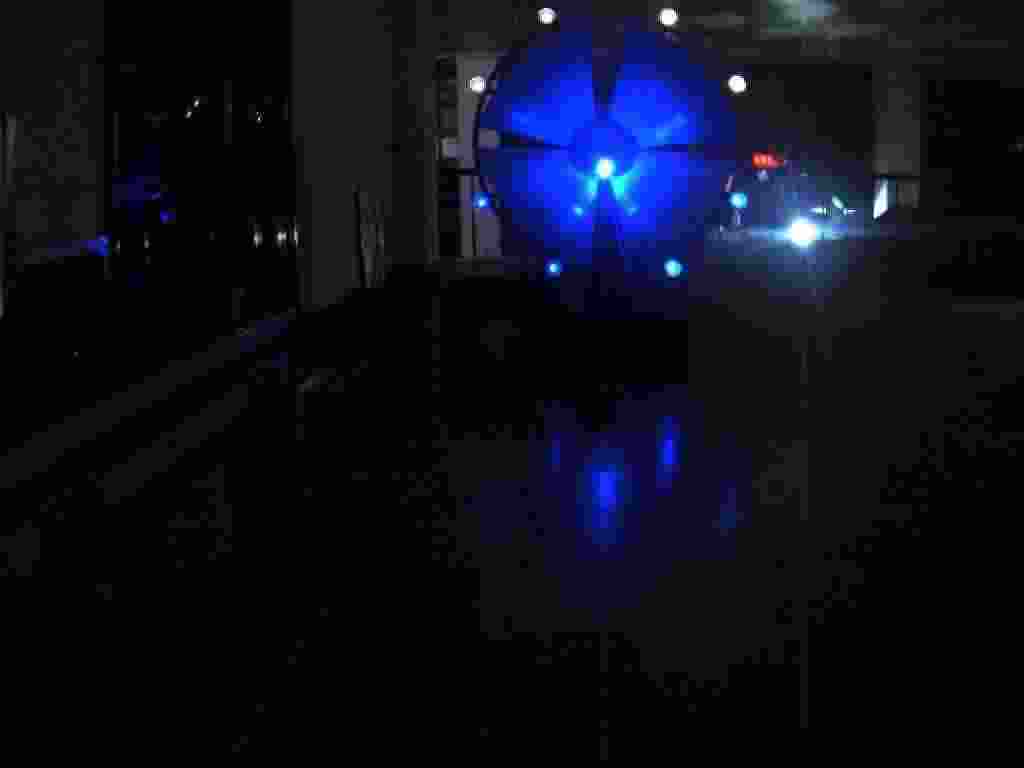}
    }
    \subfloat[][]{
    \centering
    \includegraphics[width= 0.23\linewidth]{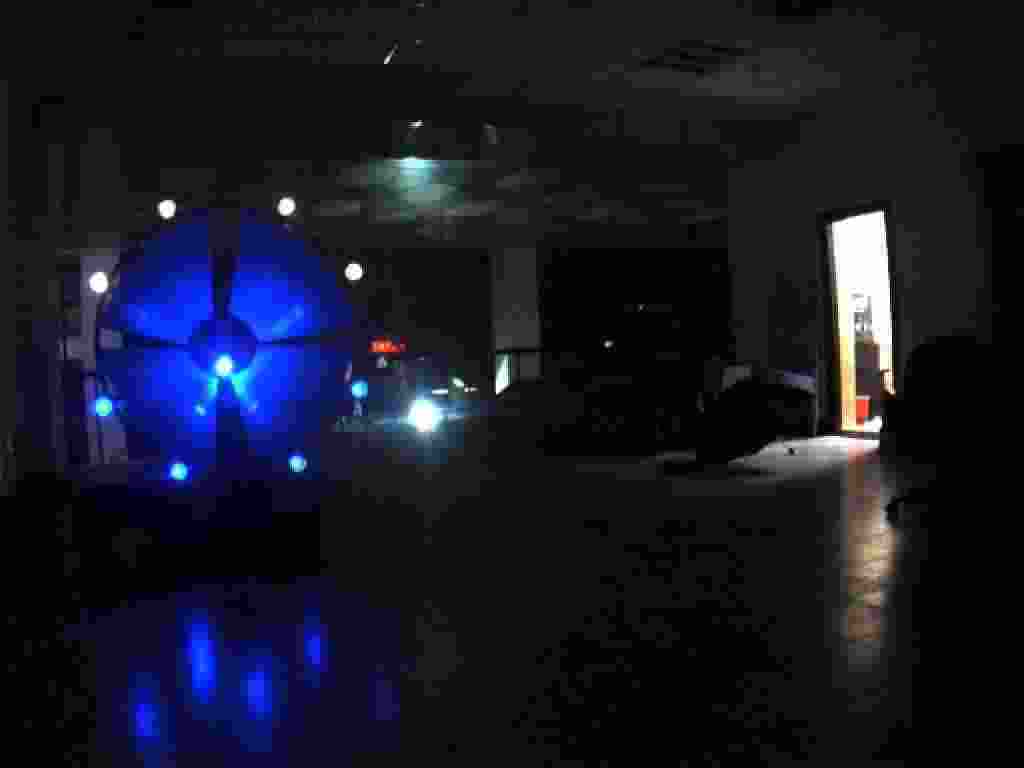}
    }\\
    \subfloat[][]{
    \centering
    \includegraphics[width= 0.23\linewidth]{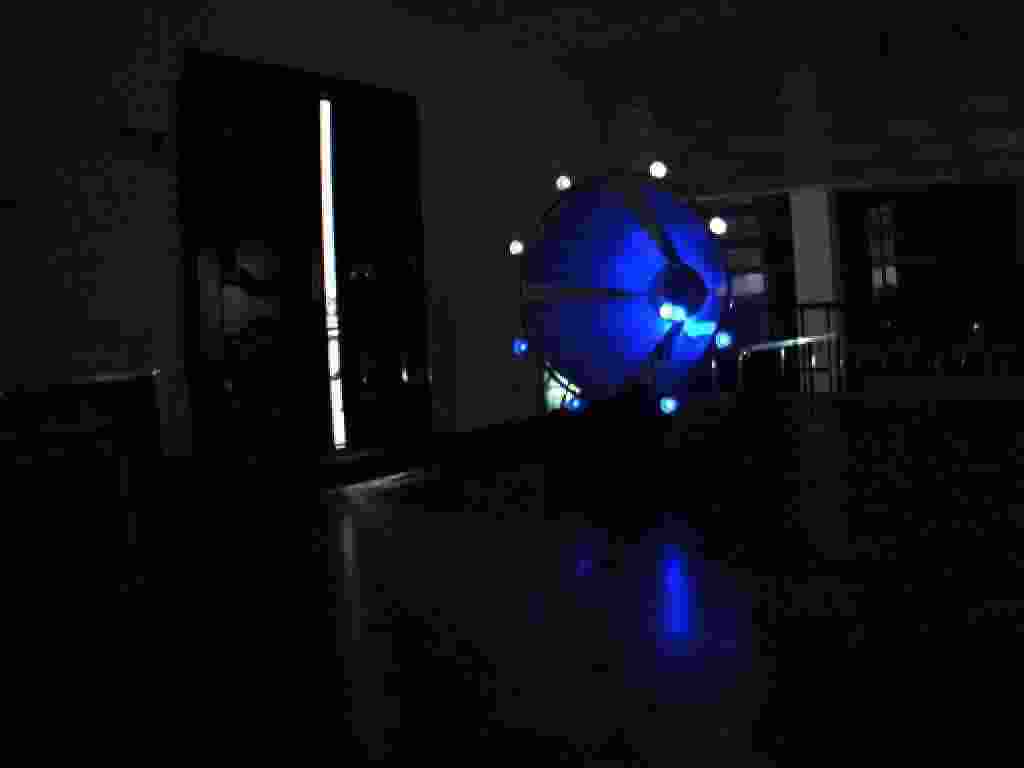}
    }
    \subfloat[][]{
    \centering
    \includegraphics[width= 0.23\linewidth]{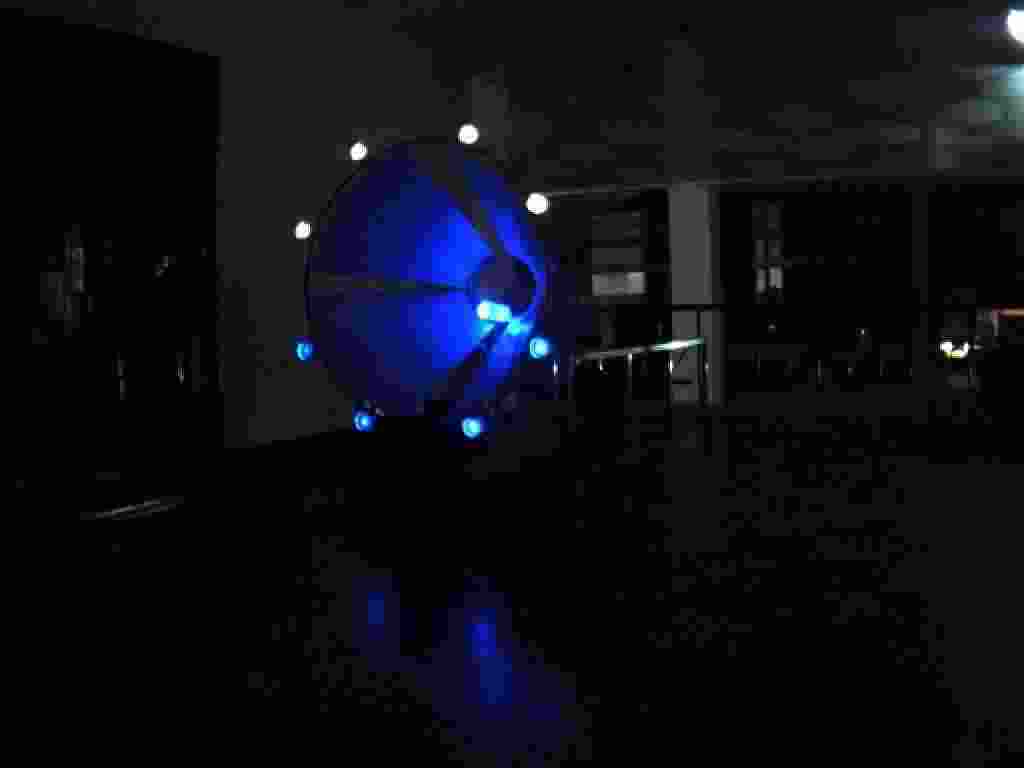}
    }
    \subfloat[][]{
    \centering
    \includegraphics[width= 0.23\linewidth]{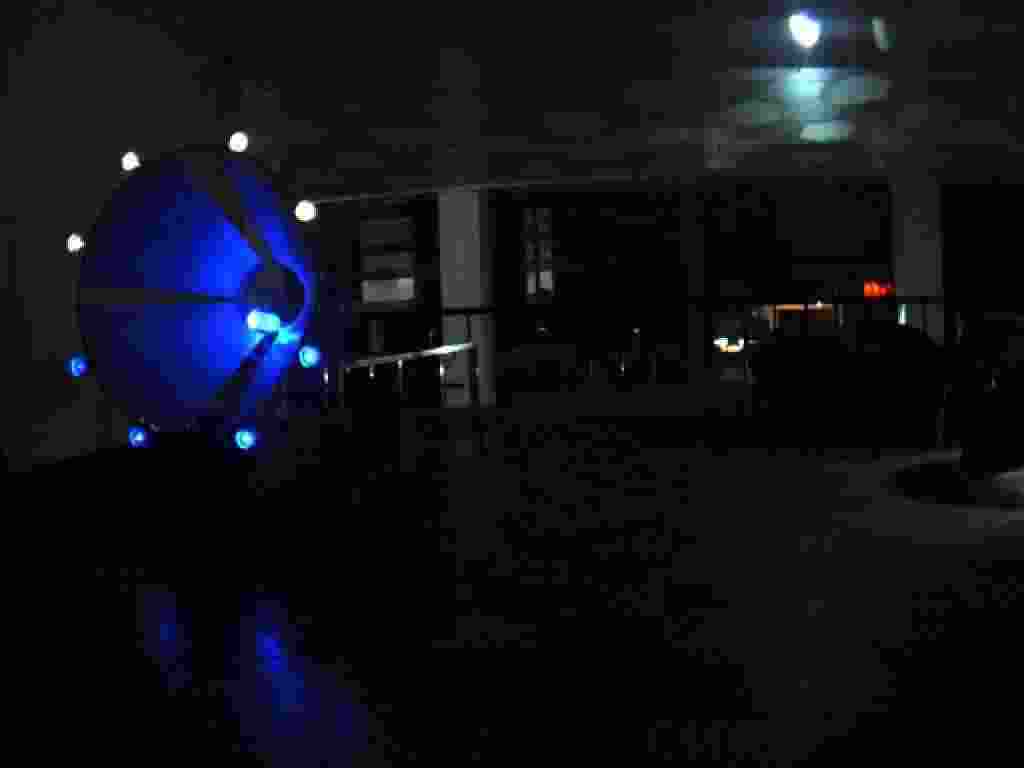}
    }
    \subfloat[][]{
    \centering
    \includegraphics[width= 0.23\linewidth]{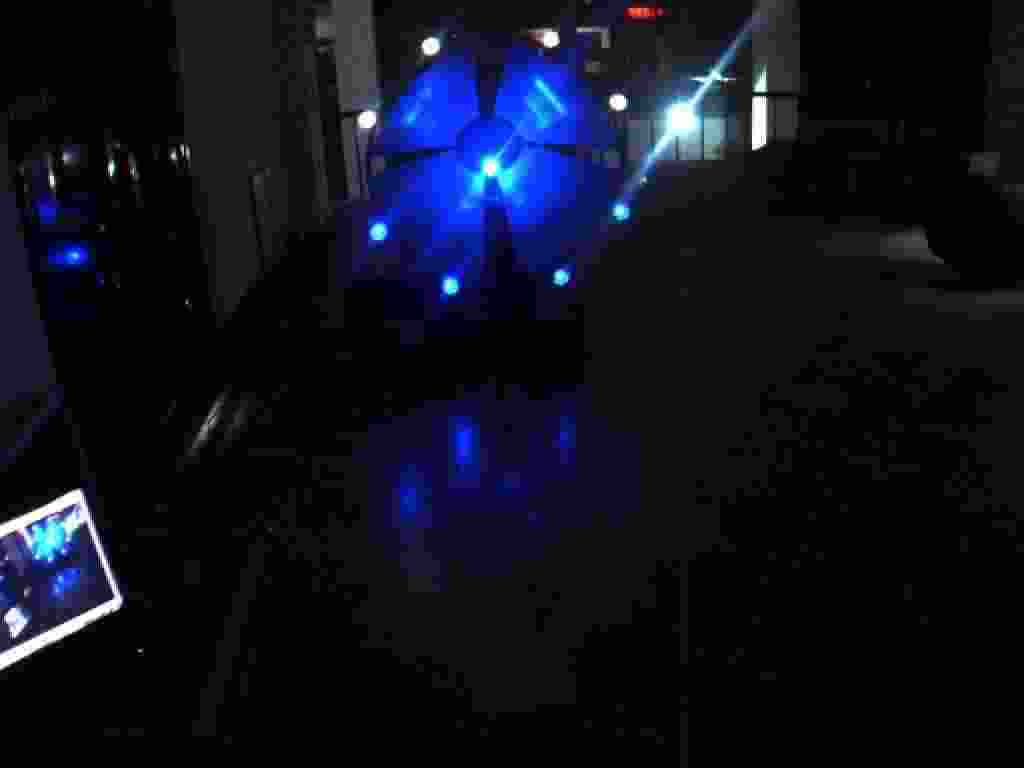}
    }
\caption{Ground test. Images were taken from different positions and orientations while keeping the docking station still.}
\label{fig:groundimages}
\end{figure}
\begin{table*}[ht]
    \centering
    \caption{Ground pose estimation performance. This table shows results corresponding to the figures given in Figure \ref{fig:groundimages}.}
\begin{tabular}{m{1.2cm}|c|c|c|c|c|c|c|c}
\hline
\multirow{2}{1em}{No.} & \multicolumn{2}{c|}{Ground Truth}& \multicolumn{2}{c|}{$\sigma = 0$} & \multicolumn{2}{c|}{$\sigma = 3$}& \multicolumn{2}{c}{$\sigma = 5$}\\
\cline{2-9}
& Orien.(deg) & Pos.(mm) & Orien.(deg) & Pos.(mm)& Orien.(deg) & Pos.(mm)& Orien.(deg) & Pos.(mm)\\
\hline
\multirow{3}{1em}{(a)} & Yaw:0.0         & $x$: 99.2      & 1.231          & 10.366    &0.960   &99.916   &0.293   &98.862\\
                     & Pitch:4.5       & $y$: -73.6     & 2.809          & -71.517   &1.932   &-72.216  &2.069   &-71.717\\
                     & Roll:-0.8       & $z$: 3686.5    & -1.273         & 3687.955  &-1.258  &3679.886 &-1.244  &3666.954\\
\hline
\multirow{3}{1em}{(b)} & Yaw:6.8         & $x$: 98.1      & 2.115          & 96.559    &1.196   &95.318   &1.127   &95.288\\
                     & Pitch:-9.2     & $y$: 839.8      & -11.302        & 837.126   &-11.911 &834.394  &-12.314 &832.257\\
                     & Roll:-1.3       & $z$: 3598.1    & -2.142         & 3590.579  &-2.135  &3598.303  &-2.128  &3572.824\\
\hline
\multirow{3}{1em}{(c)} & Yaw:2.7         & $x$: 435.8     & -3.714        & 433.207    &-4.509   &431.473   &-4.443   &430.364\\
                     & Pitch:12.2      & $y$: -1036.2   & 16.827        & -1029.845  &16.277   &-1028.301  &16.198   &-1024.642\\
                     & Roll:-1.2       & $z$: 3553.8    & -1.915        & 3538.244   &-1.841   &3531.964  &-1.898   &3518.536\\
\hline
\multirow{3}{1em}{(d)} & Yaw:18.9        & $x$:-1212.8    & 22.741        & -1213.576  &22.503   &-1211.537   &22.265   &-1206.448\\
                     & Pitch:-0.4      & $y$: -167.8    & 3.573         & -168.660   &3.067    &-168.734    &3.050    &-167.966\\
                     & Roll:-2.3       & $z$: 3453.7    & -2.042        & 3460.211   &-2.237   &3453.575    &-2.183   &3438.054\\
\hline

\multirow{3}{1em}{(e)} & Yaw:-37.0         & $x$: 515.5      & -35.471    & 513.549   &-33.238   &518.459   &-29.508 &522.176\\
                     & Pitch:1.9         & $y$: -422.3     & 2.576      & -427.395  &2.563     &-430.459  &3.374   &-430.230\\
                     & Roll:-2.9         & $z$: 4098.4     & -2.571     & 4088.068  &-2.624    &4115.437  &-2.817  &4120.387\\
\hline
\multirow{3}{1em}{(f)} & Yaw:-28.3       & $x$:-393.6     & -27.256       & -394.989   &-26.977  &-395.335    &-25.288   &-396.326\\
                     & Pitch:2.5       & $y$: -374.2    & 3.573         & -374.406   &3.472    &-375.174    &3.615     &-377.715\\
                     & Roll:-1.2       & $z$: 3307.9    & -1.363        & 3304.232   &-1.328   &3309.443    &-1.402    &3329.823\\
\hline
\multirow{3}{1em}{(g)} & Yaw:-12.0       & $x$:-1296.5    & -11.813      & -1284.283  &-11.366  &-1285.865   &-10.421   &-1290.430\\
                     & Pitch:2.6       & $y$: -327.8    & 3.991        & -323.670   &4.033    &-324.082    &4.006     &-325.703\\
                     & Roll:-1.9       & $z$: 3102.0    & -2.317       & 3073.298   &-2.326   &3078.568    &-2.331    &3090.632\\
\hline

\multirow{3}{1em}{(h)} & Yaw:10.4         & $x$:-59.8      & 4.375       & -62.959    &4.274    &-63.164     &3.078     &-63.927\\
                     & Pitch:22.8       & $y$: -966.9    & 25.979      & -972.440   &23.945   &-973.179    &22.672    &-971.425\\
                     & Roll:-3.0        & $z$: 3728.8    & -3.420      & 3745.472   &-3.508   &3741.764    &-3.665    &3730.413\\
\hline
\parbox[c]{1.2cm}{Average Error}        & N/A            & N/A          &1.970     &5.927   &2.066    & 6.715   & 2.353   &9.432\\
\hline

 
\end{tabular}

    \label{tab:groundtest}
\end{table*}

\subsubsection{Underwater experiments}
We analyzed our pose estimation method using experiments performed in our water pool with 10m in width, 15m in length and 9m in depth, where UDID was collected. The SIA-9 (see Section \ref{systemoverview}) is employed for docking in this set of experiments. The docking station is mounted underwater in depth 2m as shown in Figure \ref{fig:infrasb}. The approaching speed of AUV is 0.5m/s on average. In order to eliminate other distractions, such as ambient light and false detection, we first shut down all ambient light in the experimental water pool, remaining only landmarks emitting light. We launched the SIA-9 at initial locations out of the scope of critical angle so that mirror images are impossible. Meanwhile, the whole docking station is assured to observe in the captured images. Under these settings, a binarization-based detection method is employed for detection.

\begin{figure}[htbp]
\captionsetup[subfigure]{justification=centering}
    \subfloat[][t=0]{
    \centering
    \includegraphics[width= 0.23\linewidth]{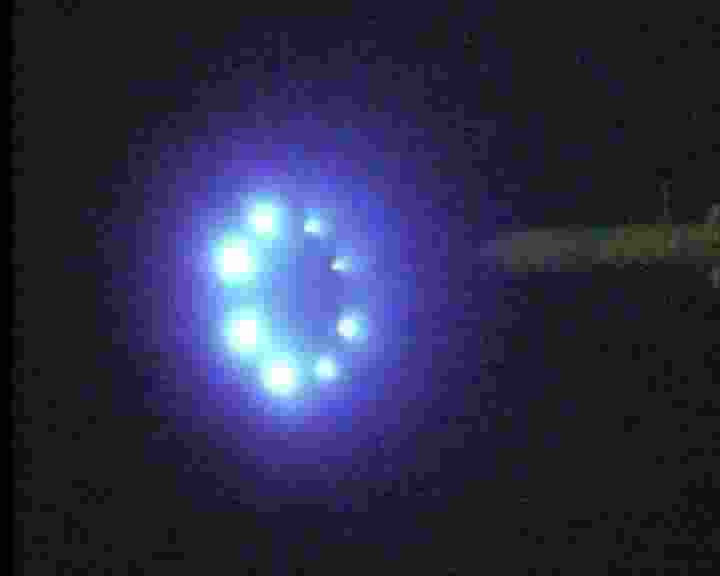}
    }
    \subfloat[][t=8]{
    \centering
    \includegraphics[width= 0.23\linewidth]{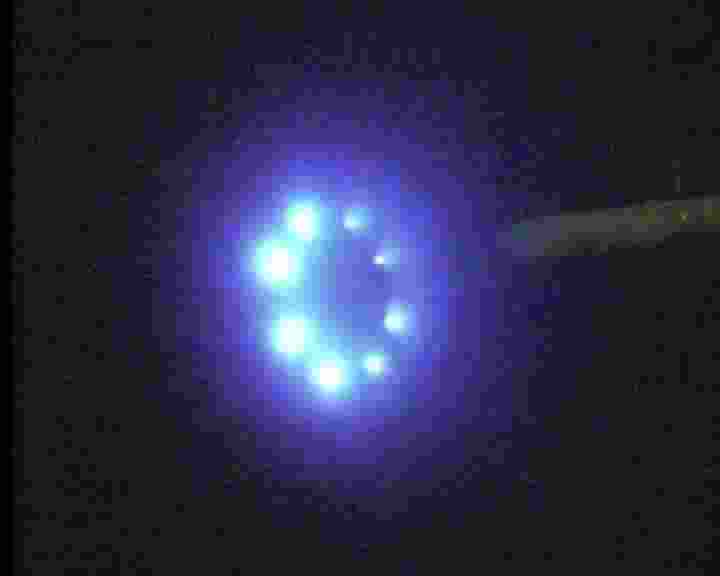}
    }
    \subfloat[][t=16]{
    \centering
    \includegraphics[width= 0.23\linewidth]{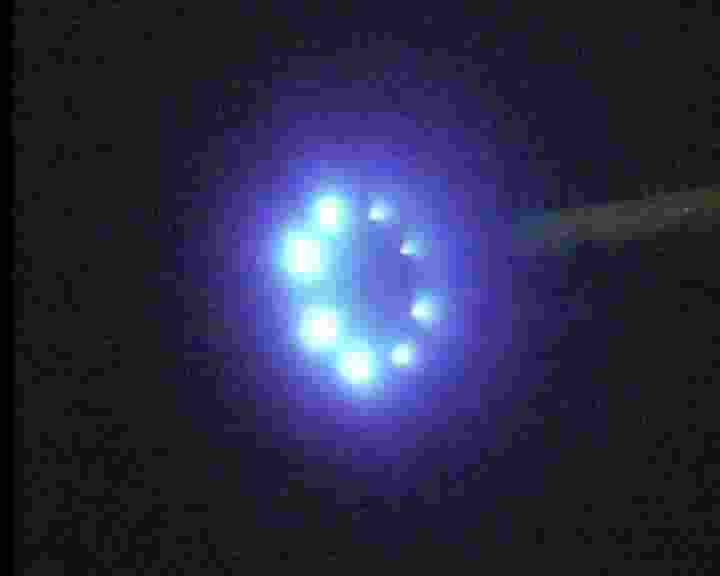}
    }
    \subfloat[][t=24]{
    \centering
    \includegraphics[width= 0.23\linewidth]{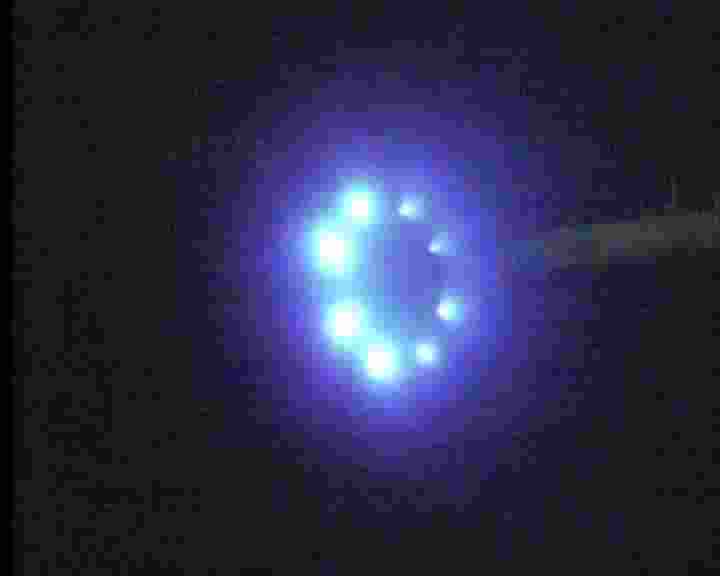}
    }\\
    \subfloat[][t=32]{
    \centering
    \includegraphics[width= 0.23\linewidth]{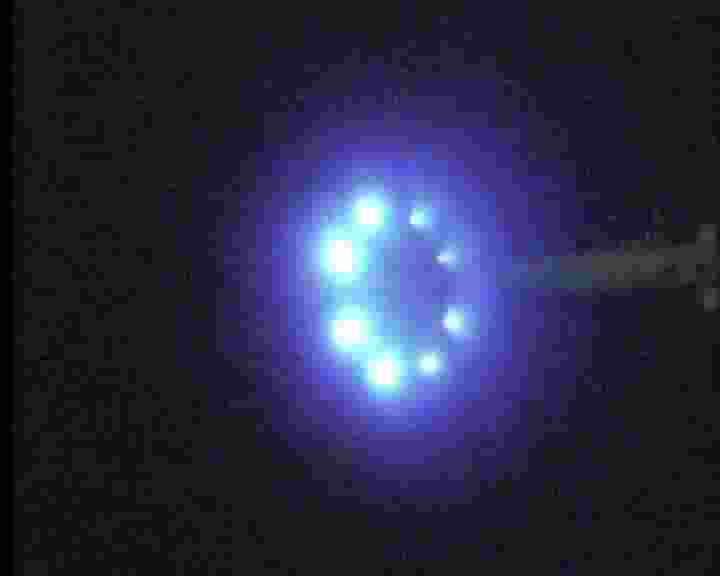}
    }
    \subfloat[][t=40]{
    \centering
    \includegraphics[width= 0.23\linewidth]{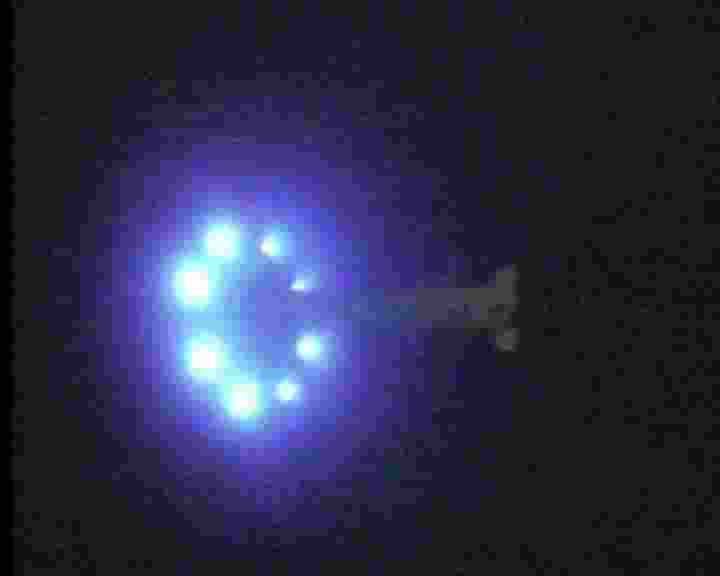}
    }
    \subfloat[][t=48]{
    \centering
    \includegraphics[width= 0.23\linewidth]{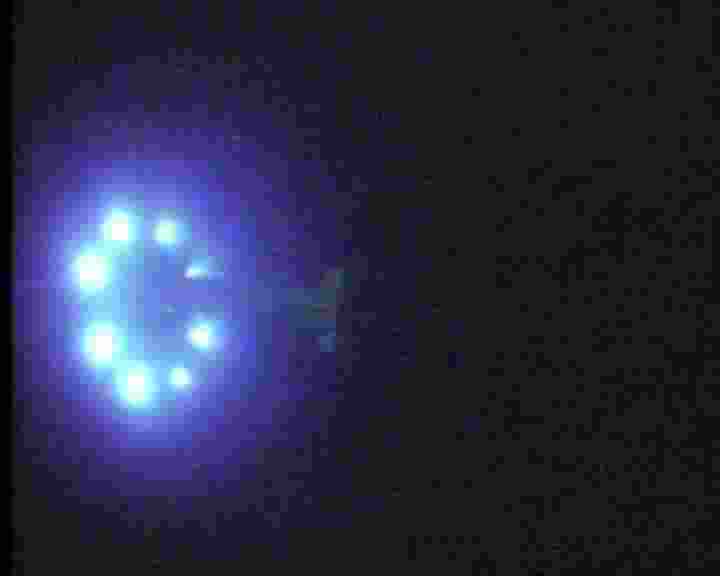}
    }
    \subfloat[][t=56]{
    \centering
    \includegraphics[width= 0.23\linewidth]{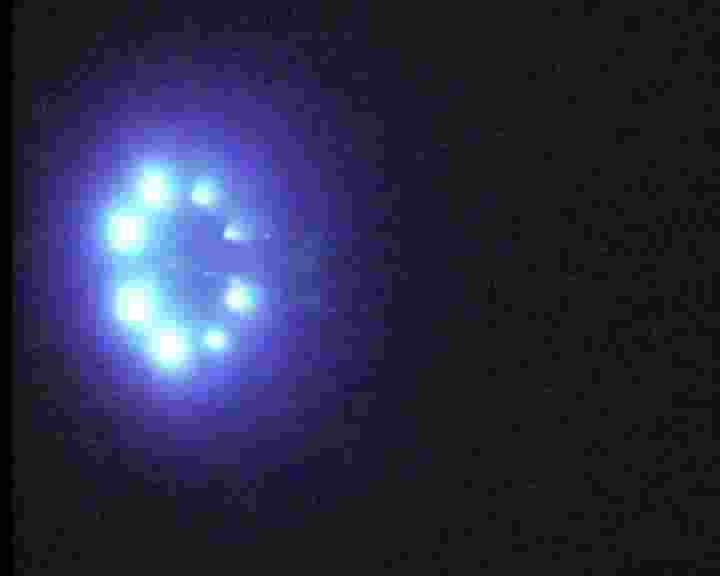}
    }
\caption{Final docking process used in our underwater experiments.}
\label{fig:underwatertest}
\end{figure}

The SIA-9 is launched at directly facing, left side and right side initial points in distance $10$m-$15$m. Figure \ref{fig:underwatertest} shows a successful docking process. A video of a demonstration of the whole docking process is given in the supplemental material\footnote{The video is also provided on the webpage \url{http://vision.is.tohoku.ac.jp/~liushuang/IJRR2017/video}.}. It is worth noting that intrinsic matrices used in the air and water are quite different from each other due to the change of medium. Thus, the camera was re-calibrated in the water before performing underwater experiments.

\subsection{Experimental results of the integrated underwater docking framework}
\label{sect:integrated}
In this subsection, we provide experimental results of our integrated underwater docking framework. We implement our underwater docking framework which integrates detection of docking stations and pose estimation. We have not implemented an integrated underwater docking framework in the SIA-9 since the SIA-9 is too small to fit a graphics processing units (GPU) device, such as NVIDIA's Jeston TX2. A GPU device helps for real-time detection of our proposed DoNN.

DoNN is implemented by using the framework Darknet \citep{darknet13} written in C. Both implementations in Matlab and C are available for pose estimation in our underwater docking algorithm. We integrate DoNN and the Matlab version of pose estimation by using Matlab.

\begin{figure}[h]
\captionsetup[subfigure]{justification=centering}
    \subfloat[][t=0]{
    \centering
    \includegraphics[width= 0.48\linewidth]{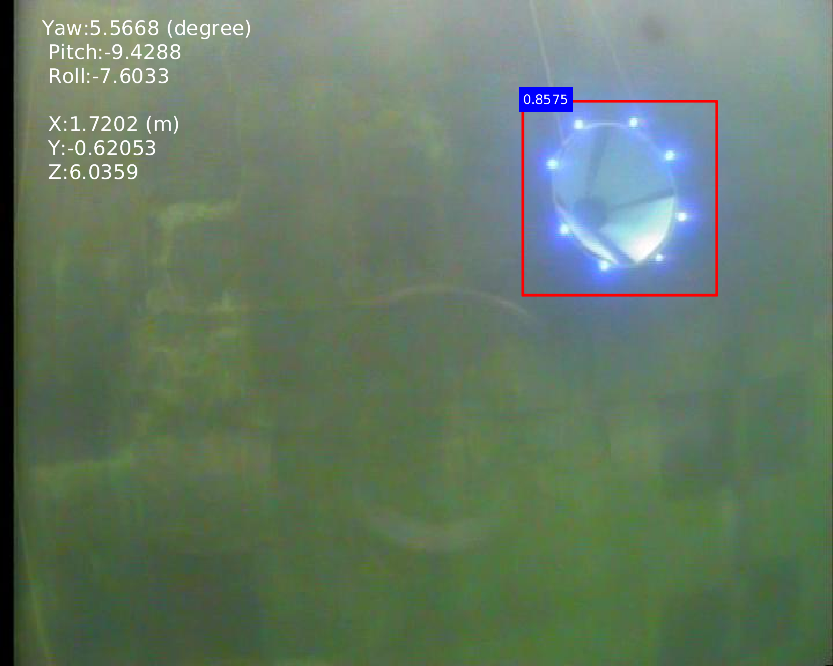}
    }
    \subfloat[][t=8]{
    \centering
    \includegraphics[width= 0.48\linewidth]{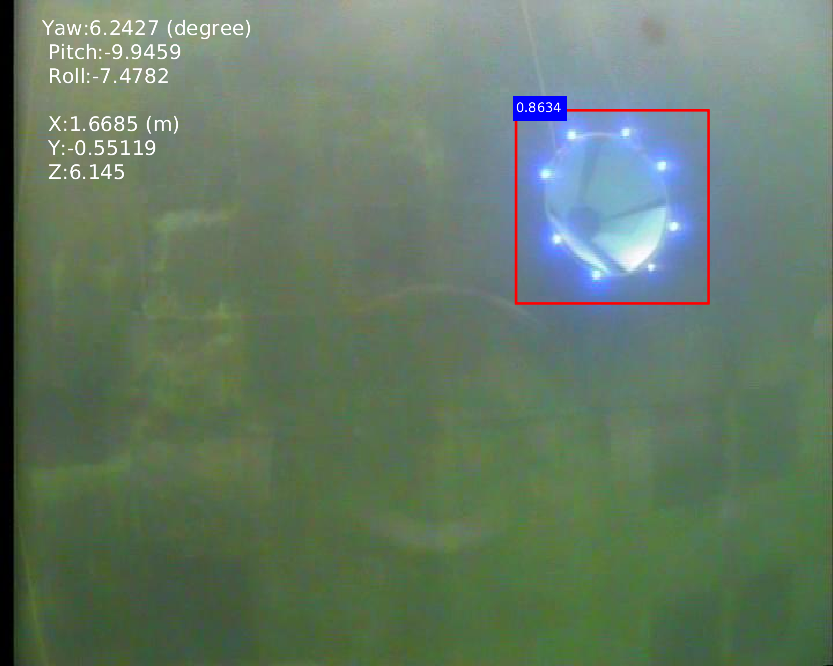}
    }\\
    \subfloat[][t=16]{
    \centering
    \includegraphics[width= 0.48\linewidth]{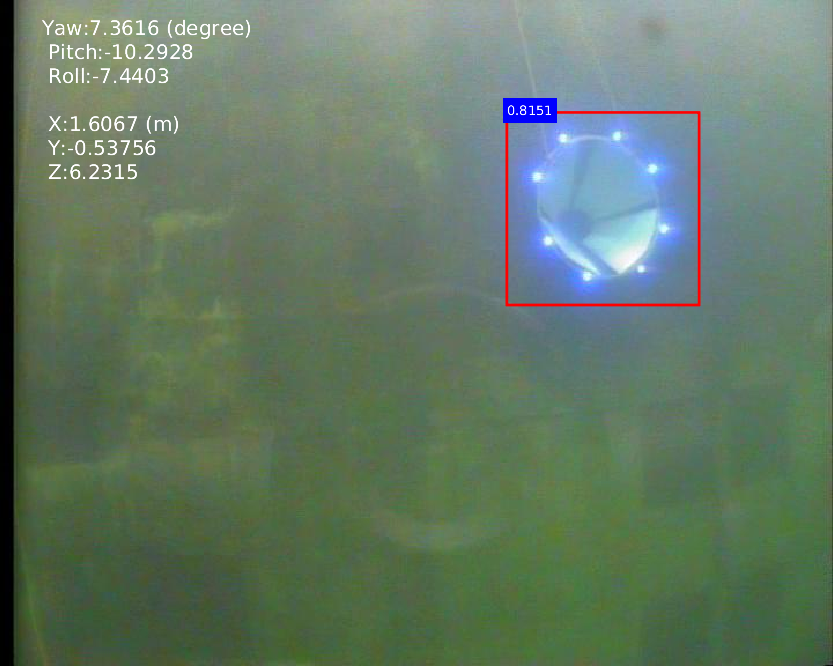}
    }
    \subfloat[][t=24]{
    \centering
    \includegraphics[width= 0.48\linewidth]{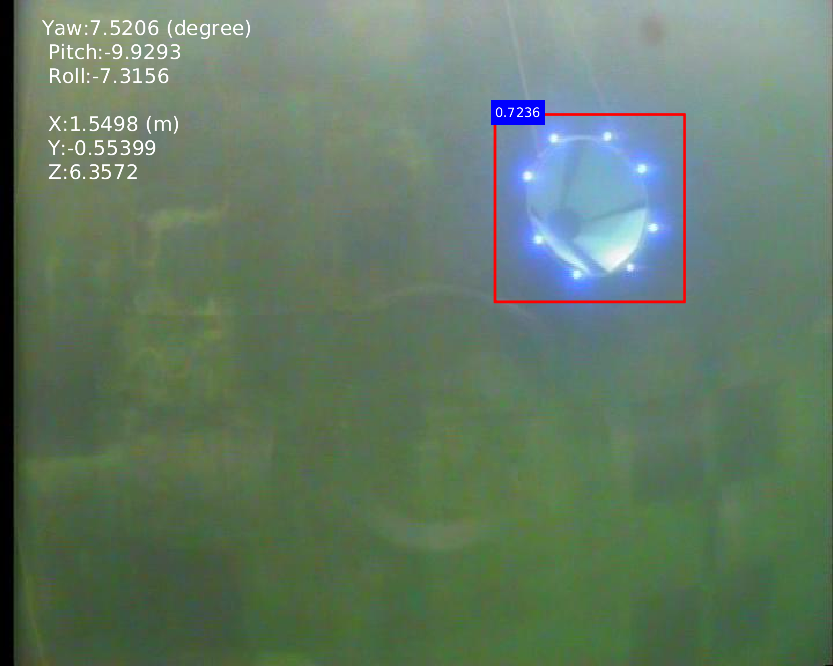}
    }\\
    \subfloat[][t=32]{
    \centering
    \includegraphics[width= 0.48\linewidth]{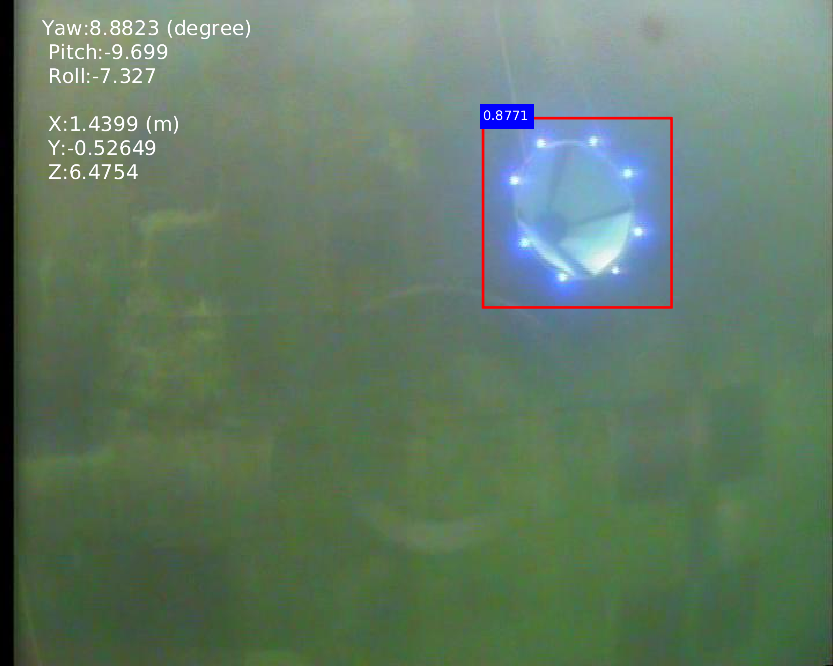}
    }
    \subfloat[][t=40]{
    \centering 
    \includegraphics[width= 0.48\linewidth]{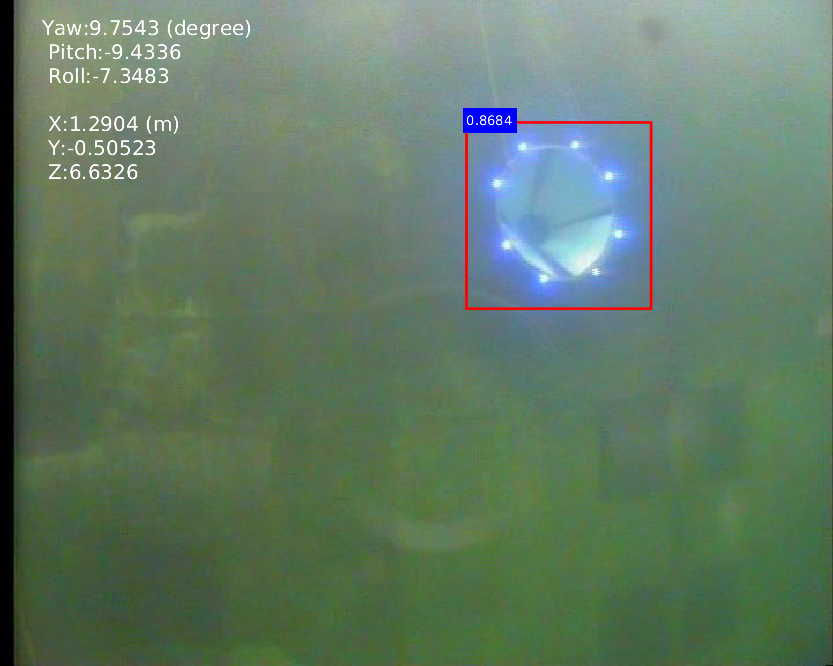}
    }\\
    \subfloat[][t=48]{
    \centering
    \includegraphics[width= 0.48\linewidth]{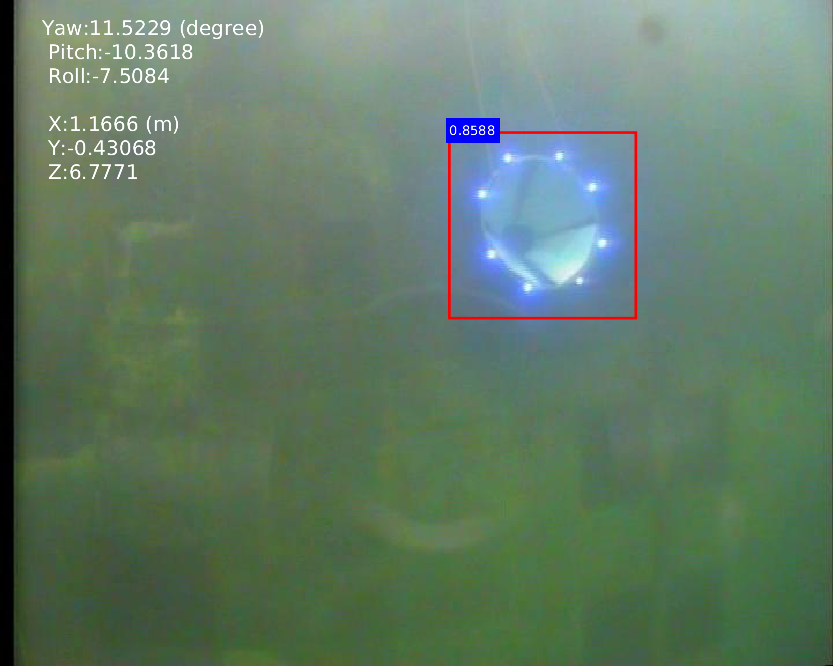}
    }
    \subfloat[][t=56]{
    \centering
    \includegraphics[width= 0.48\linewidth]{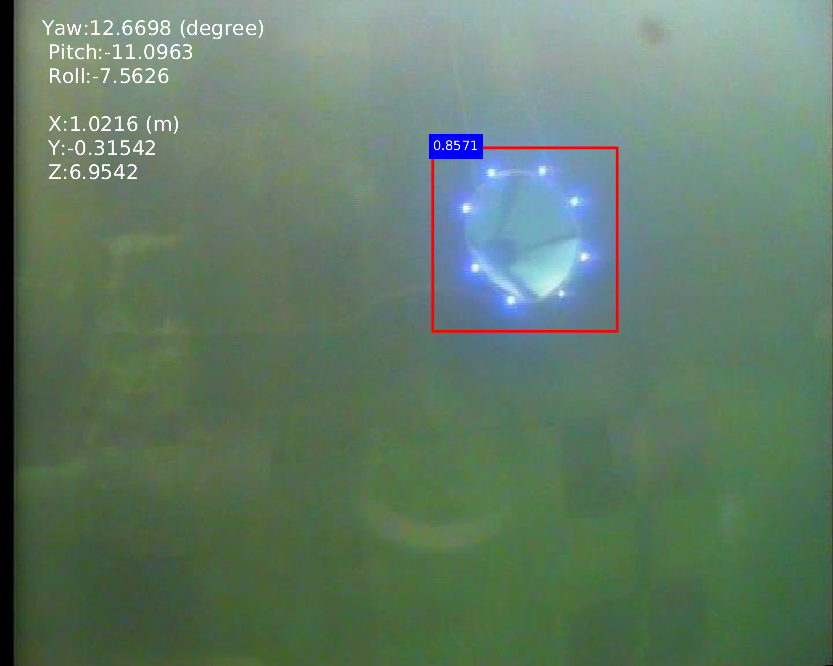}
    }
\caption{A demonstration of our integrated underwater docking algorithm. Images are captured at every eight frames as the SIA-9 is moving away from the docking station.}
\label{fig:integrateddocking}
\end{figure}

Since it is difficult to obtain ground truth of relative pose in water, we present consecutive frames in Figure~\ref{fig:integrateddocking} for a demonstration of our integrated underwater docking algorithm. Figure~\ref{fig:integrateddocking} shows that the $Z$ component of coordinate increases as the SIA-9 moves far away from the docking station. We also provide a video to present experimental results of our integrated underwater docking algorithm in our supplemental materials.
\begin{table}[!h]
    \centering
    \caption{Running time of modules in underwater docking algorithm. The detection and pose estimation modules were implemented in the desktop computer using C and Matlab, respectively. The pose estimation module was implemented in the embedded computer using C.}
\begin{tabular}{ccc}
\hline
 Module & Desktop& Embedded Computer\\
 \hline
 Detection & \multicolumn{1}{c}{0.096s/frame (C)} & \multicolumn{1}{c}{N/A} \\
 \\[-1ex]
 \shortstack[c]{Pose\\Estimation} & \multicolumn{1}{c}{\shortstack[c]{0.043s/frame\\(Matlab)}} & \multicolumn{1}{c}{\shortstack[c]{0.036s/frame\\ (C)}}\\
 \hline
\end{tabular}
\label{tab:timeconsumption}
\end{table}
Next, we present running time of detection and pose estimation modules of our underwater docking algorithm in Table \ref{tab:timeconsumption}. The desktop used for test is equipped with Intel Xeon 3.60GHz CPU and NVIDIA GTX 1080Ti GPU. The embedded computer is employed in the SIA-9, described in Section \ref{systemoverview}.

\section{Conclusion}
We introduced a vision based underwater docking framework which consists of (i) a detection module used for localization of docking stations in 2D images, and (ii) a pose estimation module used for estimation of the position and orientation between AUVs and docking stations from the docking stations detected in the images. For credible and robust detection of underwater docking stations, we proposed an algorithm, called DoNN. In order to analyze the performance of DoNN under various conditions, we provided a dataset UDID which was collected in our experimental water pool. Experiments show that DoNN achieves $0.99964$ performance in terms of AUC on UDID, and is more robust to various deformations, such as blurring, color shift, contrast shift and mirror images, compared to the baseline models in average. A perspective-n-point algorithm called RPnP, which is accurate, fast and robust to noise, is integrated to our vision based underwater docking framework for pose estimation. The running time of pose estimation is $0.043$ seconds per frame. Ground experiments show that the average error of position and orientation is $5.927$ mm and $1.970^{\circ}$, respectively, when no artificial noise is added. We observe that the average error of position and orientation is $9,432$ mm and $2.353^{\circ}$, respectively, when strong artificial noise is added. Successful underwater docking experiments were performed to demonstrate the effectiveness of pose estimation module. 

In future work, we plan to perform underwater docking experiments in the sea after equipping the SIA-9 with graphics processing units (GPU) device, and implementing our framework in the updated SIA-9. We will also explore new methods to integrate our vision based underwater docking framework with acoustic sensors which are useful for long range localization. 

\begin{acks}
We would like to thank Lingshuai Meng, Qingyong Jia, Min Jiang, Wenkang Yi and Jun Du for their help in underwater experiments.
\end{acks}

\begin{funding}
This work was partly supported by China State Key Laboratory of Robotics [grant number 2016-Z08].
\end{funding}

\begin{dci}
The Authors declare that there is no conflict of interest. 
\end{dci}

\end{document}